\pdfoutput=1

\documentclass[11pt]{article}

\usepackage[final]{acl}

\usepackage{times}
\usepackage{latexsym}

\usepackage[T1]{fontenc}

\usepackage[utf8]{inputenc}

\usepackage{microtype}

\usepackage{inconsolata}

\usepackage{graphicx}
\usepackage{algorithm}
\usepackage{algpseudocode}

\usepackage{amsfonts}   
\usepackage{graphicx}
\usepackage{textcomp}

\usepackage{tabulary}
\usepackage[most]{tcolorbox}
\usepackage{longtable}
\usepackage{enumitem}
\usepackage{breqn}
\usepackage{multirow}
\usepackage{array}
\usepackage{longtable}
\usepackage{xcolor}
\usepackage{amssymb}
\usepackage{booktabs}
\usepackage{float}
\usepackage{amsmath}

\title{MCQG-SRefine: Multiple Choice Question Generation and Evaluation with Iterative Self-Critique, Correction, and Comparison Feedback}

\author{Zonghai Yao \thanks{indicates equal contribution} $^1$, 
Aditya Parashar \footnotemark[1] $^1$, \\
\bf{Huixue Zhou} $^2$,
\bf{Won Seok Jang} $^3$,
\bf{Feiyun Ouyang} $^3$,
\bf{Zhichao Yang}  $^1$,
\bf{Hong Yu}  $^{1, 3, 4}$\\
University of Massachusetts, Amherst$^1$, University of Minnesota$^2$\\
University of Massachusetts, Lowell$^3$,
UMass Chan Medical School$^4$\\
{\{zonghaiyao, 
aparashar,zhichaoyang\}@umass.edu},
zhou1742@umn.edu, \\
{\{WonSeok\_Jang, feiyun\_ouyang, Hong\_Yu\}@uml.edu}
}

\begin{document}
\maketitle
\begin{abstract}
Automatic question generation (QG) is essential for AI and NLP, particularly in intelligent tutoring, dialogue systems, and fact verification.
Generating multiple-choice questions (MCQG) for professional exams, like the United States Medical Licensing Examination (USMLE), is particularly challenging, requiring domain expertise and complex multi-hop reasoning for high-quality questions.
However, current large language models (LLMs) like GPT-4 struggle with professional MCQG due to outdated knowledge, hallucination issues, and prompt sensitivity, resulting in unsatisfactory quality and difficulty.
To address these challenges, we propose MCQG-SRefine, an LLM self-refine-based (\textbf{C}ritique and \textbf{C}orrection) framework for converting medical cases into high-quality USMLE-style questions. 
By integrating expert-driven prompt engineering with iterative self-critique and self-correction feedback, MCQG-SRefine significantly enhances human expert satisfaction regarding both the quality and difficulty of the questions.
Furthermore, we introduce an \texttt{LLM-as-Judge}-based automatic metric to replace the complex and costly expert evaluation process, ensuring reliable and expert-aligned assessments.
~\footnote{Our code and data is released at \url{https://github.com/bio-nlp/MedQG} with CC-BY-NC 4.0 license}
\end{abstract}

\section{Introduction}
\label{Sec:intro}

In Artificial Intelligence (AI) and Natural Language Processing (NLP), automatic question generation (QG) from knowledge bases, texts, and images~\cite{guo2024survey} plays a crucial role in enhancing question-answering (QA) models~\cite{chen2023toward,guo2022dsm}, supporting intelligent tutoring systems~\cite{zhao2022educational,cai2023paniniqa}, improving dialogue systems, and aiding fact verification~\cite{pan2021zero,zhang2023towards}. 
Multiple-choice question generation (MCQG), a specialized type of QG, is extensively used in exams to assess students' knowledge efficiently~\cite{zhang2021review}. 
However, creating MCQs is labor-intensive, requiring the design of effective stems, prediction of common errors as distractors, and provision of corrective feedback~\cite{ch2018automatic}.
In professional fields, MCQs often require field experts because they need to reflect real-world scenarios and involve complex multi-hop reasoning.
These are unique challenges not typically encountered in general QG tasks. 

The United States Medical Licensing Examination (USMLE) exemplifies the need for high-quality MCQG~\cite{usmle_summary}. 
Preparing for the USMLE costs medical students over \$5000 on average~\cite{usmle_prepare_cost}. 
For exam boards and instructors, creating MCQs is both time-consuming and expensive~\cite{gierl2012using}. 
Any application that can automate this process is highly valuable to medical educators~\cite{homolak2023opportunities,gilardi2023chatgpt}. 
Due to the high difficulty with the need for domain knowledge and complex reasoning, USMLE questions are becoming important large language models (LLMs) benchmarks~\cite{jin2021disease}.
Top LLMs like GPT-4 have shown over 90\% accuracy on sample USMLE questions~\cite{achiam2023gpt}.
Recent research explores leveraging GPT-4's potential in USMLE-MCQG to improve question generation efficiency for educators and assist students in exam preparation~\cite{klang2023advantages,agarwal2023analysing,biswas2023passing}.

\begin{figure*}[!ht]
    \vspace{-5mm}
    \centering
    \includegraphics[width=\textwidth]{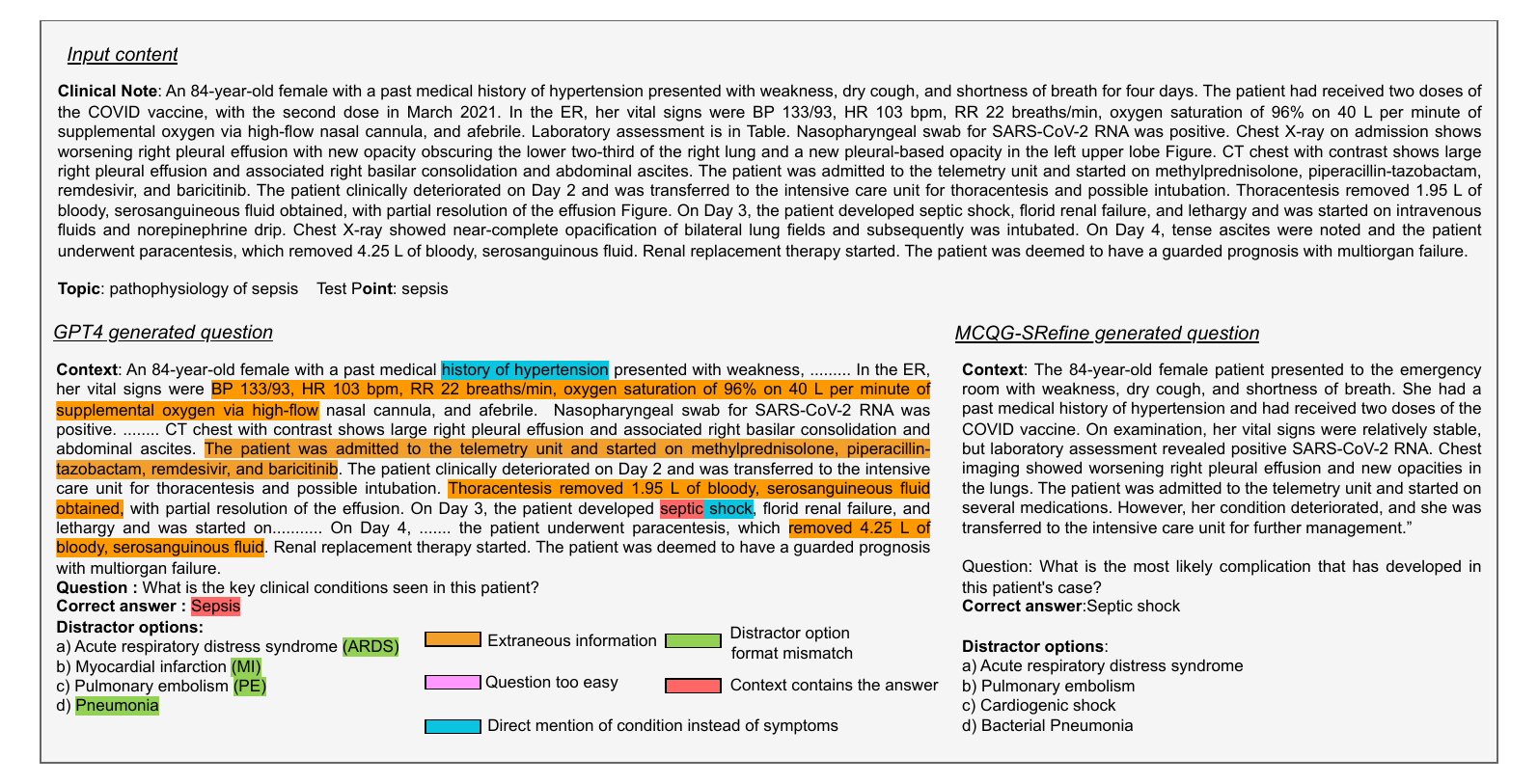}
    \vspace{-8mm}
    \caption{USMLE MCQ generated by GPT-4 and MCQG-SRefine.
    The GPT-4 question contains several errors and inconsistencies, such as extraneous information, a distractor option format mismatch, mentioning symptoms instead of conditions, and a context that contains the answer. 
    The MCQG-SRefine addresses these issues, resulting in a higher quality question that aligns the context, question, and answer options more coherently. Irrelevant details are removed, the question focuses on the key clinical condition of sepsis, distractor options are presented in a consistent format, and the context no longer gives away the answer.}
    \label{fig:mcqg_SRefine}
    \vspace{-5mm}
\end{figure*}

However, relying solely on LLMs like GPT-4 to generate USMLE questions presents several challenges.
Firstly, their performance is constrained by their training data, leading to two major issues: outdated knowledge~\cite{mousavi2024your} and hallucination~\cite{zhang2023siren} (\textbf{Limit1}). 
Outdated knowledge means that LLMs can only repeat or integrate old USMLE questions or medical documents in their training data during the generation. Consequently, they struggle to create new questions based on the latest medical cases or guidelines like medical experts. 
Hallucination refers to LLMs potentially providing misinformation in questions, which could harm students' learning outcomes.
Secondly, generating questions on specific concepts requires precise prompts (\textbf{Limit2}), which students might not know how to formulate~\cite{sahoo2024systematic}. 
For example, given one USMLE question about cardiovascular physiology, if GPT-4 is asked to "\texttt{Generate a more difficult USMLE question.}", it only provides a longer stem. 
It will focus on the specific topics and test points only when specifically asked to generate a question "\texttt{testing knowledge of the hemodynamic profile of aortic stenosis}."
Lastly, the quality and difficulty of the questions often do not meet expectations~\cite{artsi2024large,benitez2024harnessing} (\textbf{Limit3}). 
As shown in Figure~\ref{fig:mcqg_SRefine}, the output from GPT-4 contains several errors and inconsistencies, such as extraneous information, mismatched distractor option formats, direct mention of the condition instead of symptoms, and context that includes the answer. 
The generated questions often lack the depth required for students' critical thinking.

To address these challenges, we propose a new task: converting medical cases that appear in case reports or clinical notes into USMLE-style MCQs.
Our approach involves several key design elements to alleviate the above limitations:

For \textbf{Limit1}, previous research has attempted to prompt LLMs to generate USMLE-MCQs by following expert-crafted instructions in zero-shot setting or further adding existing questions as few-shot examples~\cite{artsi2024large}. 
To our knowledge, we are the first to qualitatively and quantitatively study how to convert medical cases into USMLE-MCQs. 
These medical cases provide valuable real-world information on disease progression, accurate assessments, diagnoses, and potential treatment plans. 
Using the latest medical cases as input, LLMs can generate up-to-date questions, thereby minimizing the limitations of outdated knowledge. 
Additionally, grounding questions' key elements in the original medical cases can help LLMs reduce hallucinations, enhancing the reliability of the generated content.

For \textbf{Limit2}, we follow the National Board of Medical Examiners (NBME) guidelines~\footnote{\url{https://www.nbme.org/educators/item-writing-guide}} to establish a checklist of 41 target topics covering all potential exam areas.
We then deployed a ColBERT retriever~\cite{santhanam2021colbertv2} using USMLE Content Outline~\footnote{\url{https://www.usmle.org/sites/default/files/2021-08/USMLE_Content_Outline.pdf}} as a collection for test points retrieval.
Each input medical case in our experiments is evaluated by experts with exam experience to determine if it contains sufficient information to generate questions related to the specified target topics and test points. 
We also compared the topics and test points identified by experts with those generated by LLMs, assessing their impact on the quality and difficulty of the resulting questions.

For \textbf{Limit3}, as illustrated in Figure~\ref{fig:mcqg_SRefine}, we used the triplets (medical case, topic, and test point) as our question generation pipeline input.
We work with experts for prompt engineering based on USMLE guidelines.
We then created our MCQG-SRefine (e.g., self-refine with iterative Critique and Correction feedback) following three steps.
S1 - Initial MCQ Generation: Generate an initial USMLE-MCQ based on the triplets.
S2 - Critique Feedback: Prompt the LLM itself to provide feedback on the S1 USMLE-MCQ.
S3 - Correction Feedback: Correct the USMLE-MCQ based on the S2-generated critique feedback.
Through iterative critique and correction, MCQG-SRefine significantly enhances the quality and difficulty of generated USMLE-style MCQs. 
Human evaluations confirm its effectiveness, showing a strong preference for MCQG-SRefine generated questions, with a preference ratio of 72.5\% in win, 10\% tie, and 17.5\% loss when compared to GPT-4 generated questions.
In terms of difficulty, MCQG-SRefine generates more challenging questions. Specifically, when provided with expert-identified topics and test points, there is an 80\% reduction in easy questions, a 2.25-fold increase in medium questions, and a 4-fold increase in hard questions.

\begin{figure*}[hbt!]
    \centering
    \vspace{-5mm}
    \includegraphics[width=\textwidth]{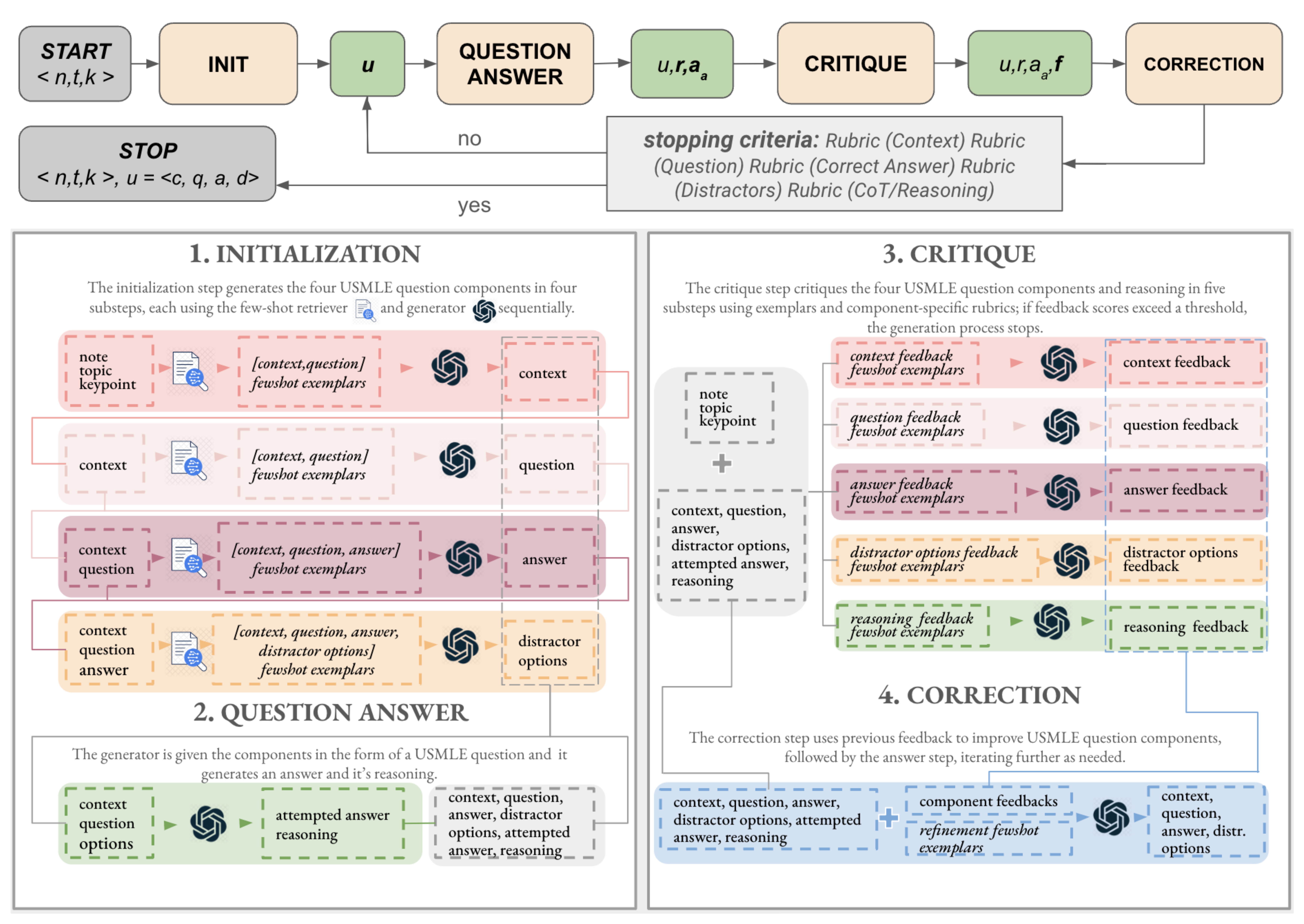}
    \vspace{-8mm}
    \caption{The framework for generating USMLE-style questions involves four main steps, as illustrated in the figure. First, the initialization generates the context, question, answer, and distractor options using retrieval and generation models. The generation model then answers the generated question along with a reasoning. Next, the feedback step evaluates the generated components on various rubrics and generates textual feedback and scores, stopping if feedback scores exceed a threshold. Finally, the refine step iterates by using the feedback to improve the components before cycling back to the answer step.}
    \vspace{-5mm}
    \label{fig:usmle_qg_ppl}
\end{figure*}

Finally, designing a reliable reference-free metric to automatically evaluate the quality of system-generated USMLE-MCQs is challenging. Recent research indicates that \texttt{LLM-as-Judge} correlates more closely with human evaluations than traditional metrics~\cite{chen2023storyer,chiang2023can,kocmi2023large,zheng2024judging,zhang2024comprehensive,kim2023prometheus,kim2024prometheus}, though these methods remain underexplored in medical NLP tasks. 
Our goal is to replace the costly expert evaluation process in USMLE-MCQG with \texttt{LLM-as-Judge}.
We used 30 criteria designed by experts in their human evaluation, covering different aspects of USMLE questions, to guide \texttt{LLM-as-Judge} in providing rating or comparison feedback on different systems' questions. 
By further exploring the filtering methods for the 30 criteria, we finally screened out a combination of 10 key criteria.
This improved the correlation between \texttt{LLM-as-Judge} and expert evaluations, as measured by Cohen's kappa, from 0.226 (slight reliability) to 0.539 (moderate reliability).
Using the results from this automated evaluation system, it was shown that the preference rate for questions generated by MCQG-SRefine over those generated by GPT-4 was 79.97\% in favor and 20.03\% against.
Moreover, MCQG-SRefine demonstrated overall improvements across 10 criteria within the 5 components, not just in a specific area.

\section{Method}
\label{Sec:method}

\textbf{Problem statement:} 
Given a medical case $n$ detailing a patient's history, diagnosis, treatment, and outcome, we aim to generate a USMLE question $u$. Here, $u =\textlangle c,q,a,d\textrangle $ consists of a context ($c$), which is a modified excerpt from $n$ tailored to align with the target style and obscure evidence information that can easily lead directly to the correct answer; a question ($q$) based on the generated context, which may be one or several sentences; the correct answer ($a$) to this question; and several distractor options ($d$).
    
\subsection {Topic and test point identification}
\label{Sec:topic_testpoint}
As discussed in \textbf{Limit2}, generating questions using LLMs without specific guidance, such as defined topics \(t\) and test points \(k\), often results in questions that lack relevance, quality, and appropriate difficulty. These questions may fall outside the scope of the USMLE exam, being either too simple or overly complex. Therefore, the quality and difficulty of the generated USMLE questions are significantly influenced by the selection of topics \(t\) and test points \(k\).

\noindent\textbf{Topics \(t\)} refer to a list of target topics selected from 41 potential topics outlined in the NBME official guidelines, categorized into 10 sections~\footnote{Details can be found in Appendix~\ref{apx:topic_testpoint}.}. Both the LLM and human experts are provided with this list to generate a maximum of five topics (\(t\)) that are highly relevant to the medical case.

\noindent\textbf{Test points \(k\)} refer to the core concepts closely related to the correct answer. We employ the ColBERT retriever~\footnote{\url{https://github.com/stanford-futuredata/ColBERT}}, denoted as \(\pi_{rtr}\), to retrieve suitable test points from 18 sections~\footnote{Details can be found in Appendix~\ref{apx:topic_testpoint}.} of the USMLE content outline.
This test points retrieval process involves querying the model with the medical case (\(n\)) and the identified topics (\(t\)). The result is a list of highly specific USMLE concepts, such as 'keloid formation'. \(\pi_{llm}\) then generates specific test point concepts using these filtered concepts as a foundation, ensuring they are directly relevant to the context (\(c\)) and topics (\(t\)). These test points can either originate from the filtered list of USMLE concepts or be derived from the content of the medical case itself. Additionally, we ask human experts to identify specific test points they believe are related to their selected topics, referring to the key medical concepts mentioned in the USMLE content outline. 
More details can be found in Appendix Table~\ref{tab:human_topic_keypoint_example}. 
In Section~\ref{Sec:main_results}, we compare the topics and test points generated by humans and LLMs for their impact on the quality of the resulting questions.

\subsection{Initialization}
\label{Sec:init}
As illustrated in Figure~\ref{fig:usmle_qg_ppl}, the MCQG-SRefine pipeline begins with the INIT step, which comprises 4 generation steps, each targeting a component of the goal $u = ⟨c, q, a, d⟩$.
To assist the model in referencing similar examples for better generation of each component of $u$, we deploy a ColBERT retriever model $R$ to retrieve a small set of USMLE examples from the MedQA~\cite{jin2021disease} question bank.
As shown in Figure~\ref{fig:usmle_qg_ppl}, given the input $<n, t, k>$ from Topic and Test point identification ste, $R$ first uses $<n, t, k>$ as a query to retrieve few-shot examples, and then LLM follows the INIT-c prompts in Appendix Table~\ref{prompt} to generate the context $c$. 
Subsequently, after obtaining $c$, we continue to use $<n, t, k, c>$ as a query to retrieve few-shot examples and follow the INIT-q prompts in Appendix Table~\ref{prompt} to generate the question $q$.
The exact process is applied for $t$ and $k$. 
It is important to note that we trim the retrieved examples for each component; for instance, in INIT-c prompts, we only retain the context component of each example, and similarly, for the other three components $q, a, d$, only the relevant parts are kept.
As demonstrated in Appendix Table~\ref{tab:demonstration_MCQG-SRefine}, this INIT step already results in a USMLE MCQ for a given input $<n, t, k>$. 
However, as discussed in Figure~\ref{fig:mcqg_SRefine}, despite incorporating several advanced prompting engineering methods in the INIT step—including prompts designed by medical experts according to USMLE guidelines, tailored topics\&test points for each input medical case, as well as step-by-step retrieval and generation—the LLM-generated USMLE MCQs in the INIT step still fall short of the required quality and difficulty. 

\subsection{Question Answering Feedback Collection}
\label{Sec:qa}
Inspired by recent work that augments the standard QG model with an additional QA module to further constrain the generated questions~\cite{su2022qa4qg, xie2020exploring, sun2019joint}, we add a Question Answering Feedback Collection module. This provides additional feedback from the question-answering perspective to further challenge the quality and difficulty of the questions.
Our motivation stems from the fact that LLMs like GPT-4 have proven to perform exceptionally well on USMLE QA tasks, achieving human-expert levels in both accuracy and reasoning processes. By analyzing the rationale and correctness of the final answers produced by the language models during the QA process, we can gather valuable insights into the quality and difficulty of the questions.
Specifically, given the context $c$, question $q$, and options composed of $a \cup d$ (with their order shuffled), we collect the LLM's generated attempt $a_a$ along with the reasoning $r$ that supports $a_a$ in this step. An example output is shown in Appendix Table~\ref{tab:demonstration_MCQG-SRefine}.

\subsection{Critique}
\label{Sec:critique}
After generating all components in INIT and QA step, the LLM is asked to critique each component. 
The set to be critiqued is $u_{\text{critique}} = \langle c, q, a, d, r \rangle$. 
The LLM receives a scoring guide $G$, which includes all aspects that need to be evaluated for each component.
The prompt includes this guide $G$ as well as several manually written example critiques of scored components $E^{\text{fs}}_{\text{critique}} = \langle e^{\text{fs}}_{c}, e^{\text{fs}}_{q}, e^{\text{fs}}_{a}, e^{\text{fs}}_{d}, e^{\text{fs}}_{r} \rangle$ and $\langle G, n, t, k, c, q, a, d, a_a \rangle$. 
The final output of this step is LLM critique feedback on all components, $f = \langle f_c, f_q, f_a, f_d, f_r \rangle$, which includes short text critiques and scores for each aspect.
The aspects for scoring different components in $G$ are as follows:
\textbf{Context:} Relevant, Concise, Coherent, Consistent, Specific, Fluent, Clueing, Completeness, and Misdirection. 
\textbf{Question:} Relevant, Clear, Concluding, Difficulty, and Clarity.
\textbf{Correct Answer:} Relevant, Occurrence, Justification, Depth of Understanding, and Prevention of Guesswork.
\textbf{Distractors:} Format, Length, Relation, Variation, Plausibility, Differentiation, and Common Mistakes.
\textbf{Chain of Thought/Reasoning:} Logical Flow, Evidence-Based Reasoning, and Consideration of Options.
We provide detailed explanations for each aspect of every component in Appendix Table~\ref{tab:demonstration_MCQG-SRefine}, and LLM-Critique prompts in Table~\ref{prompt}. 
The total score for each component is calculated by summing up all individual aspect scores. A sample output is provided in Appendix Table~\ref{tab:demonstration_MCQG-SRefine}.

\subsection{Correction}
\label{Sec:correction}
The Correction step aims to correct each of the generated components of $u$ in the INIT step. 
LLM is prompted with $\textlangle E^{fs}_{correction},n,t,k,f,c,q,a,d,a_a,r \textrangle$ and asked to generate $u_{correction}$, which can perform better on all the component's critique aspects. Here  $E^{fs}_{correction}$ is a set of manually written few shot examples which are incrementally improving using the previous output's feedback.
$u_{correction}$ is again given to the Critique step to check if the feedback scores are greater than a fixed threshold, which, if true, stops the iterative Critique and Correction and, if not, continues~\footnote{Stopping criteria can be found in Appendix~\ref{apx:stop}.}.

\section{Experimental Design and Setup}
\label{Sec:experiments}

\paragraph{Experimental Setup}
For all our experiments with MCQG-SRefine, we use the chat completions API from OpenAI and the \texttt{gpt-4-0125-preview}, which has a context window of 8192 tokens, and the values of the hyperparameters temperature and top-p are set to 1. Similarly, for all other models used for the \texttt{LLM-as-Judge} comparison feedback generation, we used their default hyperparameters.

\paragraph{Dataset}
\label{dataset}
For the medical cases utilized in the generation of USMLE questions, we used unidentified patient summaries from the PMC-Patients dataset \cite{Zhao_2023}. The average length of these patient summaries was \textasciitilde 419 words. The frequency of topics used is listed in Appendix Table \ref{tab:hum_frequency} \ref{tab:mach_frequency}.

\paragraph{Experimental Design}
Our experimental design is motivated to answer the following research questions:
Evaluate whether MCQG-SRefine improves both the quality and difficulty of the generated questions.
Specifically, we employed the MCQG-SRefine pipeline to generate USMLE questions and compared them with baseline questions generated by GPT-4 under identical inputs and settings. Our inputs consisted of medical cases \(n\) from the PMC-Patients dataset, with topics \(t\) and test points \(k\) that were either human-annotated or generated by the LLM. We generated 373 questions from the human-annotated \(\langle n, t, k \rangle\) set and 385 questions using the LLM-generated set.

To assess the quality of the questions (RQ1), we first engaged two medical experts~\footnote{Two medical students with 2+ years hospital experience.} to express their preferences between the two sets (GPT-4 and GPT-4 + MCQG-SRefine) of system-generated questions based on an annotation guideline (Appendix Table~\ref{tab:human_eval_guide}). 
The evaluators were blinded to the source of the questions, and the order of the questions was randomized for each data point. 
We calculated the Percentage Agreement (87.5\%) and Cohen's kappa (0.66722) for the two evaluators' preferences, indicating substantial reliability of our human evaluation settings. 
Subsequently, a third human expert~\footnote{One licensed physician.} facilitated discussions with the initial evaluators to make the final decision for each data point, representing the final human expert preference (referred to as \texttt{Expert X}).

For evaluating the difficulty level of the questions (RQ2), the human evaluators were also asked to classify the difficulty of both questions into one of three categories: Easy, Medium, and Difficult.
Specifically, we randomly selected 50 real-world USMLE-style questions from the AMBOSS~\footnote{\url{https://www.amboss.com/us}} dataset (10 for each difficulty level) as examples for the experts to reference. AMBOSS categorizes question difficulty from 1 to 5, where 1 is the easiest and 5 is the most difficult. We grouped levels 1 and 2 as Easy examples, levels 3 and 4 as Medium examples, and level 5 as Hard examples.

\paragraph{LLM-as-Judge for evaluation metrics} 
In addition to human evaluation, recent work has demonstrated that LLM-as-Judge (particularly GPT-4-based) has a high correlation with human assessment in reference-free settings~\cite{chen2023storyer,chiang2023can,kocmi2023large,zheng2024judging,zhang2024comprehensive,kim2023prometheus,kim2024prometheus,lan2024criticbench}. 
Recent work in the medical domain has also shown the potential of LLM-as-Judge to replace traditional evaluation metrics~\cite{yao2024medqa,schmidgall2024agentclinic,mehandru2024evaluating}.
In this work, finding reliable automatic evaluation metrics in a reference-free setting is crucial, as it can reduce the burden of expert evaluation and help improve LLMs in future work (e.g., as a reward model). To achieve this, we explored two common LLM-as-Judge modes: rating and comparison. 
Regarding the evaluation criteria for each part of the MCQ, we found that directly using the criteria from the critique section of MCQG-SRefine did not correlate well with the human evaluation results of \texttt{Expert X}.
Therefore, we conducted a detailed correlation analysis between each criterion's score and human evaluation, ranking them accordingly. Based on this analysis, we applied different filtering methods to identify the most relevant combination of criteria~\footnote{Details can be found in the Discussion section~\ref{discussion}}.
Finally, we selected the following ten aspects: Context (concision, relevance, misdirection), Question (concluding, clarity), Correct Answer (occurrence, depth of understanding), Distractor (common mistakes), and Reasoning (logical flow, evidence-based reasoning). 
Specifically, GPT4-as-judge provides two evaluation indicators: 1. Detailed ratings and reasons for the above five sections and ten aspects; 2. Preference between MCQA-SRefine and GPT-4 generated questions.

\section{Results}
\label{Sec:main_results}

\begin{figure}[!t]
    \centering
    \vspace{-5mm}
    \includegraphics[width=\linewidth]{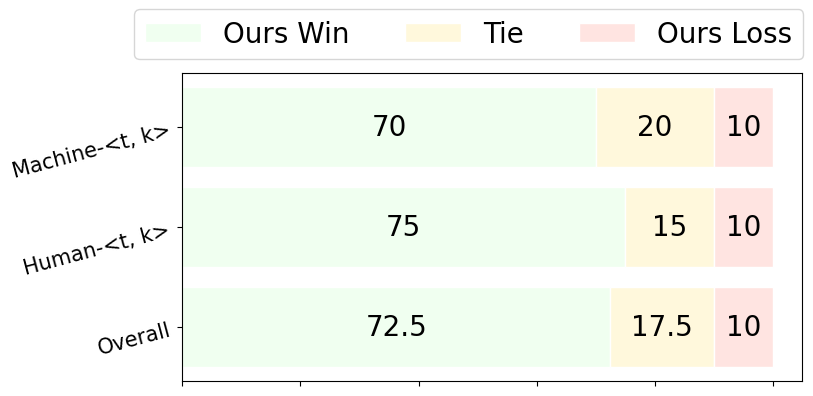}
    \vspace{-5mm}
    \caption{The quality expert preference for the GPT-4 and the GPT-4 + MCQG-SRefine question. 
    The data is divided into Human and Machine based on how the topic $t$ and key points $k$ were generated. 
    We only put the final Expert X preferences here, but we provide more results in the Appendix Table~\ref{tab:evaluations}.
    The percentage agreement between experts is 87.5\% (Human<$t$, $k$>: 90\%, Machine<$t$, $k$>: 85\%).
    The Cohen's kappa between experts is 0.66722 (Human<$t$, $k$>: 0.75, Machine<$t$, $k$>: 0.57), indicating substantial reliability.}
    \vspace{-5mm}
    \label{fig:human_eval_quality}
\end{figure}

\begin{figure}[!t]
    \vspace{-5mm}
    \centering
    \includegraphics[width=\linewidth]{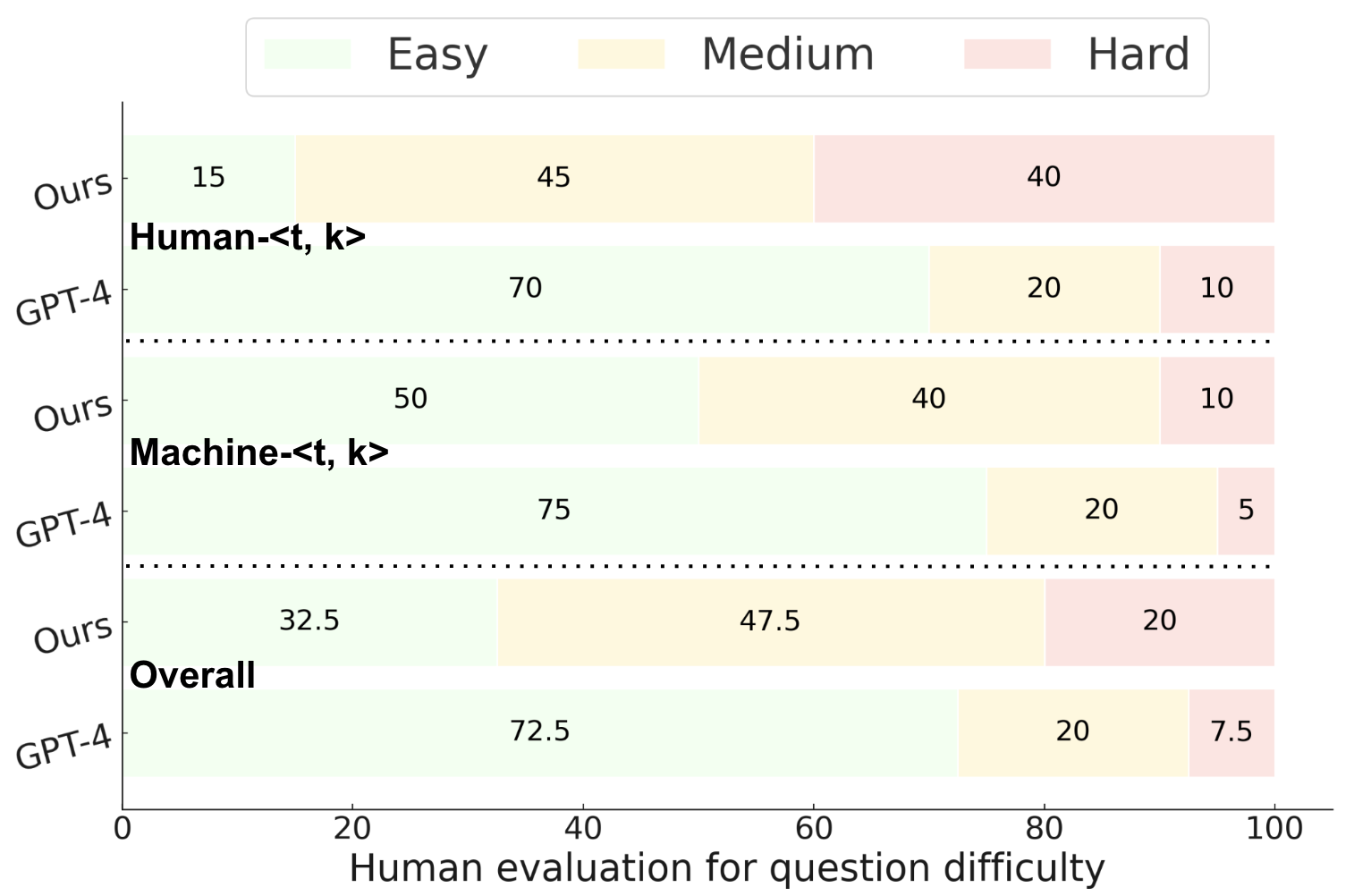}
    \vspace{-5mm}
    \caption{The difficulty expert evaluation for the GPT-4 generated and the GPT-4 + MCQG-SRefine questions.}
    \vspace{-5mm}
    \label{fig:human_eval_difficutly}
\end{figure}

\paragraph{Main results}
Figure~\ref{fig:human_eval_quality} demonstrates the overwhelming advantage of GPT-4 + MCQG-SRefine over GPT-4 with a 70-80\% win rate in human preference about question quality (RQ1). 
In Figure~\ref{fig:human_eval_difficutly}, we also observe a decrease of about 80\% in easy questions, a 2.25 times increase in medium questions, and a 4 times increase in hard questions with the input of these medical cases with expert-provided topics and key points.
For machine-provided topics and key point cases, the proportion of easy questions decreased by 33.3\%, a 2 times increase in medium questions, but there was no increase in the proportion of hard questions (RQ2).
This demonstrates the effectiveness of MCQG-SRefine in increasing the difficulty of questions.
This also indicates that the quality of topics and key points provided by experts is higher, so LLM can generate more difficult questions by thinking more deeply during the critique and correction steps.
Further improving the quality of the machine-provided topics and key points can be a future direction for improvement.

\begin{figure}[!t]
    \centering
    \vspace{-5mm}
    \includegraphics[width=\linewidth]{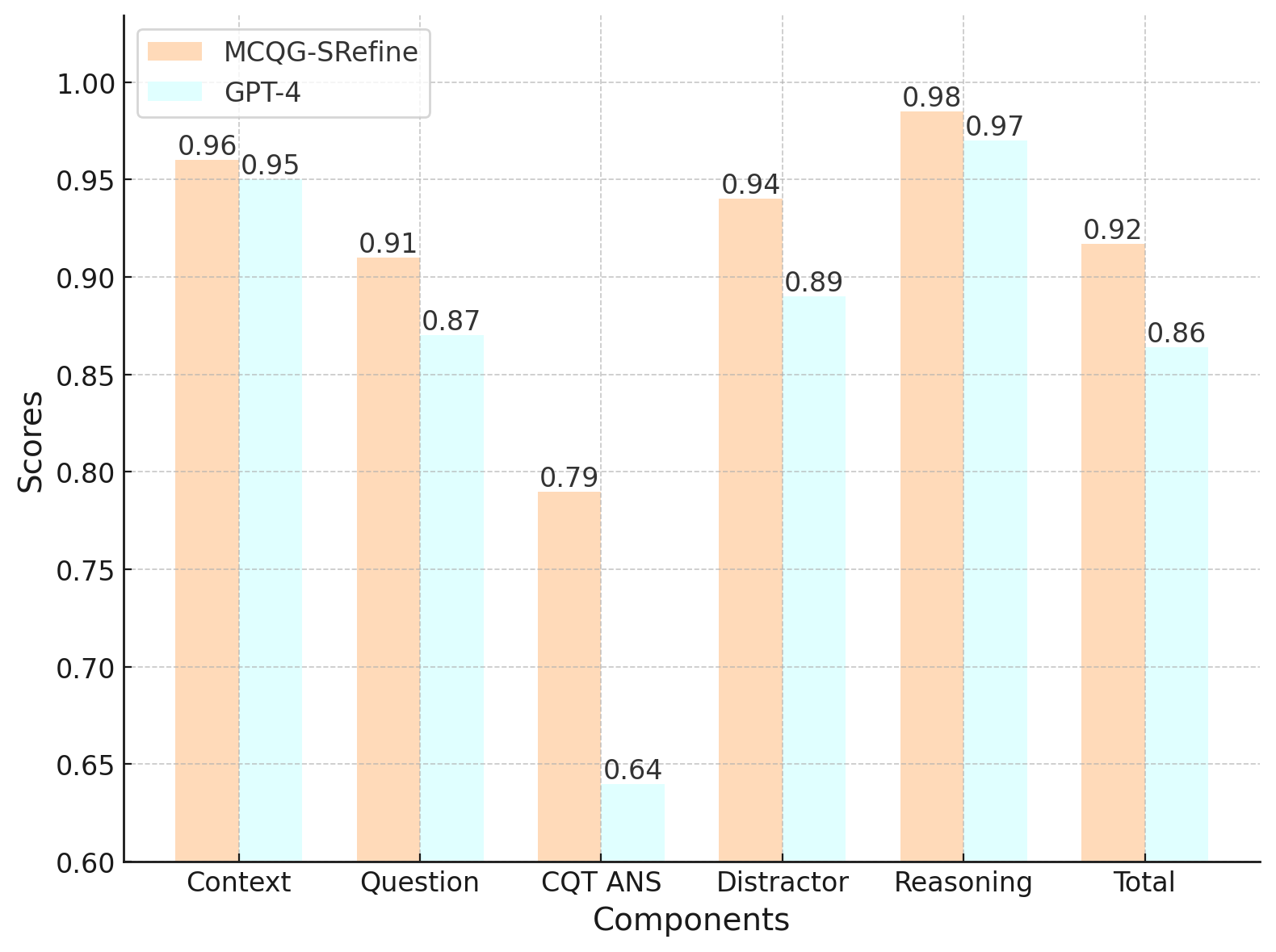}
    \vspace{-6mm}
    \caption{LLM-as-Judge (Rating) results for different components (e.g., Context, Question, Correct Answer, Distractor, Reasoning) and the final score.
    }
    \vspace{-5mm}
    \label{fig:LLM_judge_rating}
\end{figure}

Figure~\ref{fig:LLM_judge_rating} and Table~\ref{tab:LLM_judge_preference} present the results of the LLM-as-judge evaluation.
For human-provided topics and key point cases, GPT-4 + MCQG-SRefine achieved a win rate of 79.8\% compared to 20.2\% for GPT-4. 
Similarly, for machine-provided topics and key point cases, GPT-4 + MCQG-SRefine achieved a win rate of 80.1\%, outperforming GPT-4's 19.9\%.
Notably, GPT-4 + MCQG-SRefine consistently outperformed GPT-4 across all five evaluated components with LLM-as-judge (Rating) results rather than demonstrating an advantage in only a single aspect~\footnote{The aspect-level score can be found in Appendix Table~\ref{tab:llm-as-judge-results}.}. 
This indicates a balanced and comprehensive improvement of GPT-4 + MCQG-SRefine.

\noindent\textbf{Qualitative analysis}
One of the main issues with the questions generated by GPT-4 is that they often directly include the correct answer or too obvious relevant keywords within the context component (Appendix Table \ref{tab:human_eval_case} Case Studies 1 and 3). 
So the questions directly generated by GPT-4 often make the answers obvious, but GPT-4 + MCQG-SRefine can modify this information into hints for the correct answer through its critical and corrective steps, which experts consider a better way to construct questions for candidates (Case Study 4).

Another finding is that the questions generated by GPT-4 + MCQG-SRefine are more concise compared to those by GPT-4 (Case Study 1). 
Experts pointed out that this conciseness makes them more similar to real USMLE questions.
Our experiments show that GPT-4 adopted a very conservative strategy when generating context due to our emphasis on hallucination issues in the prompts. 
This strategy involves copying and pasting much information from raw medical case inputs to avoid generating potentially incorrect new facts.
Although this does reduce the occurrence of hallucinations - our human assessment shows that 7.5\% of GPT-4 problems contain factual errors - it inevitably sacrifices the typical simplicity and highly refined information presentation of USMLE problems, as well as the logical coherence of multi-hop reasoning between information in context, question, correct answer, and distractors. 
This is a significant stylistic difference between USMLE questions and the original medical cases. 
We also found that prompt engineering was ineffective in resolving this issue. 
We interpret this as a shortcut behavior learned by GPT-4 during aligning with human preferences stage training (e.g., RLHF~\cite{ouyang2022training}) to reduce output diversity to mitigate hallucinations~\cite{kirk2023understanding}.
In contrast, MCQG-SRefine maintains a high level of factual accuracy (with 5\% factual errors) and further improves the quality of USMLE questions by iteratively criticizing and correcting each component of the question, as well as shifting the perspective from QG to QA to make it closer to actual exam questions.
This is another major reason why GPT-4 + MCQG-SRefine outperforms GPT-4 in human evaluation.
However, when MCQG-SRefine reduced the amount of useless information provided in the questions to increase quality and difficulty, omitting too much information sometimes made inferring the correct answer more challenging (Case Study 2), although this was a rare occurrence in our human evaluation (7.5\%).

\section{Discussion}
\label{discussion}
\noindent\textbf{Round-wise analysis}
Since MCQG-SRefine operates as an iterative system, an interesting question is whether GPT-4 can consistently provide meaningful critiques and corrections for itself in such a specialized and complex setting. 
To explore this, we conducted round-wise analyses, with key findings presented in Figures~\ref{fig:Best scoring rounds}, while additional analyses are discussed in the appendix.
Figure~\ref{fig:Best scoring rounds} shows the best-scoring rounds for human- and machine-generated topic + key points. For human-generated topics and key points, 30\% of the best scores came from the first round of output, while the remaining 70\% were evenly distributed across rounds 2, 3, and 4. This suggests that the LLM cannot consistently ensure that its critique and corrections will continuously improve the quality of the generated questions. However, based on the main results, selecting the best round after multiple iterations generally leads to a much higher quality of final questions compared to the initial output.
We observed similar results for machine-generated topics and key points, but the proportion of best scores from the first round was lower (24.9\%). This is consistent with the main results, where topics and key points provided by humans were clearer, making it more likely for the LLM to generate high-quality questions in the first round, while those provided by machines required more improvement.

\begin{table}[!t]
\centering
\scalebox{0.85}{
\begin{tabular}{l|c|c}
\hline
\textbf{LLM-as-Judge* preference} & \textbf{t k (H.)} & \textbf{t k (M.)} \\
\hline
MCQG-SRefine win & 79.80\% & 80.10\% \\ \hline
GPT-4 win & 20.20\% & 19.90\% \\ \hline
\end{tabular}
}
\caption{The win rate of GPT-4 + MCQG-SRefine and GPT-4 questions using LLM-as-Judge (Comparision).}
\vspace{-5mm}
\label{tab:LLM_judge_preference}
\end{table}

\noindent\textbf{Improving LLM-as-judge reliability}
We found that directly using the critique criteria from Section~\ref{Sec:critique} for LLM-as-Judge only resulted in slight reliability when correlated with \texttt{Expert X}. 
We explored two heuristic algorithms to improve the effectiveness of LLM-as-Judge through aspect filtering~\footnote{5 components with 30 aspects in Appendix Table~\ref{tab:demonstration_MCQG-SRefine}.}.
Specifically, we calculated the correlation of each aspect's score from the collected rating feedback with \texttt{Expert X}, then sorted these aspects in descending order based on percentage agreement or Cohen's kappa. 
In the Greedy~\footnote{The Appendix provides the pseudocode for the Greedy Aspect Selection and All-Combo algorithms.}, we added aspects sequentially to the final rating score calculation based on their correlation, from highest to lowest, and recalculated the correlation between \texttt{LLM-as-Judge (rating)} and \texttt{Expert X}.
Similarly, in the All-Combo, we calculated the final rating score for all possible combinations of the top n aspects (1 $<=$ n $<=$ 11)\footnote{We observed diminishing returns for the top 9, 10, and 11 aspects, and thus stopped at 11.}, selecting the combination with the highest correlation to \texttt{Expert X} as the output of the All-Combo algorithm. 
As shown in Appendix Table~\ref{tab:aspect_filtering_all_combination_PA} and Table~\ref{tab:aspect_filtering_all_combination_CK}, the All-Combo method identified the optimal aspect combination.
In Table~\ref{tab:llm-as-judge-corr}, we observed that the percentage agreement and Cohen's kappa of LLM-as-Judge significantly improved~\footnote{According to Cohen's kappa, the reliability improved from slight (0.226) to moderate (0.539).}.

\begin{figure}[!t]
\vspace{-5mm}
    \centering
    \includegraphics[width=\linewidth]{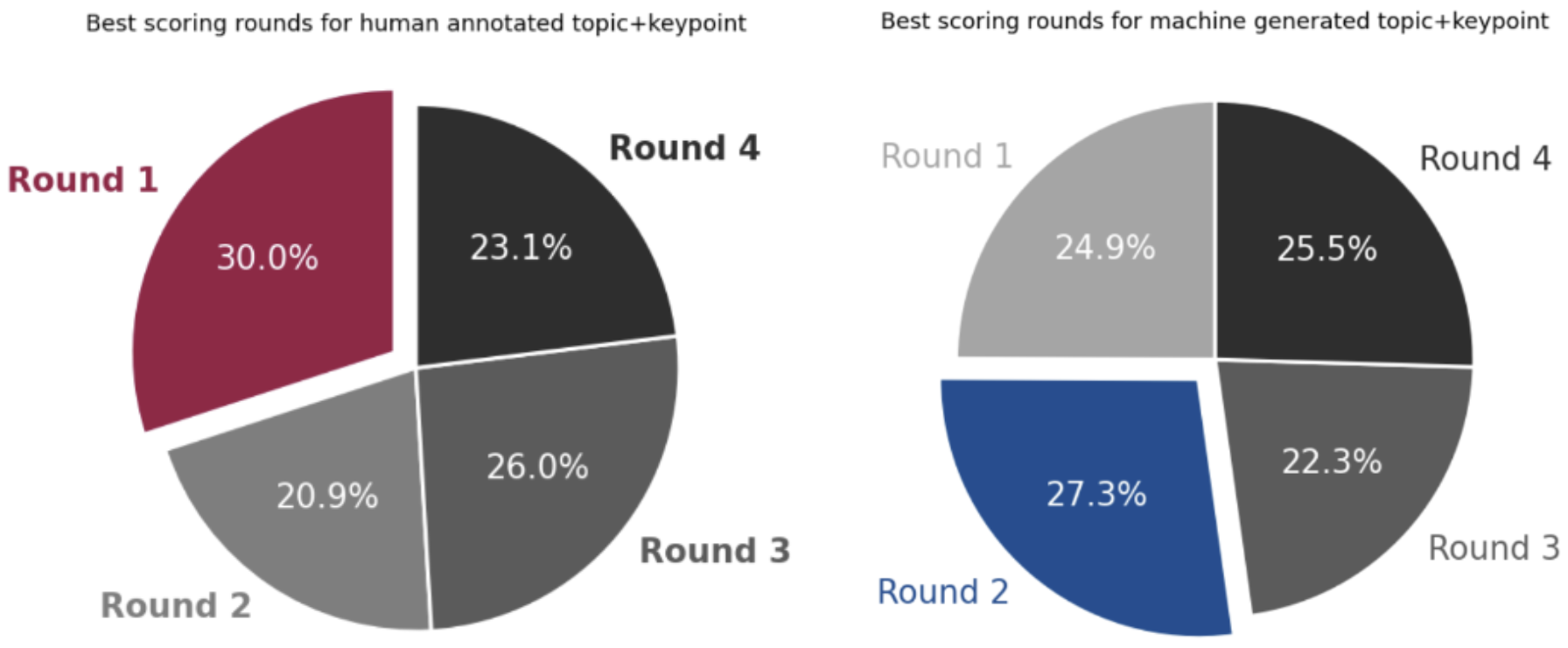}
    \caption{Best scoring rounds.}
    \label{fig:Best scoring rounds}
    \vspace{-3mm}
\end{figure}

\begin{table}[!t]
\centering
\scalebox{0.85}{
\begin{tabular}{l|c|c}
\hline
\textbf{Corr. w/ Expert-x} & \textbf{P. Agreement} & \textbf{C. Kappa} \\ \hline
w/o aspect filtering & 67.5\% & 0.226190476 \\ \hline
PA-based Greedy & 72.5\% & 0.342301943 \\ \hline
CK-based Greedy & 72.5\% & 0.423328965 \\ \hline
PA-based All-Comb & 80.0\% & 0.505409582 \\ \hline
CK-based All-Comb* & 80.0\% & 0.538904899 \\ \hline
\end{tabular}
}
\vspace{-2mm}
\caption{The correlation between the LLM-as-Judge and Expert-X across different aspect filtering methods.}
\vspace{-5mm}
\label{tab:llm-as-judge-corr}
\end{table}

\section{Related Work}
\label{Sec:related work}

\subsection{LLMs for Generating Medical MCQs}

\citet{cheung2023chatgpt} conducted the first study comparing LLMs with humans in medical exam MCQ generation. 
Using GPT-3.5, they generated MCQs from textbooks. GPT produced 50 MCQs in 21 minutes—10\% of the time taken by humans—but human-written questions were of higher quality and effectiveness, outperforming AI in 60\% of acceptable MCQs. Other studies, such as \citet{klang2023advantages}, evaluated AI-generated MCQs using GPT-4 and found them mostly effective, though lacking in clinical context. \citet{agarwal2023analysing} compared GPT-3.5, Bard, and Bing in generating MCQs for a physiology course, noting that GPT produced more effective but less difficult questions. 
Overall, while LLMs can quickly generate MCQs, their quality and difficulty often do not meet expectations. 
We build on this work by qualitatively and quantitatively studying how to convert medical cases into high-quality and challenging USMLE-style MCQs.

\subsection{Self-Refine using LLM Feedback}

Learning from feedback helps align LLMs with desired outcomes, enhancing their ability to follow instructions through different forms of feedback, such as human preference feedback~\cite{ouyang2022training}, AI-generated preference feedback~\cite{lee2023rlaif,dubois2024alpacafarm}, or fine-grained feedback~\cite{wu2024fine,lightman2023let}.
Unlike preference and fine-grained feedback, which provide scalar values as training signals, natural language or correction feedback offers richer information~\cite{scheurer2022training,ma2023eureka,yao2023improving,mishra2024synfac}, making it particularly effective for self-correcting language models~\cite{welleck2022generating,pan2023automatically}.
Recent research has demonstrated that LLMs~\cite{achiam2023gpt,heng2024proggen,wang2024biorag} can self-correct their responses to meet various user requirements, such as reducing harmful content, incorporating specific keywords, diversity requirement generation, or debugging code~\cite{madaan2024self,chen2023teaching}. This self-correction process generally involves generating a critique that identifies shortcomings in the initial response, followed by revising it based on the self-critique—an iterative process that has shown promise in enhancing LLM output quality~\cite{pan2023automatically}.
Inspired by the success of these iterative self-refinement methods, we are the first to explore using this approach for generating USMLE-style MCQs. By leveraging self-feedback, we aim to create high-quality, clinically relevant questions that adhere to the rigorous standards of medical education.

\subsection{LLM-as-Judge using LLM feedback}

The critique capabilities of LLMs have been extensively used for the automatic evaluation of response quality, often employing models like GPT-4~\cite{achiam2023gpt,liu2023gpteval,fu2023gptscore} or critique-adjusted LLMs~\cite{ke2023critiquellm,li2023generative}. Despite their success, these methods have demonstrated instability in certain complex task scenarios~\cite{wang2023shepherd,zhang2024self}.
LLMs have shown a high correlation with human evaluations in tasks such as summarization and story generation, effectively scoring candidate texts or comparing them based on specific evaluation aspects~\cite{chen2023storyer,li2024llms,li2024generation,gu2024survey}. 
For example, studies by \citet{chiang2023can} and \citet{kocmi2023large} have shown that LLM evaluations produce results comparable to those of expert human evaluators in story generation and translation tasks. Similarly, research by \citet{zheng2024judging} and \citet{zhang2024comprehensive} indicates that powerful LLM reviewers, such as GPT-4, achieve over 80\% consistency with human preferences in multi-turn dialogue scenarios, both in controlled and crowdsourced settings, reaching agreement levels similar to human evaluators.
Further evidence of GPT-4's effectiveness as an evaluator is demonstrated by its performance in the PROMETHEUS~\cite{kim2023prometheus,kim2024prometheus} and CRITICBENCH~\cite{lan2024criticbench} benchmarks. 
However, most LLM-as-judge research has primarily focused on general NLP fields, with limited exploration in specialized domains like clinical NLP. This gap is largely due to challenges such as the need for domain-specific knowledge, difficulties in designing evaluation prompts that meet domain standards, and the inherent challenges of generalizing evaluation tools to specialized fields~\cite{singhal2023large,li2024leveraging}.

In the medical field~\cite{ li2024llms,raju2024constructing}, LLMs-as-judges have demonstrated certain potential in areas such as diagnostic support~\cite{yao2024medqa}, medical documentation~\cite{brake2024comparing,brake2024comparing}, clinical conversation~\cite{wang2023notechat, li2024automatic}, medical question answering~\cite{wang2024jmlr,krolik2024towards}, medical reasoning~\cite{jeong2024improving}, and patient education~\cite{yao2023readme}.
To the best of our knowledge, we are the first to explore the application of LLM-as-judge in the context of USMLE-MCQ generation and evaluation. Our work addresses these challenges to create reliable, domain-specific evaluations that meet the unique requirements of medical education.

\section{Conclusion}

In conclusion, MCQG-SRefine improves LLM ability to generate high-quality USMLE-style MCQs by integrating expert-driven prompts with iterative self-refinement. The framework significantly enhances question quality and difficulty while aligning closely with expert evaluations. Additionally, our \texttt{LLM-as-Judge} metric offers a scalable alternative to costly expert assessments.


\section{Limitations and Ethical Considerations}

Our research presents promising advances in the generation of medical exam questions. However, several limitations and ethical considerations require further exploration to ensure the robustness, fairness, and social value of this work.

\noindent\textbf{Adaptability to Other Domains and Languages:}
Our study focuses on medical education, specifically the USMLE format. Although the methods and frameworks show potential, their generalizability to other domains, languages, or question types remains unexplored. Future research should validate the adaptability of these methods to broader educational contexts and linguistic variations.

\noindent\textbf{Evaluation Bias and Reliability of LLM-as-Judge:}
The LLM-as-Judge approach, where GPT-4 is used for both generating and evaluating output, carries a significant risk of self-evaluation bias. 
First, the evaluation and generation abilities are not equivalent. According to higher-order thinking theories~\footnote{\url{https://en.wikipedia.org/wiki/Higher-order_thinking}} in educational research, evaluation is a cognitively higher-dimensional task compared to task completion. Recent studies~\cite{west2023generative} also reveal that generative AI models can excel at creating content while lacking deep understanding, which limits their reliability as evaluators. 
Second, while GPT-4 was selected due to its relatively smaller position bias compared to other models such as Claude3-Opus, Sonnet, Haiku, and GPT-3.5 (as shown in Appendix Table~\ref{tab:llm_as_judge_comparison_full}), it still exhibits inconsistencies in approximately 35\% of cases. These inconsistencies, where GPT-4's judgments vary based on the order of presented options (A-B vs. B-A), highlight the challenges of relying on LLMs for evaluation. 
Third, inconsistencies across different critique dimensions (e.g., individual scoring vs. pairwise comparison) further compound these challenges, which is aligned with recent work findings~\cite{lan2024criticbench}.
Addressing these limitations will require novel approaches to improve the reliability and fairness of LLM-based evaluation frameworks.

\noindent\textbf{Potential for Over-Self-Critique:}
Our iterative self-refinement method (MCQG-SRefine) improves question quality and difficulty based on critique and correction stages.
However, there is a risk of over-self-critique, where excessive iterations lead to suboptimal or overly complex outputs. While our Round-wise Analysis (Figure~\ref{fig:Best scoring rounds}) and Appendix~\ref{apx:ablation_study} highlight how question quality evolves across iterations, we observed cases where later refinements did not outperform earlier ones. This underscores the need for careful calibration of self-refinement processes to strike a balance between improvement and diminishing returns. Future work should investigate mechanisms to mitigate over-self-critique, such as introducing external evaluation checkpoints or dynamic stopping criteria.

\noindent\textbf{Privacy and Ethical Considerations:}
The use of medical data to generate questions raises critical privacy concerns. Although our work uses publicly available clinical resources as input, and we adhere to deidentification protocols, ensuring compliance with ethical standards and safeguarding patient privacy is important. Moreover, fairness in question generation remains an open challenge. Biases present in the LLM original training data as well as input clinical notes, such as underrepresentation of certain medical conditions or demographic groups, can lead to biased outputs. Future research must prioritize fairness-aware techniques to mitigate these issues and ensure equity in medical education tools.

\noindent\textbf{Impact on Medical Education and Educator-Student Dynamics:}
While automated USMLE-MCQG systems hold the potential to boost the efficiency, scalability, and accuracy of medical education, they may inadvertently reduce direct student-educator interactions. Medical educators are essential in providing context, clinical reasoning insights, and mentorship that automated systems cannot fully replicate. Over-reliance on automated tools could undermine these critical learning experiences. Moreover, hallucinations or inaccuracies generated by LLMs could misinform learners, posing risks to both education quality and eventual clinical practice. It is, therefore, crucial to position these tools as assistive systems that support, rather than replace, educators.

\noindent\textbf{Broader Societal and Ethical Implications:}
The societal impacts of (semi)-automated MCQG systems extend beyond education. Ensuring that these technologies are accessible, fair, and transparent is vital to prevent exacerbating educational inequities. Moreover, the automation of question generation should be accompanied by rigorous human oversight to identify and correct potential errors. Continued collaboration with medical professionals, educators, and ethicists will be critical to addressing these challenges and ensuring the responsible deployment of AI-driven educational tools.


\bibliography{custom}

\begin{thebibliography}{81}
\providecommand{\natexlab}[1]{#1}

\bibitem[{Achiam et~al.(2023)Achiam, Adler, Agarwal, Ahmad, Akkaya, Aleman, Almeida, Altenschmidt, Altman, Anadkat et~al.}]{achiam2023gpt}
Josh Achiam, Steven Adler, Sandhini Agarwal, Lama Ahmad, Ilge Akkaya, Florencia~Leoni Aleman, Diogo Almeida, Janko Altenschmidt, Sam Altman, Shyamal Anadkat, et~al. 2023.
\newblock Gpt-4 technical report.
\newblock \emph{arXiv preprint arXiv:2303.08774}.

\bibitem[{Agarwal et~al.(2023)Agarwal, Sharma, and Goswami}]{agarwal2023analysing}
Mayank Agarwal, Priyanka Sharma, and Ayan Goswami. 2023.
\newblock Analysing the applicability of chatgpt, bard, and bing to generate reasoning-based multiple-choice questions in medical physiology.
\newblock \emph{Cureus}, 15(6).

\bibitem[{Artsi et~al.(2024)Artsi, Sorin, Konen, Glicksberg, Nadkarni, and Klang}]{artsi2024large}
Yaara Artsi, Vera Sorin, Eli Konen, Benjamin~S Glicksberg, Girish Nadkarni, and Eyal Klang. 2024.
\newblock Large language models for generating medical examinations: systematic review.
\newblock \emph{BMC Medical Education}, 24(1):354.

\bibitem[{Ben{\'\i}tez et~al.(2024)Ben{\'\i}tez, Xu, Boudreau, Kow, Bello, Van~Phuoc, Wang, Sun, Leung, Lan et~al.}]{benitez2024harnessing}
Trista~M Ben{\'\i}tez, Yueyuan Xu, J~Donald Boudreau, Alfred Wei~Chieh Kow, Fernando Bello, Le~Van~Phuoc, Xiaofei Wang, Xiaodong Sun, Gilberto Ka-Kit Leung, Yanyan Lan, et~al. 2024.
\newblock Harnessing the potential of large language models in medical education: promise and pitfalls.
\newblock \emph{Journal of the American Medical Informatics Association}, 31(3):776--783.

\bibitem[{Biswas(2023)}]{biswas2023passing}
Som Biswas. 2023.
\newblock Passing is great: Can chatgpt conduct usmle exams?
\newblock \emph{Annals of Biomedical Engineering}, 51(9):1885--1886.

\bibitem[{Brake and Schaaf(2024)}]{brake2024comparing}
Nathan Brake and Thomas Schaaf. 2024.
\newblock Comparing two model designs for clinical note generation; is an llm a useful evaluator of consistency?
\newblock \emph{arXiv preprint arXiv:2404.06503}.

\bibitem[{Cai et~al.(2023)Cai, Yao, Liu, Wang, Reilly, Zhou, Li, Cao, Kapoor, Bajracharya et~al.}]{cai2023paniniqa}
Pengshan Cai, Zonghai Yao, Fei Liu, Dakuo Wang, Meghan Reilly, Huixue Zhou, Lingxi Li, Yi~Cao, Alok Kapoor, Adarsha Bajracharya, et~al. 2023.
\newblock Paniniqa: Enhancing patient education through interactive question answering.
\newblock \emph{Transactions of the Association for Computational Linguistics}, 11:1518--1536.

\bibitem[{Ch and Saha(2018)}]{ch2018automatic}
Dhawaleswar~Rao Ch and Sujan~Kumar Saha. 2018.
\newblock Automatic multiple choice question generation from text: A survey.
\newblock \emph{IEEE Transactions on Learning Technologies}, 13(1):14--25.

\bibitem[{Chen et~al.(2023{\natexlab{a}})Chen, Vo, Takamura, Miyao, and Nakayama}]{chen2023storyer}
Hong Chen, Duc~Minh Vo, Hiroya Takamura, Yusuke Miyao, and Hideki Nakayama. 2023{\natexlab{a}}.
\newblock Storyer: Automatic story evaluation via ranking, rating and reasoning.
\newblock \emph{Journal of Natural Language Processing}, 30(1):243--249.

\bibitem[{Chen et~al.(2023{\natexlab{b}})Chen, Lin, Sch{\"a}rli, and Zhou}]{chen2023teaching}
Xinyun Chen, Maxwell Lin, Nathanael Sch{\"a}rli, and Denny Zhou. 2023{\natexlab{b}}.
\newblock Teaching large language models to self-debug.
\newblock \emph{arXiv preprint arXiv:2304.05128}.

\bibitem[{Chen et~al.(2023{\natexlab{c}})Chen, Wu, and Zaki}]{chen2023toward}
Yu~Chen, Lingfei Wu, and Mohammed~J Zaki. 2023{\natexlab{c}}.
\newblock Toward subgraph-guided knowledge graph question generation with graph neural networks.
\newblock \emph{IEEE Transactions on Neural Networks and Learning Systems}.

\bibitem[{Cheung et~al.(2023)Cheung, Lau, Wong, Lee, Kulkarni, Seow, Wong, and Co}]{cheung2023chatgpt}
Billy Ho~Hung Cheung, Gary Kui~Kai Lau, Gordon Tin~Chun Wong, Elaine Yuen~Phin Lee, Dhananjay Kulkarni, Choon~Sheong Seow, Ruby Wong, and Michael Tiong-Hong Co. 2023.
\newblock Chatgpt versus human in generating medical graduate exam multiple choice questions—a multinational prospective study (hong kong sar, singapore, ireland, and the united kingdom).
\newblock \emph{PLoS One}, 18(8):e0290691.

\bibitem[{Chiang and Lee(2023)}]{chiang2023can}
Cheng-Han Chiang and Hung-yi Lee. 2023.
\newblock Can large language models be an alternative to human evaluations?
\newblock \emph{arXiv preprint arXiv:2305.01937}.

\bibitem[{Dubois et~al.(2024)Dubois, Li, Taori, Zhang, Gulrajani, Ba, Guestrin, Liang, and Hashimoto}]{dubois2024alpacafarm}
Yann Dubois, Chen~Xuechen Li, Rohan Taori, Tianyi Zhang, Ishaan Gulrajani, Jimmy Ba, Carlos Guestrin, Percy~S Liang, and Tatsunori~B Hashimoto. 2024.
\newblock Alpacafarm: A simulation framework for methods that learn from human feedback.
\newblock \emph{Advances in Neural Information Processing Systems}, 36.

\bibitem[{Fu et~al.(2023)Fu, Ng, Jiang, and Liu}]{fu2023gptscore}
Jinlan Fu, See-Kiong Ng, Zhengbao Jiang, and Pengfei Liu. 2023.
\newblock Gptscore: Evaluate as you desire.
\newblock \emph{arXiv preprint arXiv:2302.04166}.

\bibitem[{Gierl et~al.(2012)Gierl, Lai, and Turner}]{gierl2012using}
Mark~J Gierl, Hollis Lai, and Simon~R Turner. 2012.
\newblock Using automatic item generation to create multiple-choice test items.
\newblock \emph{Medical education}, 46(8):757--765.

\bibitem[{Gilardi et~al.(2023)Gilardi, Alizadeh, and Kubli}]{gilardi2023chatgpt}
Fabrizio Gilardi, Meysam Alizadeh, and Ma{\"e}l Kubli. 2023.
\newblock Chatgpt outperforms crowd workers for text-annotation tasks.
\newblock \emph{Proceedings of the National Academy of Sciences}, 120(30):e2305016120.

\bibitem[{Gu et~al.(2024)Gu, Jiang, Shi, Tan, Zhai, Xu, Li, Shen, Ma, Liu et~al.}]{gu2024survey}
Jiawei Gu, Xuhui Jiang, Zhichao Shi, Hexiang Tan, Xuehao Zhai, Chengjin Xu, Wei Li, Yinghan Shen, Shengjie Ma, Honghao Liu, et~al. 2024.
\newblock A survey on llm-as-a-judge.
\newblock \emph{arXiv preprint arXiv:2411.15594}.

\bibitem[{Guo et~al.(2024)Guo, Liao, Li, and Chua}]{guo2024survey}
Shasha Guo, Lizi Liao, Cuiping Li, and Tat-Seng Chua. 2024.
\newblock A survey on neural question generation: Methods, applications, and prospects.
\newblock \emph{arXiv preprint arXiv:2402.18267}.

\bibitem[{Guo et~al.(2022)Guo, Zhang, Wang, Zhang, Li, and Chen}]{guo2022dsm}
Shasha Guo, Jing Zhang, Yanling Wang, Qianyi Zhang, Cuiping Li, and Hong Chen. 2022.
\newblock Dsm: Question generation over knowledge base via modeling diverse subgraphs with meta-learner.
\newblock In \emph{Proceedings of the 2022 Conference on Empirical Methods in Natural Language Processing}, pages 4194--4207.

\bibitem[{Heng et~al.(2024)Heng, Deng, Li, Yu, Li, Zhang, and Zhang}]{heng2024proggen}
Yuzhao Heng, Chunyuan Deng, Yitong Li, Yue Yu, Yinghao Li, Rongzhi Zhang, and Chao Zhang. 2024.
\newblock Proggen: Generating named entity recognition datasets step-by-step with self-reflexive large language models.
\newblock \emph{arXiv preprint arXiv:2403.11103}.

\bibitem[{Homolak(2023)}]{homolak2023opportunities}
Jan Homolak. 2023.
\newblock Opportunities and risks of chatgpt in medicine, science, and academic publishing: a modern promethean dilemma.
\newblock \emph{Croatian Medical Journal}, 64(1):1.

\bibitem[{Jeong et~al.(2024)Jeong, Sohn, Sung, and Kang}]{jeong2024improving}
Minbyul Jeong, Jiwoong Sohn, Mujeen Sung, and Jaewoo Kang. 2024.
\newblock Improving medical reasoning through retrieval and self-reflection with retrieval-augmented large language models.
\newblock \emph{Bioinformatics}, 40(Supplement\_1):i119--i129.

\bibitem[{Jin et~al.(2021)Jin, Pan, Oufattole, Weng, Fang, and Szolovits}]{jin2021disease}
Di~Jin, Eileen Pan, Nassim Oufattole, Wei-Hung Weng, Hanyi Fang, and Peter Szolovits. 2021.
\newblock What disease does this patient have? a large-scale open domain question answering dataset from medical exams.
\newblock \emph{Applied Sciences}, 11(14):6421.

\bibitem[{Ke et~al.(2023)Ke, Wen, Feng, Liu, Lei, Cheng, Wang, Zeng, Dong, Wang et~al.}]{ke2023critiquellm}
Pei Ke, Bosi Wen, Zhuoer Feng, Xiao Liu, Xuanyu Lei, Jiale Cheng, Shengyuan Wang, Aohan Zeng, Yuxiao Dong, Hongning Wang, et~al. 2023.
\newblock Critiquellm: Scaling llm-as-critic for effective and explainable evaluation of large language model generation.
\newblock \emph{arXiv preprint arXiv:2311.18702}.

\bibitem[{Kim et~al.(2023)Kim, Shin, Cho, Jang, Longpre, Lee, Yun, Shin, Kim, Thorne et~al.}]{kim2023prometheus}
Seungone Kim, Jamin Shin, Yejin Cho, Joel Jang, Shayne Longpre, Hwaran Lee, Sangdoo Yun, Seongjin Shin, Sungdong Kim, James Thorne, et~al. 2023.
\newblock Prometheus: Inducing fine-grained evaluation capability in language models.
\newblock \emph{arXiv preprint arXiv:2310.08491}.

\bibitem[{Kim et~al.(2024)Kim, Suk, Longpre, Lin, Shin, Welleck, Neubig, Lee, Lee, and Seo}]{kim2024prometheus}
Seungone Kim, Juyoung Suk, Shayne Longpre, Bill~Yuchen Lin, Jamin Shin, Sean Welleck, Graham Neubig, Moontae Lee, Kyungjae Lee, and Minjoon Seo. 2024.
\newblock Prometheus 2: An open source language model specialized in evaluating other language models.
\newblock \emph{arXiv preprint arXiv:2405.01535}.

\bibitem[{Kirk et~al.(2023)Kirk, Mediratta, Nalmpantis, Luketina, Hambro, Grefenstette, and Raileanu}]{kirk2023understanding}
Robert Kirk, Ishita Mediratta, Christoforos Nalmpantis, Jelena Luketina, Eric Hambro, Edward Grefenstette, and Roberta Raileanu. 2023.
\newblock Understanding the effects of rlhf on llm generalisation and diversity.
\newblock \emph{arXiv preprint arXiv:2310.06452}.

\bibitem[{Klang et~al.(2023)Klang, Portugez, Gross, Brenner, Gilboa, Ortal, Ron, Robinzon, Meiri, Segal et~al.}]{klang2023advantages}
E~Klang, S~Portugez, R~Gross, A~Brenner, M~Gilboa, T~Ortal, S~Ron, V~Robinzon, H~Meiri, G~Segal, et~al. 2023.
\newblock Advantages and pitfalls in utilizing artificial intelligence for crafting medical examinations: a medical education pilot study with gpt-4.
\newblock \emph{BMC Medical Education}, 23.

\bibitem[{Kocmi and Federmann(2023)}]{kocmi2023large}
Tom Kocmi and Christian Federmann. 2023.
\newblock Large language models are state-of-the-art evaluators of translation quality.
\newblock \emph{arXiv preprint arXiv:2302.14520}.

\bibitem[{Krolik et~al.(2024)Krolik, Mahal, Ahmad, Trivedi, and Saket}]{krolik2024towards}
Jack Krolik, Herprit Mahal, Feroz Ahmad, Gaurav Trivedi, and Bahador Saket. 2024.
\newblock Towards leveraging large language models for automated medical q\&a evaluation.
\newblock \emph{arXiv preprint arXiv:2409.01941}.

\bibitem[{Lan et~al.(2024)Lan, Zhang, Xu, Huang, Lin, Chen, and Mao}]{lan2024criticbench}
Tian Lan, Wenwei Zhang, Chen Xu, Heyan Huang, Dahua Lin, Kai Chen, and Xian-ling Mao. 2024.
\newblock Criticbench: Evaluating large language models as critic.
\newblock \emph{arXiv preprint arXiv:2402.13764}.

\bibitem[{Lee et~al.(2023)Lee, Phatale, Mansoor, Lu, Mesnard, Bishop, Carbune, and Rastogi}]{lee2023rlaif}
Harrison Lee, Samrat Phatale, Hassan Mansoor, Kellie Lu, Thomas Mesnard, Colton Bishop, Victor Carbune, and Abhinav Rastogi. 2023.
\newblock Rlaif: Scaling reinforcement learning from human feedback with ai feedback.
\newblock \emph{arXiv preprint arXiv:2309.00267}.

\bibitem[{Li et~al.(2024{\natexlab{a}})Li, Lu, Song, Zhang, Ma, and Lan}]{li2024automatic}
Anqi Li, Yu~Lu, Nirui Song, Shuai Zhang, Lizhi Ma, and Zhenzhong Lan. 2024{\natexlab{a}}.
\newblock Automatic evaluation for mental health counseling using llms.
\newblock \emph{arXiv preprint arXiv:2402.11958}.

\bibitem[{Li et~al.(2024{\natexlab{b}})Li, Jiang, Huang, Beigi, Zhao, Tan, Bhattacharjee, Jiang, Chen, Wu et~al.}]{li2024generation}
Dawei Li, Bohan Jiang, Liangjie Huang, Alimohammad Beigi, Chengshuai Zhao, Zhen Tan, Amrita Bhattacharjee, Yuxuan Jiang, Canyu Chen, Tianhao Wu, et~al. 2024{\natexlab{b}}.
\newblock From generation to judgment: Opportunities and challenges of llm-as-a-judge.
\newblock \emph{arXiv preprint arXiv:2411.16594}.

\bibitem[{Li et~al.(2024{\natexlab{c}})Li, Dong, Chen, Su, Zhou, Ai, Ye, and Liu}]{li2024llms}
Haitao Li, Qian Dong, Junjie Chen, Huixue Su, Yujia Zhou, Qingyao Ai, Ziyi Ye, and Yiqun Liu. 2024{\natexlab{c}}.
\newblock Llms-as-judges: A comprehensive survey on llm-based evaluation methods.
\newblock \emph{arXiv preprint arXiv:2412.05579}.

\bibitem[{Li et~al.(2023)Li, Sun, Yuan, Fan, Zhao, and Liu}]{li2023generative}
Junlong Li, Shichao Sun, Weizhe Yuan, Run-Ze Fan, Hai Zhao, and Pengfei Liu. 2023.
\newblock Generative judge for evaluating alignment.
\newblock \emph{arXiv preprint arXiv:2310.05470}.

\bibitem[{Li et~al.(2024{\natexlab{d}})Li, Xu, Shen, Xu, Gu, and Tao}]{li2024leveraging}
Zhen Li, Xiaohan Xu, Tao Shen, Can Xu, Jia-Chen Gu, and Chongyang Tao. 2024{\natexlab{d}}.
\newblock Leveraging large language models for nlg evaluation: A survey.
\newblock \emph{arXiv preprint arXiv:2401.07103}.

\bibitem[{Lightman et~al.(2023)Lightman, Kosaraju, Burda, Edwards, Baker, Lee, Leike, Schulman, Sutskever, and Cobbe}]{lightman2023let}
Hunter Lightman, Vineet Kosaraju, Yura Burda, Harri Edwards, Bowen Baker, Teddy Lee, Jan Leike, John Schulman, Ilya Sutskever, and Karl Cobbe. 2023.
\newblock Let's verify step by step.
\newblock \emph{arXiv preprint arXiv:2305.20050}.

\bibitem[{Liu et~al.(2023)Liu, Iter, Xu, Wang, Xu, and Zhu}]{liu2023gpteval}
Yang Liu, Dan Iter, Yichong Xu, Shuohang Wang, Ruochen Xu, and Chenguang Zhu. 2023.
\newblock Gpteval: Nlg evaluation using gpt-4 with better human alignment.
\newblock \emph{arXiv preprint arXiv:2303.16634}.

\bibitem[{Ma et~al.(2023)Ma, Liang, Wang, Huang, Bastani, Jayaraman, Zhu, Fan, and Anandkumar}]{ma2023eureka}
Yecheng~Jason Ma, William Liang, Guanzhi Wang, De-An Huang, Osbert Bastani, Dinesh Jayaraman, Yuke Zhu, Linxi Fan, and Anima Anandkumar. 2023.
\newblock Eureka: Human-level reward design via coding large language models.
\newblock \emph{arXiv preprint arXiv:2310.12931}.

\bibitem[{Madaan et~al.(2024)Madaan, Tandon, Gupta, Hallinan, Gao, Wiegreffe, Alon, Dziri, Prabhumoye, Yang et~al.}]{madaan2024self}
Aman Madaan, Niket Tandon, Prakhar Gupta, Skyler Hallinan, Luyu Gao, Sarah Wiegreffe, Uri Alon, Nouha Dziri, Shrimai Prabhumoye, Yiming Yang, et~al. 2024.
\newblock Self-refine: Iterative refinement with self-feedback.
\newblock \emph{Advances in Neural Information Processing Systems}, 36.

\bibitem[{Mehandru et~al.(2024)Mehandru, Miao, Almaraz, Sushil, Butte, and Alaa}]{mehandru2024evaluating}
Nikita Mehandru, Brenda~Y Miao, Eduardo~Rodriguez Almaraz, Madhumita Sushil, Atul~J Butte, and Ahmed Alaa. 2024.
\newblock Evaluating large language models as agents in the clinic.
\newblock \emph{npj Digital Medicine}, 7(1):84.

\bibitem[{Mishra et~al.(2024)Mishra, Yao, Vashisht, Ouyang, Wang, Mody, and Yu}]{mishra2024synfac}
Prakamya Mishra, Zonghai Yao, Parth Vashisht, Feiyun Ouyang, Beining Wang, Vidhi~Dhaval Mody, and Hong Yu. 2024.
\newblock Synfac-edit: Synthetic imitation edit feedback for factual alignment in clinical summarization.
\newblock \emph{arXiv preprint arXiv:2402.13919}.

\bibitem[{Mousavi et~al.(2024)Mousavi, Alghisi, and Riccardi}]{mousavi2024your}
Seyed~Mahed Mousavi, Simone Alghisi, and Giuseppe Riccardi. 2024.
\newblock Is your llm outdated? benchmarking llms \& alignment algorithms for time-sensitive knowledge.
\newblock \emph{arXiv preprint arXiv:2404.08700}.

\bibitem[{Ouyang et~al.(2022)Ouyang, Wu, Jiang, Almeida, Wainwright, Mishkin, Zhang, Agarwal, Slama, Ray et~al.}]{ouyang2022training}
Long Ouyang, Jeffrey Wu, Xu~Jiang, Diogo Almeida, Carroll Wainwright, Pamela Mishkin, Chong Zhang, Sandhini Agarwal, Katarina Slama, Alex Ray, et~al. 2022.
\newblock Training language models to follow instructions with human feedback.
\newblock \emph{Advances in neural information processing systems}, 35:27730--27744.

\bibitem[{Pan et~al.(2021)Pan, Chen, Xiong, Kan, and Wang}]{pan2021zero}
Liangming Pan, Wenhu Chen, Wenhan Xiong, Min-Yen Kan, and William~Yang Wang. 2021.
\newblock Zero-shot fact verification by claim generation.
\newblock \emph{arXiv preprint arXiv:2105.14682}.

\bibitem[{Pan et~al.(2023)Pan, Saxon, Xu, Nathani, Wang, and Wang}]{pan2023automatically}
Liangming Pan, Michael Saxon, Wenda Xu, Deepak Nathani, Xinyi Wang, and William~Yang Wang. 2023.
\newblock Automatically correcting large language models: Surveying the landscape of diverse self-correction strategies.
\newblock \emph{arXiv preprint arXiv:2308.03188}.

\bibitem[{Philip A~Bucur(2019)}]{usmle_prepare_cost}
Sebastian R~Diaz Philip A~Bucur, Vikrant~Bhatnagar. 2019.
\newblock \href {https://www.ncbi.nlm.nih.gov/pmc/articles/PMC6822909/} {A “u-shaped" curve: Appreciating how primary care residency intention relates to the cost of board preparation and examination}.
\newblock \emph{cureus}.

\bibitem[{Raju et~al.(2024)Raju, Jain, Li, Li, and Thakker}]{raju2024constructing}
Ravi Raju, Swayambhoo Jain, Bo~Li, Jonathan Li, and Urmish Thakker. 2024.
\newblock Constructing domain-specific evaluation sets for llm-as-a-judge.
\newblock \emph{arXiv preprint arXiv:2408.08808}.

\bibitem[{Sahoo et~al.(2024)Sahoo, Singh, Saha, Jain, Mondal, and Chadha}]{sahoo2024systematic}
Pranab Sahoo, Ayush~Kumar Singh, Sriparna Saha, Vinija Jain, Samrat Mondal, and Aman Chadha. 2024.
\newblock A systematic survey of prompt engineering in large language models: Techniques and applications.
\newblock \emph{arXiv preprint arXiv:2402.07927}.

\bibitem[{Santhanam et~al.(2021)Santhanam, Khattab, Saad-Falcon, Potts, and Zaharia}]{santhanam2021colbertv2}
Keshav Santhanam, Omar Khattab, Jon Saad-Falcon, Christopher Potts, and Matei Zaharia. 2021.
\newblock Colbertv2: Effective and efficient retrieval via lightweight late interaction.
\newblock \emph{arXiv preprint arXiv:2112.01488}.

\bibitem[{Scheurer et~al.(2022)Scheurer, Campos, Chan, Chen, Cho, and Perez}]{scheurer2022training}
J{\'e}r{\'e}my Scheurer, Jon~Ander Campos, Jun~Shern Chan, Angelica Chen, Kyunghyun Cho, and Ethan Perez. 2022.
\newblock Training language models with language feedback.
\newblock \emph{arXiv preprint arXiv:2204.14146}.

\bibitem[{Schmidgall et~al.(2024)Schmidgall, Ziaei, Harris, Reis, Jopling, and Moor}]{schmidgall2024agentclinic}
Samuel Schmidgall, Rojin Ziaei, Carl Harris, Eduardo Reis, Jeffrey Jopling, and Michael Moor. 2024.
\newblock Agentclinic: a multimodal agent benchmark to evaluate ai in simulated clinical environments.
\newblock \emph{arXiv preprint arXiv:2405.07960}.

\bibitem[{Scoles(2008)}]{usmle_summary}
Peter~V Scoles. 2008.
\newblock Comprehensive review of the usmle.
\newblock \emph{Advances in Physiology Education}, 32(2):109--110.

\bibitem[{Singhal et~al.(2023)Singhal, Azizi, Tu, Mahdavi, Wei, Chung, Scales, Tanwani, Cole-Lewis, Pfohl et~al.}]{singhal2023large}
Karan Singhal, Shekoofeh Azizi, Tao Tu, S~Sara Mahdavi, Jason Wei, Hyung~Won Chung, Nathan Scales, Ajay Tanwani, Heather Cole-Lewis, Stephen Pfohl, et~al. 2023.
\newblock Large language models encode clinical knowledge.
\newblock \emph{Nature}, 620(7972):172--180.

\bibitem[{Su et~al.(2022)Su, Xu, and Fung}]{su2022qa4qg}
Dan Su, Peng Xu, and Pascale Fung. 2022.
\newblock Qa4qg: using question answering to constrain multi-hop question generation.
\newblock In \emph{ICASSP 2022-2022 IEEE International Conference on Acoustics, Speech and Signal Processing (ICASSP)}, pages 8232--8236. IEEE.

\bibitem[{Sun et~al.(2019)Sun, Tang, Duan, Qin, Liu, Yan, Zhou, Lv, Yin, Feng et~al.}]{sun2019joint}
Yibo Sun, Duyu Tang, Nan Duan, Tao Qin, Shujie Liu, Zhao Yan, Ming Zhou, Yuanhua Lv, Wenpeng Yin, Xiaocheng Feng, et~al. 2019.
\newblock Joint learning of question answering and question generation.
\newblock \emph{IEEE Transactions on Knowledge and Data Engineering}, 32(5):971--982.

\bibitem[{Wang et~al.(2024{\natexlab{a}})Wang, Long, Xiao, Cai, Wu, Meng, Wang, and Zhou}]{wang2024biorag}
Chengrui Wang, Qingqing Long, Meng Xiao, Xunxin Cai, Chengjun Wu, Zhen Meng, Xuezhi Wang, and Yuanchun Zhou. 2024{\natexlab{a}}.
\newblock Biorag: A rag-llm framework for biological question reasoning.
\newblock \emph{arXiv preprint arXiv:2408.01107}.

\bibitem[{Wang et~al.(2024{\natexlab{b}})Wang, Yang, Yao, and Yu}]{wang2024jmlr}
Junda Wang, Zhichao Yang, Zonghai Yao, and Hong Yu. 2024{\natexlab{b}}.
\newblock Jmlr: Joint medical llm and retrieval training for enhancing reasoning and professional question answering capability.
\newblock \emph{arXiv preprint arXiv:2402.17887}.

\bibitem[{Wang et~al.(2023{\natexlab{a}})Wang, Yao, Yang, Zhou, Li, Wang, Xu, and Yu}]{wang2023notechat}
Junda Wang, Zonghai Yao, Zhichao Yang, Huixue Zhou, Rumeng Li, Xun Wang, Yucheng Xu, and Hong Yu. 2023{\natexlab{a}}.
\newblock Notechat: a dataset of synthetic doctor-patient conversations conditioned on clinical notes.
\newblock \emph{arXiv preprint arXiv:2310.15959}.

\bibitem[{Wang et~al.(2023{\natexlab{b}})Wang, Li, Chen, Cai, Zhu, Lin, Cao, Liu, Liu, and Sui}]{wang2023large}
Peiyi Wang, Lei Li, Liang Chen, Zefan Cai, Dawei Zhu, Binghuai Lin, Yunbo Cao, Qi~Liu, Tianyu Liu, and Zhifang Sui. 2023{\natexlab{b}}.
\newblock Large language models are not fair evaluators.
\newblock \emph{arXiv preprint arXiv:2305.17926}.

\bibitem[{Wang et~al.(2023{\natexlab{c}})Wang, Yu, Tan, O'Brien, Pasunuru, Dwivedi-Yu, Golovneva, Zettlemoyer, Fazel-Zarandi, and Celikyilmaz}]{wang2023shepherd}
Tianlu Wang, Ping Yu, Xiaoqing~Ellen Tan, Sean O'Brien, Ramakanth Pasunuru, Jane Dwivedi-Yu, Olga Golovneva, Luke Zettlemoyer, Maryam Fazel-Zarandi, and Asli Celikyilmaz. 2023{\natexlab{c}}.
\newblock Shepherd: A critic for language model generation.
\newblock \emph{arXiv preprint arXiv:2308.04592}.

\bibitem[{Wang et~al.(2023{\natexlab{d}})Wang, Ivison, Dasigi, Hessel, Khot, Chandu, Wadden, MacMillan, Smith, Beltagy et~al.}]{wang2023far}
Yizhong Wang, Hamish Ivison, Pradeep Dasigi, Jack Hessel, Tushar Khot, Khyathi Chandu, David Wadden, Kelsey MacMillan, Noah~A Smith, Iz~Beltagy, et~al. 2023{\natexlab{d}}.
\newblock How far can camels go? exploring the state of instruction tuning on open resources.
\newblock \emph{Advances in Neural Information Processing Systems}, 36:74764--74786.

\bibitem[{Welleck et~al.(2022)Welleck, Lu, West, Brahman, Shen, Khashabi, and Choi}]{welleck2022generating}
Sean Welleck, Ximing Lu, Peter West, Faeze Brahman, Tianxiao Shen, Daniel Khashabi, and Yejin Choi. 2022.
\newblock Generating sequences by learning to self-correct.
\newblock \emph{arXiv preprint arXiv:2211.00053}.

\bibitem[{West et~al.(2023)West, Lu, Dziri, Brahman, Li, Hwang, Jiang, Fisher, Ravichander, Chandu et~al.}]{west2023generative}
Peter West, Ximing Lu, Nouha Dziri, Faeze Brahman, Linjie Li, Jena~D Hwang, Liwei Jiang, Jillian Fisher, Abhilasha Ravichander, Khyathi Chandu, et~al. 2023.
\newblock The generative ai paradox:“what it can create, it may not understand”.
\newblock In \emph{The Twelfth International Conference on Learning Representations}.

\bibitem[{Wu et~al.(2024)Wu, Hu, Shi, Dziri, Suhr, Ammanabrolu, Smith, Ostendorf, and Hajishirzi}]{wu2024fine}
Zeqiu Wu, Yushi Hu, Weijia Shi, Nouha Dziri, Alane Suhr, Prithviraj Ammanabrolu, Noah~A Smith, Mari Ostendorf, and Hannaneh Hajishirzi. 2024.
\newblock Fine-grained human feedback gives better rewards for language model training.
\newblock \emph{Advances in Neural Information Processing Systems}, 36.

\bibitem[{Xie et~al.(2020)Xie, Pan, Wang, Kan, and Feng}]{xie2020exploring}
Yuxi Xie, Liangming Pan, Dongzhe Wang, Min-Yen Kan, and Yansong Feng. 2020.
\newblock Exploring question-specific rewards for generating deep questions.
\newblock \emph{arXiv preprint arXiv:2011.01102}.

\bibitem[{Yao et~al.(2023{\natexlab{a}})Yao, Kantu, Wei, Tran, Duan, Kwon, Yang, Yu et~al.}]{yao2023readme}
Zonghai Yao, Nandyala~Siddharth Kantu, Guanghao Wei, Hieu Tran, Zhangqi Duan, Sunjae Kwon, Zhichao Yang, Hong Yu, et~al. 2023{\natexlab{a}}.
\newblock Readme: Bridging medical jargon and lay understanding for patient education through data-centric nlp.
\newblock \emph{arXiv preprint arXiv:2312.15561}.

\bibitem[{Yao et~al.(2023{\natexlab{b}})Yao, Schloss, and Selvaraj}]{yao2023improving}
Zonghai Yao, Benjamin~J Schloss, and Sai~P Selvaraj. 2023{\natexlab{b}}.
\newblock Improving summarization with human edits.
\newblock \emph{arXiv preprint arXiv:2310.05857}.

\bibitem[{Yao et~al.(2024)Yao, Zhang, Tang, Bian, Zhao, Yang, Wang, Zhou, Jang, Ouyang et~al.}]{yao2024medqa}
Zonghai Yao, Zihao Zhang, Chaolong Tang, Xingyu Bian, Youxia Zhao, Zhichao Yang, Junda Wang, Huixue Zhou, Won~Seok Jang, Feiyun Ouyang, et~al. 2024.
\newblock Medqa-cs: Benchmarking large language models clinical skills using an ai-sce framework.
\newblock \emph{arXiv preprint arXiv:2410.01553}.

\bibitem[{Zeng et~al.(2023)Zeng, Yu, Gao, Meng, Goyal, and Chen}]{zeng2023evaluating}
Zhiyuan Zeng, Jiatong Yu, Tianyu Gao, Yu~Meng, Tanya Goyal, and Danqi Chen. 2023.
\newblock Evaluating large language models at evaluating instruction following.
\newblock \emph{arXiv preprint arXiv:2310.07641}.

\bibitem[{Zhang et~al.(2024{\natexlab{a}})Zhang, D'Haro, Chen, Zhang, and Li}]{zhang2024comprehensive}
Chen Zhang, Luis~Fernando D'Haro, Yiming Chen, Malu Zhang, and Haizhou Li. 2024{\natexlab{a}}.
\newblock A comprehensive analysis of the effectiveness of large language models as automatic dialogue evaluators.
\newblock In \emph{Proceedings of the AAAI Conference on Artificial Intelligence}, volume~38, pages 19515--19524.

\bibitem[{Zhang et~al.(2021)Zhang, Guo, Chen, Fan, and Cheng}]{zhang2021review}
Ruqing Zhang, Jiafeng Guo, Lu~Chen, Yixing Fan, and Xueqi Cheng. 2021.
\newblock A review on question generation from natural language text.
\newblock \emph{ACM Transactions on Information Systems (TOIS)}, 40(1):1--43.

\bibitem[{Zhang et~al.(2024{\natexlab{b}})Zhang, Shen, Wu, Peng, Wang, Zhuang, and Lu}]{zhang2024self}
Wenqi Zhang, Yongliang Shen, Linjuan Wu, Qiuying Peng, Jun Wang, Yueting Zhuang, and Weiming Lu. 2024{\natexlab{b}}.
\newblock Self-contrast: Better reflection through inconsistent solving perspectives.
\newblock \emph{arXiv preprint arXiv:2401.02009}.

\bibitem[{Zhang and Gao(2023)}]{zhang2023towards}
Xuan Zhang and Wei Gao. 2023.
\newblock Towards llm-based fact verification on news claims with a hierarchical step-by-step prompting method.
\newblock \emph{arXiv preprint arXiv:2310.00305}.

\bibitem[{Zhang et~al.(2023)Zhang, Li, Cui, Cai, Liu, Fu, Huang, Zhao, Zhang, Chen et~al.}]{zhang2023siren}
Yue Zhang, Yafu Li, Leyang Cui, Deng Cai, Lemao Liu, Tingchen Fu, Xinting Huang, Enbo Zhao, Yu~Zhang, Yulong Chen, et~al. 2023.
\newblock Siren's song in the ai ocean: a survey on hallucination in large language models.
\newblock \emph{arXiv preprint arXiv:2309.01219}.

\bibitem[{Zhao et~al.(2023)Zhao, Jin, Chen, Peng, and Yu}]{Zhao_2023}
Zhengyun Zhao, Qiao Jin, Fangyuan Chen, Tuorui Peng, and Sheng Yu. 2023.
\newblock \href {https://doi.org/10.1038/s41597-023-02814-8} {A large-scale dataset of patient summaries for retrieval-based clinical decision support systems}.
\newblock \emph{Scientific Data}, 10(1).

\bibitem[{Zhao et~al.(2022)Zhao, Hou, Wang, Yu, Liu, and Ma}]{zhao2022educational}
Zhenjie Zhao, Yufang Hou, Dakuo Wang, Mo~Yu, Chengzhong Liu, and Xiaojuan Ma. 2022.
\newblock Educational question generation of children storybooks via question type distribution learning and event-centric summarization.
\newblock \emph{arXiv preprint arXiv:2203.14187}.

\bibitem[{Zheng et~al.(2023)Zheng, Chiang, Sheng, Zhuang, Wu, Zhuang, Lin, Li, and Li}]{zheng2023eric}
Lianmin Zheng, Wei-Lin Chiang, Ying Sheng, Siyuan Zhuang, Zhanghao Wu, Yonghao Zhuang, Zi~Lin, Zhuohan Li, and Dacheng Li. 2023.
\newblock Eric.
\newblock \emph{P Xing, Hao Zhang, Joseph E. Gonzalez, and Ion Stoica. Judging llm-as-a-judge with mt-bench and chatbot arena}, 2(6):7.

\bibitem[{Zheng et~al.(2024)Zheng, Chiang, Sheng, Zhuang, Wu, Zhuang, Lin, Li, Li, Xing et~al.}]{zheng2024judging}
Lianmin Zheng, Wei-Lin Chiang, Ying Sheng, Siyuan Zhuang, Zhanghao Wu, Yonghao Zhuang, Zi~Lin, Zhuohan Li, Dacheng Li, Eric Xing, et~al. 2024.
\newblock Judging llm-as-a-judge with mt-bench and chatbot arena.
\newblock \emph{Advances in Neural Information Processing Systems}, 36.

\end{thebibliography}

\appendix

\newpage


\section{Topics and Test points}
\label{apx:topic_testpoint}

\textbf{Topics \(t\)} refer to a list of target topics selected from 41 potential topics outlined in the NBME official guidelines, categorized into the following 10 sections:

\begin{enumerate}
    \item Diagnosis - Causes and Mechanisms
    \item Diagnosis - Obtaining and Predicting History and Physical Examination
    \item Diagnosis - Selecting and Interpreting Laboratory and Diagnostic Studies
    \item Diagnosis - Formulating the Diagnosis
    \item Diagnosis - Determining Prognosis/Outcome
    \item Management - Health Maintenance and Disease Prevention
    \item Management - Selecting and Monitoring Pharmacotherapy
    \item Management - Clinical Interventions/Treatments
    \item Management - Selecting Clinical Interventions (Mixed Management)
    \item Management - Monitoring/Surveillance for Disease Recurrence or Progression
\end{enumerate}

\noindent\textbf{Test points \(k\)} refer to the core concepts closely related to the correct answer. We employ the ColBERT retriever~\footnote{\url{https://github.com/stanford-futuredata/ColBERT}}, denoted as \(\pi_{rtr}\), to retrieve suitable test points from 18 sections.

\begin{enumerate}
    \item General Principles of Foundational Science
    \item Immune System
    \item Blood \& Lymphoreticular System
    \item Behavioral Health
    \item Nervous System \& Special Senses
    \item Skin \& Subcutaneous Tissue
    \item Musculoskeletal System
    \item Cardiovascular System
    \item Respiratory System
    \item Gastrointestinal System
    \item Renal \& Urinary System
    \item Pregnancy, Childbirth, \& the Puerperium
    \item Female Reproductive System \& Breast
    \item Male Reproductive System
    \item Endocrine System
    \item Multisystem Processes \& Disorders
    \item Biostatistics, Epidemiology/Population Health, \& Interpretation of the Medical Literature
    \item Social Sciences
\end{enumerate}

\paragraph{Stopping Criteria}
\label{apx:stop}
The stopping criteria for the MCQG-SRefine included two checks. One was that the iteration would stop if the total critique score of all the components exceeded 90\% of the maximum possible score, and the other was if the number of iterations exceeded 4.

\begin{table}[!ht]
\centering
\begin{tabular}{|>{\raggedright\arraybackslash}m{7.2cm}|}
\hline
\textbf{Task Input and Response} \(R_a, R_b\) \\ \hline
\textbf{I}: Generate a USMLE question based on the following \textbf{medical case}: A 84-year-old male presents with chest pain ....... \textbf{Topic}: pathophysiology of sepsis \textbf{Test point}: sepsis \\ 
\(R_a\): system-a generated USMLE question\\ 
\(R_b\): system-b generated USMLE question\\ 
\hline

\textbf{Critique} \(F_{crq}(I, R_b)\)\\ \hline
The generated question fails to meet the quality and difficulty requirements for the USMLE because ...... The following suggestions could be considered: ..... \\ 
\textbf{Score for Question} \(R_b\): 1/10 \\ 
\hline

\textbf{Correction} \(F_{crr}(I, R_b, F_{crq})\) \textbf{with} \(F_{crq}\) \\ \hline
Here is the revision for question B:\\
\textbf{Context}: The 84-year-old female patient presented to the emergency room with ...... \\
\textbf{Question}: What is the most likely complication that has developed in this patient's case?
\textbf{Correct answer}: Septic shock
\textbf{Distractor options}: a) b) c) d) \\
\hline

\textbf{Comparison} \(F_{cmp}(I, R_a, R_b)\) \\ \hline
Question A has higher quality than B as it ...... On the contrary, question B ...... \\ 
\textbf{Preference Response: A} \\ \hline
\end{tabular}
\caption{Examples of different LLM feedback (Critique, Correction, Comparison) for USMLE MCQG.}
\end{table}


\section{Ablation study and output analysis of MCQG-SRefine}
\label{apx:ablation_study}

\subsection{Round-wise metrics}

As shown in Table~\ref{tab:Round-wise metrics - Human data}, the round-wise metrics for human data reveal nuanced trends in the model's performance through multiple feedback iterations. The total score exhibited a modest increase from Round 1 (0.9031) to Round 3 (0.9062), followed by a slight decline in Round 4 (0.9006), suggesting that while the self-feedback process contributes positively, its benefits may diminish with excessive iterations. The context score notably improved from 0.9683 in Round 1 to 0.9712 in Round 3, reflecting enhanced model comprehension of the context through feedback. However, performance in other areas fluctuated: the question score showed a slight decline across rounds, and while the correct answer score improved from Round 1 to Round 3, it decreased slightly in Round 4. Conversely, the distractor option score showed steady improvement, culminating in the highest score by Round 4. The reasoning score, however, demonstrated a gradual decline over the rounds. The standard deviation for most metrics either decreased or remained stable, indicating more consistent performance. Overall, while certain components of the model benefited from the feedback process, others did not, highlighting the complexity of balancing improvements across different aspects of question generation.

\begin{table*}
\centering
\caption{Round-wise metrics - Human data}
\begin{tabular}{lcccc}
\hline
\textbf{Mean | Std. dev} & \textbf{Round 1} & \textbf{Round 2} & \textbf{Round 3} & \textbf{Round 4} \\
\hline
\textbf{Context score} & 0.9683 \,|\, 0.068 & 0.96705 \,|\, 0.060 & \textbf{0.9712 \,|\, 0.055} & 0.9679 \,|\, 0.062 \\
\textbf{Question score} & \textbf{0.8394} \,|\, 0.098 & 0.8322 \,|\, 0.09 & 0.8277 \,|\, \textbf{0.089} & 0.8195 \,|\, 0.091 \\
\textbf{C. Answer score} & 0.8051 \,|\, 0.168 & 0.8212 \,|\, 0.167 & \textbf{0.8358 \,|\, 0.161} & 0.8284 \,|\, 0.172 \\
\textbf{Distractor option score} & 0.9378 \,|\, 0.096 & 0.9344 \,|\, 0.089 & 0.9376 \,|\, 0.086 & \textbf{0.9414 \,|\, 0.084} \\
\textbf{Reasoning score} & \textbf{0.9650 \,|\, 0.116} & 0.96168 \,|\, 0.122 & 0.9589 \,|\, 0.132 & 0.9456 \,|\, 0.150 \\
\textbf{Total score} & 0.9031 \,|\, 0.061 & 0.9033 \,|\, \textbf{0.052} & \textbf{0.9062} \,|\, 0.0569 & 0.9006 \,|\, 0.058 \\
\hline
\end{tabular}
\label{tab:Round-wise metrics - Human data}
\end{table*}

As shown in Table~\ref{tab:Round-wise metrics - Machine data}, the round-wise metrics for machine data demonstrate a stable overall performance, with the total score showing minimal fluctuation across rounds, peaking slightly in Round 2 (0.907) and ending at 0.903 in Round 4. This suggests that while the feedback process maintains performance, it does not lead to significant improvements. The context score exhibits a notable increase from 0.9688 in Round 1 to 0.974 in Round 4, coupled with a decrease in standard deviation, indicating enhanced and more consistent context understanding. The C. Answer score also shows a gradual upward trend, improving from 0.795 in Round 1 to 0.810 in Round 4, reflecting slow but steady progress in answer generation. The distractor option score remains relatively high and stable across rounds, while the reasoning score experiences a decline from 0.967 in Round 1 to 0.948 in Round 3, with a slight recovery to 0.951 in Round 4. Variability in performance is evident, with some metrics, such as the context score, showing decreased standard deviation, indicating more consistency, while others, like the reasoning score, exhibit increased variability, highlighting differential effects on the consistency of various components.

\begin{table*}
\centering
\caption{Round-wise metrics - Machine data}
\begin{tabular}{lcccc}
\hline
\textbf{Mean | Std. dev} & \textbf{Round 1} & \textbf{Round 2} & \textbf{Round 3} & \textbf{Round 4} \\
\hline
\textbf{Context score} & 0.9688 \,|\, 0.0686 & \textbf{0.975 \,|\, 0.053} & 0.970 \,|\, 0.0647 & 0.974 \,|\, \textbf{0.0511} \\
\textbf{Question score} & 0.838 \,|\, \textbf{0.090} & \textbf{0.848 \,|\, 0.092} & 0.836 \,|\, 0.0941 & 0.837 \,|\, 0.0939 \\
\textbf{C. Answer score} & 0.795 \,|\, 0.150 & 0.809 \,|\, \textbf{0.158} & 0.799 \,|\, 0.151 & \textbf{0.810 \,|\, 0.154} \\
\textbf{Distractor option score} & 0.943 \,|\, 0.080 & 0.940 \,|\, 0.088 & 0.941 \,|\, 0.081 & \textbf{0.944 \,|\, 0.0825} \\
\textbf{Reasoning score} & \textbf{0.967 \,|\, 0.1098} & 0.965 \,|\, 0.121 & 0.948 \,|\, 0.147 & 0.951 \,|\, 0.149 \\
\textbf{Total score} & 0.902 \,|\, \textbf{0.051} & \textbf{0.907 \,|\, 0.0564} & 0.899 \,|\, 0.058 & 0.903 \,|\, 0.0561 \\
\hline
\end{tabular}
\label{tab:Round-wise metrics - Machine data}
\end{table*}

Based on the analysis of the round-wise metrics from both human and machine data, the self-feedback process demonstrates the most significant improvements in context understanding and answer generation. These enhancements are evidenced by the consistent upward trends in context and correct answer scores across rounds. However, the data also suggests a potential trade-off between reasoning ability and other metrics, as seen in the decline of reasoning scores, particularly after the initial rounds. The plateau or slight decline in overall performance after 2-3 rounds indicates that the benefits of the feedback process diminish with excessive iterations, implying that a limited number of rounds may be optimal for maximizing improvements without compromising other aspects of question generation. These findings highlight the importance of balancing the feedback process to achieve comprehensive improvements across all key metrics.

\begin{table*}
\centering
\caption{Round-wise metrics - Human data}
\begin{tabular}{lcccc}
\hline
\textbf{Mean | Std. dev} & \textbf{Round 1} & \textbf{Round 2} & \textbf{Round 3} & \textbf{Round 4} \\
\hline
\textbf{Question length} & 203.008 \,|\, 46.15 & 111.09 \,|\, 24.14 & 91.47 \,|\, 19.68 & \textbf{84.11 \,|\, 17.80} \\
\textbf{Accuracy (QA)} & 0.898 \,|\, 0.302 & 0.89 \,|\, 0.311 & \textbf{0.903 \,|\, 0.295} & 0.86 \,|\, 0.339 \\
\textbf{C. Answer = Keypoint} & 0.193 \,|\, 0.394 & 0.170 \,|\, 0.375 & 0.153 \,|\, 0.360 & \textbf{0.146 \,|\, 0.353} \\
\hline
\end{tabular}
\label{tab:Round-wise metrics II - Human data}
\end{table*}

\begin{table*}
\centering
\caption{Round-wise metrics - Machine data}
\begin{tabular}{lcccc}
\hline
\textbf{Mean | Std. dev} & \textbf{Round 1} & \textbf{Round 2} & \textbf{Round 3} & \textbf{Round 4} \\
\hline
\textbf{Question length} & 204.68 \,|\, 49.04 & 117.85 \,|\, 25.33 & 96.113 \,|\, 19.52 & \textbf{87.5 \,|\, 17.21} \\
\textbf{Accuracy (QA)} & 0.89 \,|\, 0.30 & \textbf{0.908 \,|\, 0.288} & 0.865 \,|\, 0.341 & 0.875 \,|\, 0.33 \\
\textbf{C. Answer = Keypoint} & 0.083 \,|\, 0.276 & 0.080 \,|\, 0.272 & 0.073 \,|\, 0.261 & \textbf{0.065 \,|\, 0.247} \\
\hline
\end{tabular}
\label{tab:Round-wise metrics II - Machine data}
\end{table*}

In addition, the analysis of the round-wise metrics in Table~\ref{tab:Round-wise metrics II - Human data} and \ref{tab:Round-wise metrics II - Machine data} reveals several key trends in the performance of the question generation pipeline. As the rounds progress, the context length consistently decreases, indicating that the pipeline effectively refines the context by excluding extraneous information, leading to more precise and focused questions. The accuracy of the QA component improves over the rounds, suggesting that the iterative process enhances the overall quality of the questions, making them more answerable by the LLM. However, this improvement reaches a point of diminishing returns in the later rounds, implying that a limited number of iterations may be optimal. On the other hand, the equality between the correct answer and the keypoint deteriorates over the rounds, indicating that the pipeline makes the correct answer more subtly related to the keypoint rather than directly copying it. This shift suggests that while the pipeline reduces redundancy, it may also introduce complexity that could impact the clarity and directness of the correct answer-keypoint relationship.

\begin{figure}
    \centering
    \includegraphics[width=\linewidth]{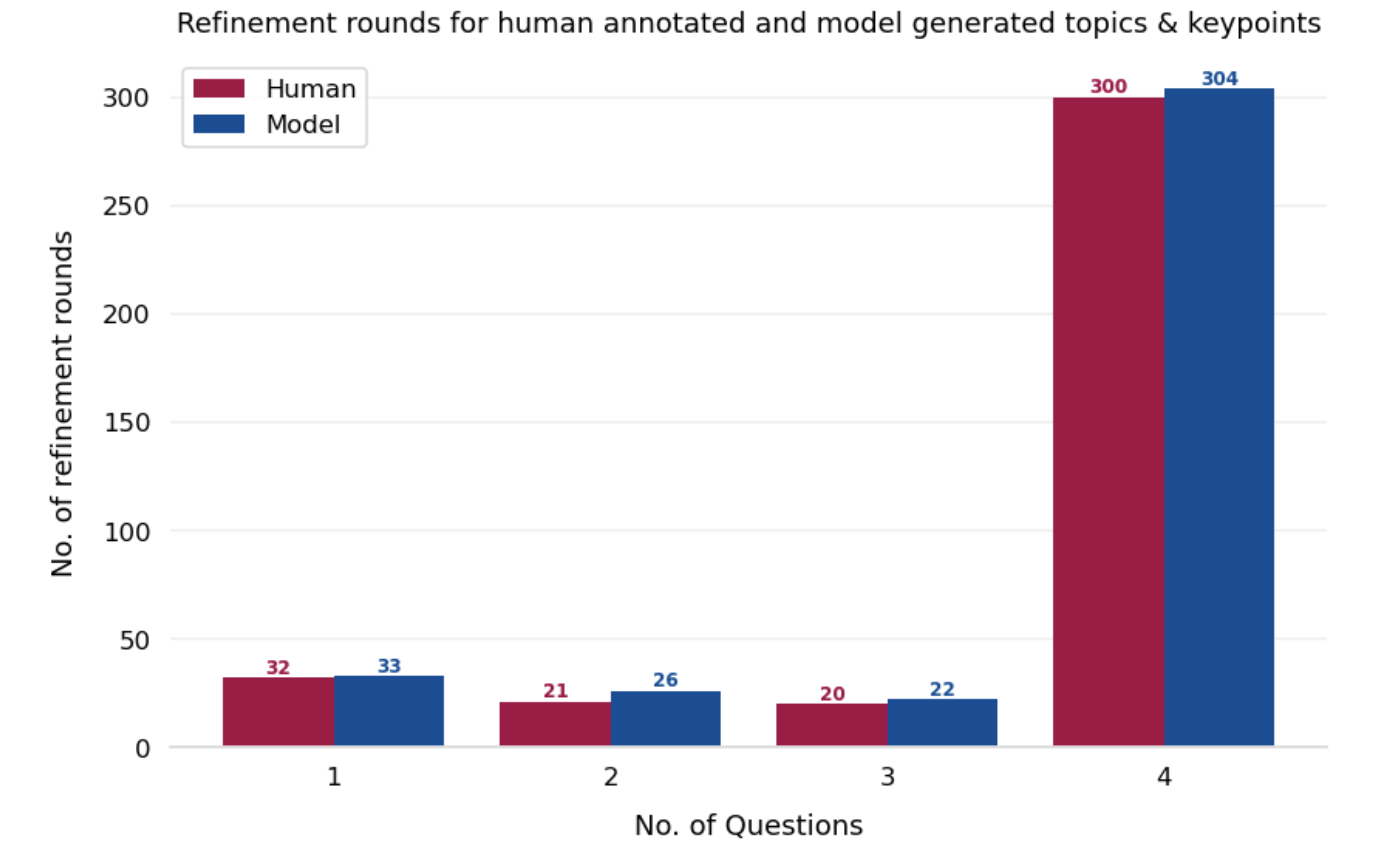}
    \caption{No. of refinement rounds. This histogram indicates that most of the data points(78.9\% for human, 80.4\% for machine) take the full 4 iterations in pursuit of the threshold score.}
    \label{fig:No. of refinement rounds.png}
\end{figure}

\begin{figure}
    \centering
    \includegraphics[width=\linewidth]{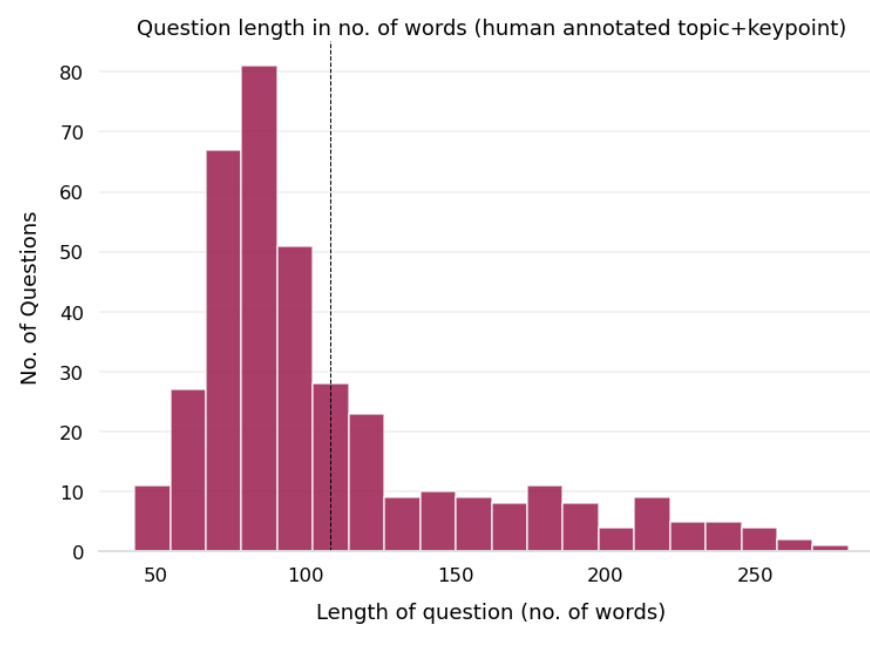}
    \caption{Length of the generated question human.}
    \label{fig:Length of the generated question human}
\end{figure}

\begin{figure}
    \centering
    \includegraphics[width=\linewidth]{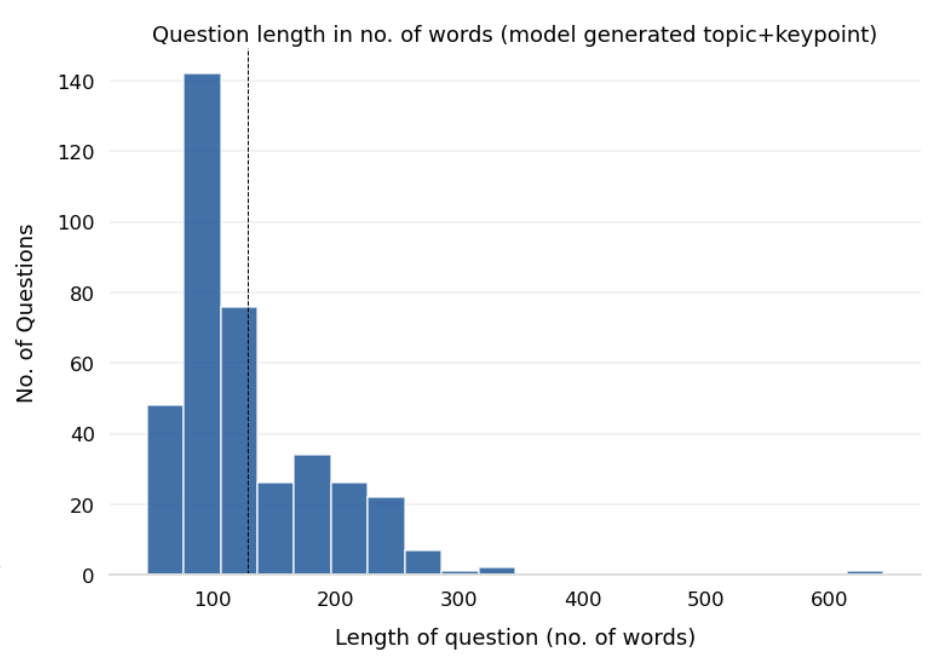}
    \caption{Length of the generated question machine.}
    \label{fig:Length of the generated question machine}
\end{figure}



\subsection{Other Basic statistics}
We also calculate Refinement rounds for human-annotated and model-generated topics \& key points in Figure~\ref{fig:No. of refinement rounds.png}; 
Question length in no. of words (human annotated topic+keypoint) in Figure~\ref{fig:Length of the generated question human}; 
Question length in no. of words (machine annotated topic+keypoint) in Figure~\ref{fig:Length of the generated question machine};

\begin{figure}[!ht]
    \centering
    \includegraphics[width=\linewidth]{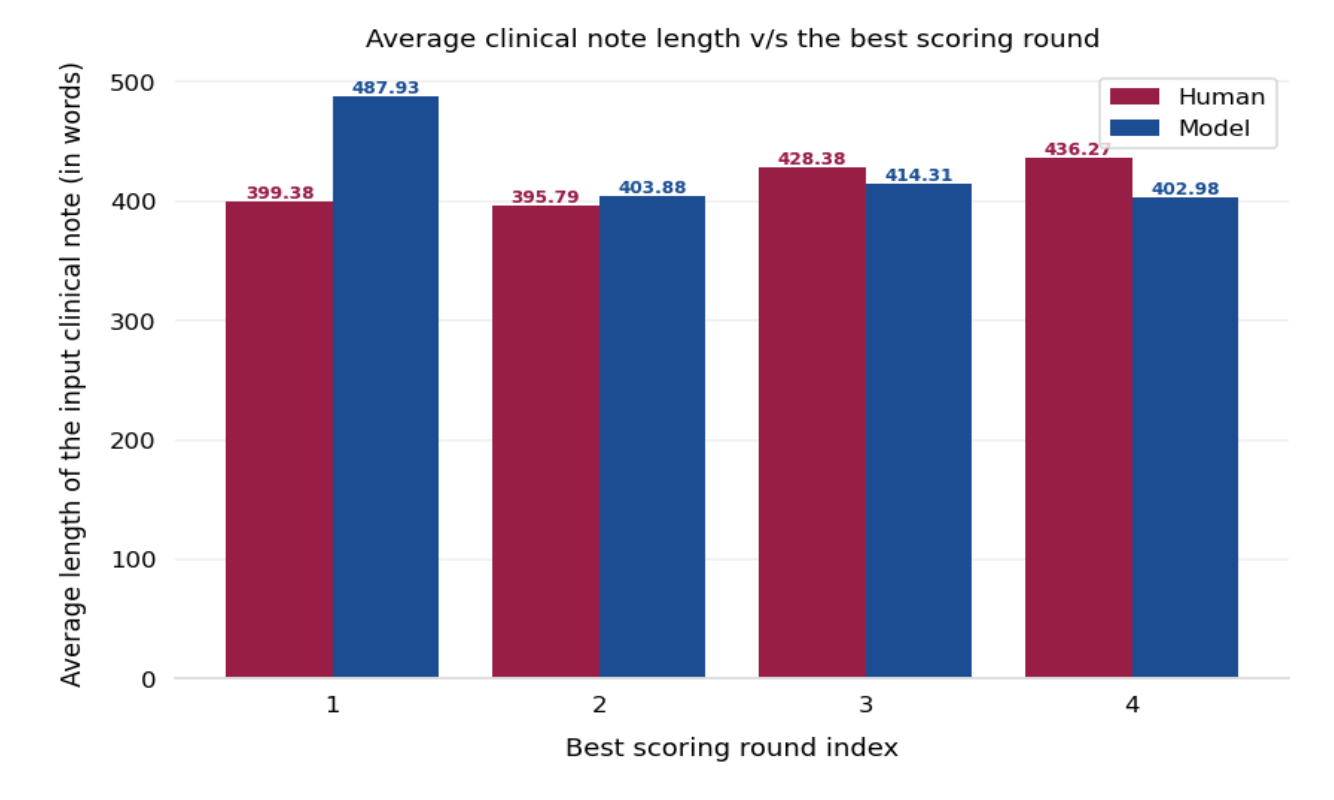}
    \caption{Clinical note length v/s best scoring round.}
    \label{fig:Clinical note length v/s best scoring round}
\end{figure}

\begin{figure}[!ht]
    \centering
    \includegraphics[width=\linewidth]{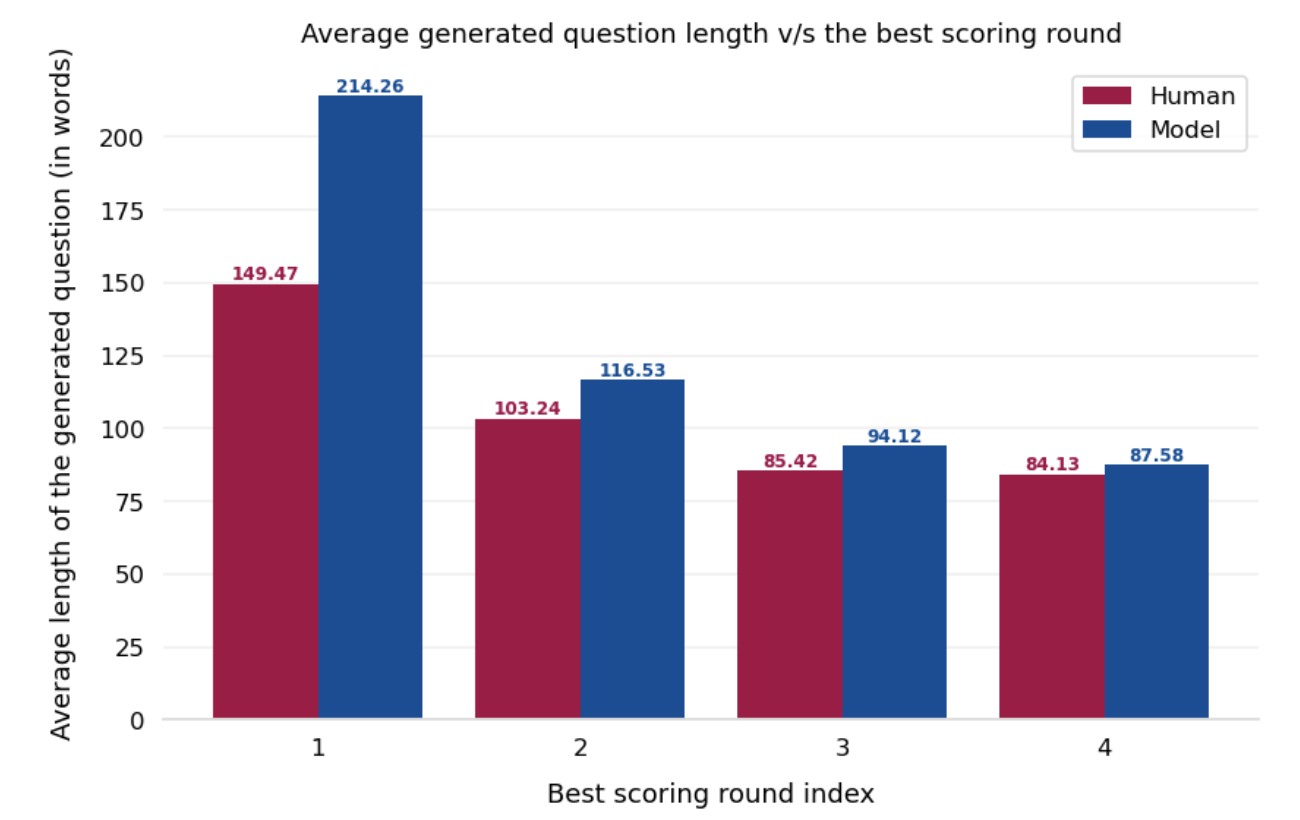}
    \caption{Question length v/s best scoring round.}
    \label{fig:Question length v/s best scoring round}
\end{figure}

\begin{table*}[ht]
\centering
\begin{tabular}{|c|c|c|c|c|}
\hline
\textbf{Data Type} & \textbf{Variable} & \textbf{Pearson} & \textbf{Spearman} & \textbf{Kendall} \\ \hline
\textbf{Human} & Rounds with Clinical Note Length & -0.9103 & -1.0 & -1.0 \\ \hline
\textbf{Machine} & Rounds with Clinical Note Length & -0.8869 & -1.0 & -1.0 \\ \hline
\textbf{Human} & Rounds with Question Length & -0.9103 & -1.0 & -1.0 \\ \hline
\textbf{Machine} & Rounds with Question Length & -0.8869 & -1.0 & -1.0 \\ \hline
\end{tabular}
\caption{Correlation between different rounds with Clinical Note Length and Question Length for both Human and Machine data.}
\label{tab:Correlation analysis}
\end{table*}

\subsection{Correlation analysis}
Our analysis in Figure~\ref{fig:Clinical note length v/s best scoring round} ~\ref{fig:Question length v/s best scoring round} and Table~\ref{tab:Correlation analysis} reveals a strong inverse relationship between the length of clinical notes/questions and the number of rounds required to achieve the best score in both human-generated and machine-generated data. The perfect negative Spearman and Kendall correlations (-1.0) indicate that as the length of inputs increases, fewer rounds are consistently needed to reach optimal performance. Human-generated content shows slightly stronger linear correlations compared to machine-generated data. These findings suggest that longer, more detailed inputs provide richer information, allowing the model to converge on optimal performance more quickly. Practically, this implies that shorter inputs may benefit from more iterative rounds, while longer inputs may require fewer rounds to achieve the best results. The consistency of this pattern across both clinical note length and question length underscores the importance of input complexity in determining the efficiency of the iterative improvement process.

\begin{table}
    \centering
    \scalebox{0.85}{
    \begin{tabular}{lccc}
        \toprule
        \textbf{Model} & \textbf{No POS Bias} & \textbf{PA.} & \textbf{$\tau$} \\
        \midrule
        GPT-4 & 65\% & 61.53\% & 0.2121 \\
        GPT-4o & 42.5\% & 58.82\% & 0.3340 \\
        GPT-3.5-turbo & 35\% & 64.29\% & 0.0572 \\
        Claude-3-haiku & 42.5\% & 41.18\% & -0.0636 \\
        Claude-3.5-sonnet & 37.5\% & 46.67\% & 0.1517 \\
        Claude-3-opus & 47.5\% & 68.42\% & 0.3152 \\
        \bottomrule
    \end{tabular}
    }
    \caption{Percentage Agreement (PA.) and Kendall's Tau ($\tau$) for Different Models Compared to Expert X using valid data (e.g., order matching for both GPT-4 generated output first setting and MCQG-SRefine generated output first setting).}
    \label{tab:llm_as_judge_comparison}
\end{table}

\section{Improving reliability of LLM-as-judge}

\subsection{Greedy Algorithm}

The Greedy algorithm iteratively constructs an optimal set of aspects by sequentially adding the most correlated aspects. Let $A = \{a_1, a_2, ..., a_n\}$ be the set of all aspects, and $h$ be the human preference data. The algorithm proceeds as follows:

\begin{algorithm}[H]
\caption{Greedy Aspect Selection}
\begin{algorithmic}[1]
\State Sort $A$ in descending order of correlation with $h$
\State $S \gets \emptyset$ \Comment{Initialize empty set of selected aspects}
\State $c_{best} \gets 0$ \Comment{Best correlation so far}
\For{each $a_i \in A$}
    \State $S' \gets S \cup \{a_i\}$
    \State $r \gets \text{CalculateRating}(S')$
    \State $c \gets \text{Correlation}(r, h)$
    \If{$c > c_{best}$}
        \State $S \gets S'$
        \State $c_{best} \gets c$
    \EndIf
\EndFor
\State \Return $S, c_{best}$
\end{algorithmic}
\end{algorithm}

where $\text{CalculateRating}(S)$ computes the final rating score using the aspects in set $S$, and $\text{Correlation}(r, h)$ calculates either the percentage agreement or Cohen's kappa between the rating $r$ and human preferences $h$.

\subsection{All-Combo Algorithm}

The All-Combo algorithm exhaustively evaluates all possible combinations of the top $n$ aspects to find the optimal subset. Let $k$ be the number of aspects in a combination, where $1 \leq k \leq n$.

\begin{algorithm}[H]
\caption{All-Combo Aspect Selection}
\begin{algorithmic}[1]
\State $C_{best}^{PA} \gets \emptyset, c_{best}^{PA} \gets 0$ \Comment{Best for Percentage Agreement}
\State $C_{best}^{\kappa} \gets \emptyset, c_{best}^{\kappa} \gets 0$ \Comment{Best for Cohen's Kappa}
\For{$k = 1$ to $n$}
    \For{each combination $C \in \binom{A}{k}$}
        \State $r \gets \text{CalculateRating}(C)$
        \State $c_{PA} \gets \text{PercentageAgreement}(r, h)$
        \State $c_{\kappa} \gets \text{CohenKappa}(r, h)$
        \If{$c_{PA} > c_{best}^{PA}$}
            \State $C_{best}^{PA} \gets C, c_{best}^{PA} \gets c_{PA}$
        \EndIf
        \If{$c_{\kappa} > c_{best}^{\kappa}$}
            \State $C_{best}^{\kappa} \gets C, c_{best}^{\kappa} \gets c_{\kappa}$
        \EndIf
    \EndFor
\EndFor
\State \Return $C_{best}^{PA}, c_{best}^{PA}, C_{best}^{\kappa}, c_{best}^{\kappa}$
\end{algorithmic}
\end{algorithm}

This algorithm returns the best aspect combinations and their corresponding correlation scores for both percentage agreement and Cohen's kappa, as these may differ depending on the evaluation metric used.

\section{More Experiments for LLM-as-Judge}
\label{Sec:RQ_LLM-as-Judge}

Determine if \texttt{LLM-as-Judge} is an effective metric. RQ3: Rating or comparison; RQ4: Which LLM is the better judge?

\subsection{Settings}

To explore \textbf{RQ3}, we utilized ratings generated during the MCQG-SRefine critique step as \texttt{LLM-as-Judge (rating)} scores. For \texttt{LLM-as-Judge (comparison)}, we employed the same guidelines and settings used in the human evaluation from RQ1 in main results, prompting the model to choose a preferred output between the two system-generated questions (e.g., GPT-4 and MCQG-SRefine). 
To ensure the results were not biased by the inherent positional preferences of the LLM (e.g., \textbf{position bias})~\cite{wang2023large,zheng2023eric,zeng2023evaluating}, we collected preferences from two sequence settings: one where the GPT-4 generated output was shown first followed by the MCQG-SRefine output, and the reverse order. We filtered out preferences that did not match across the two sequence settings, using the remaining data for further analysis. 
To ensure the results were not influenced by \textbf{length bias} (i.e., LLMs
like GPT-4 prefer longer generations during their
automatic evaluation) ~\cite{wang2023far,zeng2023evaluating}, we deliberately included those data points in the evaluation data which had the least context length ratio for the initial and the MCQG-SRefine questions. These examples constituted half the human evaluation data and the rest were chosen randomly from the dataset.
Notably, the guidelines used in human evaluation, \texttt{LLM-as-Judge (rating)}, and \texttt{LLM-as-Judge (comparison)} were consistent. Thus, by standardizing the evaluation data, we could fairly compare the judgments of Expert X and \texttt{LLM-as-Judge (rating/comparison)} under identical settings to assess the reliability of \texttt{LLM-as-Judge}.

To investigate \textbf{RQ4}, we prompted various LLMs~\footnote{\texttt{GPT-3.5-turbo}, \texttt{GPT-4}, \texttt{GPT-4o}, \texttt{Claude-3-haiku/opus}, and \texttt{Claude-3.5-sonnet}} using the \texttt{LLM-as-Judge (comparison)} settings and compare the preference with Expert X.

\subsection{Results}

For \textbf{RQ3}, we found that the \textbf{rating} method performs slightly better than the \textbf{comparison} method. Percentage Agreement between GPT-4-Comparison and Expert X: Total agreement: 53.8\% (Human: 58.33\%, Machine: 50\%). Percentage Agreement between GPT-4-Rating and Expert X: Total agreement: 61.538\% (Human: 58.33\%, Machine: 64.285\%).
For \textbf{RQ4}, as shown in Appendix Table~\ref{tab:llm_as_judge_comparison} and ~\ref{tab:llm_as_judge_comparison_full}, we observed that GPT-4 as the LLM-as-judge has the least position bias (35\% of the data showed inconsistent results under two different orders, while other models showed more than 50\% inconsistency). Moreover, GPT-4 has a relatively higher correlation with human evaluations in the valid data (i.e., data without position bias) with a percentage agreement of 61.53\% and a Cohen's kappa of 0.2121. On the other hand, Claude-3-opus showed 52.5\% position bias but had the highest correlation with human evaluation in the valid data, with a percentage agreement of 68.42\% and a Cohen's kappa of 0.3152.

\begin{table}
    \centering
    \scalebox{0.85}{
    \begin{tabular}{c|c|ccc|cc}
        \hline
        & & \multicolumn{3}{c|}{Quality} & \multicolumn{2}{c}{Difficulty} \\
        & \textbf{$t$, $k$} & \textbf{GPT4} & \textbf{Ours} & \textbf{Tie} & \textbf{GPT4} & \textbf{Ours} \\
        \hline
        \multirow{2}{*}{\textbf{1}} & H. & 15\% & \textbf{80\%} & 5\% & 50:35:15 & 35:30:35 \\
         & M. & 20\% & \textbf{80\%} & 0\% & 70:20:10 & 45:35:20 \\
        \hline
        \multirow{2}{*}{\textbf{2}} & H. & 20\% & \textbf{70\%} & 5\% & 60:35:5 & 30:65:5 \\
         & M. & 20\% & \textbf{75\%} & 0\% & 65:30:5 & 60:35:5 \\
        \hline
        \multirow{2}{*}{\textbf{x}} & H. & 15\% & \textbf{75\%} & 10\% & 70:20:10 & 15:45:40 \\
         & M. & 20\% & \textbf{70\%} & 10\% & 75:20:5 & 50:40:10 \\
        \hline
    \end{tabular}
    }
    \caption{The Quality part shows expert preference counts for the GPT-4 generated question and the MCQG-SRefine (Ours) question. The data is divided into Human (H.) and Machine (M.) based on how the topic and key points were generated. Expert x represents the preferences reached by the experts after a round of preliminary annotation.
    The percentage agreement between Expert 1 and Expert 2 is 87.5\% (Human: 90\%, Machine: 85\%).
    The Cohen's kappa between Expert 1 and Expert 2 is 0.66722 (Human: 0.75, Machine: 0.571428), indicating substantial reliability.
    The Difficulty part shows the difficulty level distribution (e.g., Easy: Medium: Hard) for GPT-4 and Ours, annotated by human experts.
    Compared with GPT-4, the MCQG-SRefine pipeline generates better quality USMLE multiple-choice questions while producing significantly more medium and hard questions.
    }
    \label{tab:evaluations}
\end{table}

\begin{table}
\centering
\scalebox{0.85}{
\begin{tabular}{l|c|c}
\hline
\textbf{avg. score} & \textbf{MCQG-SRefine} & \textbf{GPT-4} \\ \hline
Context & \footnotesize{0.96 (0.96, 0.95, 0.99)} & \footnotesize{0.95 (0.91, 0.96, 0.99)} \\ \hline
Question & 0.91 (0.89, 0.93) & 0.87 (0.84, 0.9) \\ \hline
CQT ANS & 0.79 (0.64, 0.94) & 0.64 (0.45, 0.83) \\ \hline
Distractor & 0.94 (0.94) & 0.89 (0.89) \\ \hline
Reasoning  & 0.985 (0.99, 0.98) & 0.97 (0.98, 0.96) \\ \hline
Total  & 0.917 & 0.864 \\ \hline
\end{tabular}
}
\caption{The table also shows the rating results for the 10 different aspects: Context (concision, relevance, misdirection), Question (concluding, clarity), Correct Answer (occurrence, depth of understanding), Distractor (common mistakes), and Reasoning (logical flow, evidence-based reasoning).}
\label{tab:llm-as-judge-results}
\end{table}

Finally, we also calculated self-BLEU scores as diversity metrics in two scenarios: 1. Self-BLEU for GPT-4 generated USMLE MCQs. 2. Self-BLEU for USMLE MCQs generated by the MCQG-SRefine framework.
Our analysis shows that MCQG-SRefine reduces the self-BLEU score by approximately 25.9\% compared to GPT-4's original generation (lower self-BLEU indicates less similarity and hence greater diversity). We believe this is because SRefine adjusts the initial contexts—typically more homogeneous due to covering broader portions of the clinical note—to make them more focused on individual topics or test points, thereby enhancing diversity. Below are the results:

\begin{table}[h!]
\centering
\begin{tabular}{lcc}
\hline
\textbf{Model}                & \textbf{Human}     & \textbf{Machine}    \\ \hline
GPT-4                         & 0.5839      & 0.5793      \\
MCQG-SRefine          & 0.4326      & 0.4322       \\ \hline
\end{tabular}
\caption{Performance comparison of GPT-4 and GPT-4 + MCQG-SRefine on Human Data and Machine Data.}
\label{tab:gpt4_mcqg_refine}
\end{table}

\onecolumn

\begin{longtable}{|m{15cm}|}
\hline
\textbf{Patient History}\\
An 84-year-old female with a past medical history of hypertension presented with weakness, dry cough, and shortness of breath for four days. The patient had received two doses of the COVID vaccine, with the second dose in March 2021. In the ER, her vital signs were BP 133/93, HR 103 bpm, RR 22 breaths/min, oxygen saturation of 96\% on 40 L per minute of supplemental oxygen via high-flow nasal cannula, and afebrile. Laboratory assessment is in Table. Nasopharyngeal swab for SARS-CoV-2 RNA was positive. Chest X-ray on admission shows worsening right pleural effusion with new opacity obscuring the lower two-third of the right lung and a new pleural-based opacity in the left upper lobe Figure. CT chest with contrast shows large right pleural effusion and associated right basilar consolidation and abdominal ascites. The patient was admitted to the telemetry unit and started on methylprednisolone, piperacillin-tazobactam, remdesivir, and baricitinib. The patient clinically deteriorated on Day 2 and was transferred to the intensive care unit for thoracentesis and possible intubation. Thoracentesis removed 1.95 L of bloody, serosanguineous fluid obtained, with partial resolution of the effusion Figure. On Day 3, the patient developed septic shock, florid renal failure, and lethargy and was started on intravenous fluids and norepinephrine drip. Chest X-ray showed near-complete opacification of bilateral lung fields and subsequently was intubated. On Day 4, tense ascites were noted and the patient underwent paracentesis, which removed 4.25 L of bloody, serosanguineous fluid. Renal replacement therapy started. The patient was deemed to have a guarded prognosis with multiorgan failure. \\ \hline

\textbf{Diagnosis}\\ \hline

\textbf{Causes and Mechanisms}\\
\begin{itemize}
    \item $\square$ the cause/infectious agent or predisposing factor(s)
    \item $\boxtimes$ underlying processes/pathways (pathophysiology of sepsis)
    \item $\square$ underlying anatomic structure or physical location
    \item $\boxtimes$ mechanisms, drugs (tazobactam, beta-lactamase inhibitors)
\end{itemize} \\ \hline

\textbf{Obtaining and Predicting History and Physical Examination}\\
\begin{itemize}
    \item $\square$ knows signs/symptoms of selected disorders
    \item $\square$ knows individual’s risk factors for development of condition
    \item $\square$ knows what to ask to obtain pertinent additional history
    \item $\boxtimes$ predicts the most likely additional physical finding (pleural effusion)
\end{itemize} \\ \hline

\textbf{Selecting and Interpreting Laboratory and Diagnostic Studies}\\
\begin{itemize}
    \item $\square$ select most appropriate laboratory or diagnostic study
    \item $\boxtimes$ interprets laboratory or other study findings (pleural fluid analysis)
    \item $\boxtimes$ predicts the most likely laboratory or diagnostic study result (pleural fluid protein/serum protein ratio $>$ 0.5)
    \item $\square$ most appropriate laboratory or diagnostic study after change in patient status
\end{itemize} \\ \hline

\textbf{Formulating the Diagnosis}\\
\begin{itemize}
    \item $\boxtimes$ select most likely diagnosis (correct answer: pleural effusion after pneumonia, sepsis, MODS)
\end{itemize} \\ \hline

\textbf{Determining Prognosis/Outcome}\\
\begin{itemize}
    \item $\square$ recognizes factors in the history, or physical or laboratory study findings
    \item $\square$ interprets laboratory or other diagnostic study results and identifies current/future status of patient
    \item $\boxtimes$ recognizes associated conditions of a disease (septic shock, MODS)
    \item $\square$ recognizes characteristics of disease relating to natural history or course of disease
\end{itemize} \\ \hline

\textbf{Management}\\ \hline

\textbf{Health Maintenance and Disease Prevention}\\
\begin{itemize}
    \item $\square$ risk factors for conditions amenable to prevention or detection
    \item $\square$ identifies patient groups at risk
    \item $\square$ knows common screening tests
    \item $\square$ selects appropriate preventive agent or technique
    \item $\square$ knows appropriate counseling regarding current and future problems
    \item $\square$ educates patients
\end{itemize} \\ \hline

\textbf{Selecting and Monitoring Pharmacotherapy}\\
\begin{itemize}
    \item $\square$ selects most appropriate pharmacotherapy
    \item $\square$ assesses patient adherence, recognizes techniques to increase adherence
    \item $\square$ recognizes factors that alter drug requirements
    \item $\square$ Knows adverse effects of various drugs or recognizes signs and symptoms of drug (and drug-drug) interactions
    \item $\square$ knows contraindications of various medications
    \item $\square$ knows modifications of a therapeutic regimen within the context of continuing care
    \item $\square$ appropriate monitoring to evaluate effectiveness of pharmacotherapy or adverse effects
\end{itemize} \\ \hline

\textbf{Clinical Interventions/Treatments}\\
\begin{itemize}
    \item $\boxtimes$ most appropriate management of selected conditions (sepsis)
    \item $\boxtimes$ immediate management or priority in management (septic shock)
    \item $\square$ follow-up or monitoring approach regarding the management plan
    \item $\boxtimes$ current/short-term management (sepsis)
    \item $\square$ severity of patient condition in terms of need for referral for surgical treatments/procedures
    \item $\square$ appropriate surgical management
    \item $\square$ preoperative/postoperative
\end{itemize} \\ \hline

\textbf{Selecting Clinical Interventions (Mixed Management)}\\
\begin{itemize}
    \item $\square$ Selecting Clinical Interventions (Mixed Management)
\end{itemize} \\ \hline

\textbf{Monitoring/Surveillance for Disease Recurrence or Progression}\\
\begin{itemize}
    \item $\square$ indications for surveillance for recurrence or progression of disease following treatment
    \item $\square$ how to monitor a chronic disease in a stable patient where a change in patient status might indicate a need to change therapy
    \item $\square$ most appropriate long-term treatment
\end{itemize} \\ \hline
\caption{Human annotation example for topics and test points.}
\label{tab:human_topic_keypoint_example}
\end{longtable}

\twocolumn
\begin{table*}[h!]
    \centering
    \begin{tabular}{p{0.8\textwidth}r}
        \toprule
        \textbf{Topic Name} & \textbf{Frequency} \\
        \midrule
        Select most likely diagnosis & 40 \\
        Selects most appropriate pharmacotherapy & 37 \\
        Select most appropriate laboratory or diagnostic study & 33 \\
        Most appropriate management of selected conditions & 27 \\
        Mechanisms, drugs & 21 \\
        Most appropriate long-term treatment & 21 \\
        Predicts the most likely laboratory or diagnostic study result & 20 \\
        Recognizes associated conditions of a disease & 15 \\
        The cause/infectious agent or predisposing factor(s) & 13 \\
        Underlying processes/pathways & 11 \\
        Knows signs/symptoms of selected disorders & 10 \\
        Current/short-term management & 8 \\
        Immediate management or priority in management & 7 \\
        The cause/infectious agent or predisposing factor & 7 \\
        Knows adverse effects of various drugs or recognizes signs and symptoms of drug (and drug-drug) interactions & 7 \\
        Appropriate surgical management & 6 \\
        Interprets laboratory or other study findings & 6 \\
        Underlying anatomic structure or physical location & 5 \\
        Predicts the most likely additional physical finding & 4 \\
        Interprets laboratory or other diagnostic study results and identifies current/future status of patient & 4 \\
        Selecting Clinical Interventions (Mixed Management) & 4 \\
        Follow-up or monitoring approach regarding the management plan & 4 \\
        Identifies patient groups at risk & 3 \\
        Selects appropriate preventive agent or technique & 3 \\
        Knows common screening tests & 3 \\
        Educates patients & 2 \\
        Select most likely diagnosis yes & 1 \\
        Selecting Clinical Interventions (Mixed Management) & 1 \\
        Risk factors for conditions amenable to prevention or detection & 1 \\
        Recognizes factors that alter drug requirements & 1 \\
        Indications for surveillance for recurrence or progression of disease following treatment & 1 \\
        Most appropriate long-term treatment & 1 \\
        Severity of patient condition in terms of need for referral for surgical treatments/procedures & 1 \\
        Knows what to ask to obtain pertinent additional history & 1 \\
        Preoperative/postoperative & 1 \\
        Knows contraindications of various medications & 1 \\
        Recognizes characteristics of disease relating to natural history or course of disease & 1 \\
        Appropriate monitoring to evaluate effectiveness of pharmacotherapy or adverse effects & 1 \\
        Predicts the most likely laboratory or diagnostic study result  & 1 \\
        \bottomrule
    \end{tabular}
    \caption{Frequency of Human identified topics}
    \label{tab:hum_frequency}
\end{table*}

\begin{table*}[h!]
    \centering
    \begin{tabular}{p{0.8\textwidth}r}
        \toprule
        \textbf{Topic Name} & \textbf{Frequency} \\
        \midrule
        Select most appropriate laboratory or diagnostic study & 66 \\
        The cause/infectious agent or predisposing factor(s) & 64 \\
        Selects most appropriate pharmacotherapy & 56 \\
        Appropriate surgical management & 41 \\
        Select most likely diagnosis & 37 \\
        Knows signs/symptoms of selected disorders & 37 \\
        Most appropriate management of selected conditions & 16 \\
        Most appropriate long-term treatment & 10 \\
        Underlying processes/pathways & 7 \\
        Mechanisms, drugs & 4 \\
        Knows individual’s risk factors for development of condition & 4 \\
        Appropriate monitoring to evaluate effectiveness of pharmacotherapy or adverse effects & 3 \\
        Educates patients & 2 \\
        Interprets laboratory or other diagnostic study results and identifies current/future status of patient & 2 \\
        Interpret laboratory or other study findings & 1 \\
        Recognizes factors that alter drug requirements & 1 \\
        Interprets laboratory or other study findings & 1 \\
        Underlying anatomic structure or physical location & 1 \\
        Knows modifications of a therapeutic regimen within the context of continuing care & 1 \\
        Risk factors for conditions amenable to prevention or detection & 1 \\
        Predicts the most likely additional physical finding & 1 \\
        Immediate management or priority in management & 1 \\
        Knows what to ask to obtain pertinent additional history & 1 \\
        Knows adverse effects of various drugs and signs and symptoms of drug (and drug-drug) interactions & 1 \\
        \bottomrule
    \end{tabular}
    \caption{Frequency of machine generated topics}
    \label{tab:mach_frequency}
\end{table*}

\onecolumn

\begin{longtable}{| p{15cm} |}
\hline
\textbf{Problem Statement} \\
Given a medical case \( n \) detailing a patient's history, diagnosis, treatment, and outcome, we aim to generate a USMLE question \( u \). \\
Here, \( u : <c, q, a, d> \) consists of: \\
\hspace{0.1em} - \textbf{context (\( c \))} which is a modified excerpt from $n$ tailored to align with the target style and obscure evidence information that can easily lead directly to the correct answer, based on a topic \( t \) and a keypoint \( k \);\\
\hspace{0.1em} - \textbf{question ($q$)} based on the generated context based on the generated context \( c \), which may be one or several sentences; \\
\hspace{0.1em} - \textbf{correct answer (\( a \))} to this question and, \\
\hspace{0.1em} - \textbf{distractor options (\( d \))} to challenge the test-taker. \\

Our framework consists of a MCQG-SRefine pipeline for generating the target \( u \). \\
\hline
\hline

\textbf{Input: Medical Cases} \\
A 67-year-old Caucasian female presented to our hospital with a chief complaint of persistent bright red blood per rectum. Her medical history was significant for hypertension, hyperlipidemia, diabetes mellitus type 2, coronary artery disease with three prior myocardial infarctions, recurrent cerebrovascular accidents requiring anticoagulation with warfarin, gastroesophageal reflux disease, asthma, and endometrial cancer status post radiation therapy. Fifteen months prior to the current presentation, the patient was noted to have a grade 1 endometrial adenoma but was not considered a good surgical candidate due to multiple comorbidities. Vaginal hysterectomy was considered but due to her long and narrow vagina, this option was deferred initially. Her only treatment option was radiation therapy and brachytherapy. She eventually underwent total abdominal hysterectomy with bilateral salpingo-oopherectomy due to continued pelvic pain. The patient denied any prior history of gastrointestinal (GI) bleeding. Her bleeding was described as one large episode of bright red blood per rectum associated with blood clots. She denied any abdominal pain, nausea, vomiting, diarrhea, constipation, or melena. The most recent colonoscopy was performed four months ago and revealed three diminutive polyps in the transverse colon with pathology confirming tubular adenoma. Her physical examination was significant for mild left-sided abdominal tenderness but was otherwise unremarkable. Rectal examination was notable for nonbleeding hemorrhoids and no visible blood. Blood work revealed white blood cells of 14.3k/uL (normal range 4.3-10.0 k/uL) and hemoglobin of 9.6 g/dL (normal range 11.8-14.8 g/dL), which is similar to the patient’s baseline. Creatinine was slightly elevated to 1.2 and blood urea nitrogen was elevated to 39. International normalized ratio was 2.0. Due the large volume of hematochezia and presence of anemia, the patient was admitted to the hospital and underwent a colonoscopy, which revealed a large, fungating, friable, and ulcerated nonobstructing mass in the sigmoid colon. The mass was noncircumferential, measured 4 cm in length, and was located 15-19 cm from the anal verge. Biopsies were obtained with cold forceps for histology and the proximal and distal margins of the mass were tattooed. Histology showed invasive and moderately differentiated carcinoma without visible goblet cells. Given the patient’s history of endometrial cancer, immunohistochemistry was performed and was consistent with an endometrial (endometrioid subtype) primary. Pan-computed tomography was negative for any obvious distant disease. The tumor marker carcinoembryonic antigen level was normal as well. The patient subsequently underwent surgical resection of the affected colon followed by radiation therapy. At a follow-up visit one year later, the patient was doing well without active GI complaints. \\
\hline

\textbf{Input: Topics} \\
The topics come from an official list of 16 target topics from USMLE. \\
We use both human annotated and model identified topics for generating USMLE questions. \\
Some example topics: \\
\hspace{1em} - Causes and Mechanisms \\
\hspace{1em} \hspace{1em} - the cause/infectious agent or predisposing factor(s) \\
\hspace{1em} \hspace{1em} - underlying processes/pathways \\
\hspace{1em} \hspace{1em} - underlying anatomic structure or physical location \\
\hspace{1em} \hspace{1em} - mechanisms, drugs \\
\hspace{1em} - Selecting and Interpreting Laboratory and Diagnostic Studies \\
\hspace{1em} \hspace{1em} - select most appropriate laboratory or diagnostic study \\
\hspace{1em} \hspace{1em} - interprets laboratory or other study findings \\
\hspace{1em} - Monitoring/Surveillance for Disease Recurrence or Progression \\
\hspace{1em} \hspace{1em} - most appropriate long-term treatment \textbf{(our topic for the current clinical note)} \\
\hline

\textbf{Input: Key Points} \\
The key points are the core concepts to which the target correct answer is highly related or is the same as the keypoint. \\
We use both human-annotated and model-identified key points to generate USMLE questions. \\
Example keypoints: \\
\hspace{1em} - covid-19 with copd \\
\hspace{1em} - Radiotherapy or chemotherapy \textbf{(our keypoint for the current clinical note)} \\
\hspace{1em} - ARDS \\
\hspace{1em} - osteonecrosis of the femoral head \\
\hspace{1em} - echocardiography for Ebstein’s anomaly \\
\hspace{1em} - COVID-19 pneumonia \\
\hline
\hline

\textbf{MCQG-SRefine INIT Step} \\
\textbf{Initiation Step:} \\
\hspace{0.1em} - Four sub-steps generate different components of the target output \( u \). \\
\hspace{0.1em} - Examples for few-shot prompting are retrieved from the COLBERT retriever model \( R \). \\
\hspace{0.1em} - These examples consist of USMLE questions from a question bank. \\
\textbf{Context-Question Pair Retrieval:} \\
\hspace{0.1em} - Use a query to retrieve the most relevant context-question pairs from \( R \). \\
\hspace{0.1em} - Prompt model \( M \) with these pairs, along with other relevant information, to generate the context \( c \). \\
\textbf{Question Retrieval:} \\
\hspace{0.1em} - Use the context to retrieve the next most relevant question from \( R \). \\
\hspace{0.1em} - Prompt model \( M \) with this question, along with other information, to generate the question \( q \). \\
\textbf{Iteration:} \\
\hspace{0.1em} - Repeat the process until generating the complete target output \( u \), including context, question, correct answer, and distractor options. \\
\hline

\textbf{MCQG-SRefine INIT Output} \\
\textbf{Context}: A 67-year-old Caucasian female with a significant medical history including hypertension, hyperlipidemia, diabetes mellitus type 2, coronary artery disease with three prior myocardial infarctions, recurrent cerebrovascular accidents, gastroesophageal reflux disease, asthma, and endometrial cancer presented with persistent bright red blood per rectum. She underwent radiation therapy and brachytherapy for her endometrial cancer but eventually required total abdominal hysterectomy with bilateral salpingo-oopherectomy due to persistent pelvic pain. She had no prior history of gastrointestinal bleeding. Her recent colonoscopy revealed three diminutive polyps in the transverse colon confirmed as tubular adenoma. Upon admission for an episode of hematochezia and anemia, she underwent a colonoscopy revealing a large, fungating, nonobstructing mass in the sigmoid colon. The biopsy was consistent with invasive and moderately differentiated carcinoma, and immunohistochemistry indicated an endometrial primary. The patient underwent surgical resection of the colon mass and subsequent radiation therapy. A follow-up visit a year later revealed no active GI complaints. \\
\textbf{Question}: What is the most appropriate long-term treatment for a 67-year-old female patient with a history of endometrial cancer who presented with a large, nonobstructing mass in the sigmoid colon due to invasive and moderately differentiated carcinoma? \\
\textbf{Correct answer}: Radiotherapy or chemotherapy \\
\textbf{Distractor options}: \\
\hspace{1em} - A: Oral corticosteroids \\
\hspace{1em} - B: Aspirin therapy \\
\hspace{1em} - C: Bisphosphonates \\
\hspace{1em} - D: Hormone replacement therapy \\
\hspace{1em} - E: Anticoagulation therapy \\
\hline
\hline

\textbf{MCQG-SRefine QUESTION ANSWER Step} \\
\hspace{0.1em} - In this step \( u \) is presented as a USMLE question for \( M \), i.e. \( u \) without the knowledge of the correct answer \( a \), the options presented are \( d \cup a \). \\
\hspace{0.1em} - \( M \) generates an attempted answer \( aa \) and a rationale which supports \( aa \) in the form of a reasoning \( r \) behind it. \\
\hline
\textbf{MCQG-SRefine QUESTION ANSWER Output} \\
\textbf{Attempted answer}:Radiotherapy or chemotherapy \\
\textbf{Reasoning}: The patient's history of endometrial cancer and the finding of a large, nonobstructing mass in the sigmoid colon are significant. The biopsy results revealing invasive and moderately differentiated carcinoma with immunohistochemistry indicating an endometrial primary suggest that this mass is likely a metastasis from her endometrial cancer. In such a case, systemic therapy with chemotherapy or targeted radiation therapy is often the most appropriate treatment to control the disease and prevent further metastases. \\
\hspace{1em} - Option A (Oral corticosteroids) is not typically used as a primary treatment for cancer. It may be used in certain situations to manage side effects of cancer or cancer treatment, but it wouldn't be the mainstay of treatment in this case.\\
\hspace{1em} - Option B (Aspirin therapy) is sometimes used as a preventive measure against colorectal cancer, but it is not a treatment for existing cancer. \\
\hspace{1em} - Option C (Bisphosphonates) are used to prevent bone loss and fractures in patients with certain types of cancer, such as breast and prostate cancer, that have spread to the bones. They are not typically used as a primary treatment for colorectal or endometrial cancer.\\
\hspace{1em} - Option D (Hormone replacement therapy) is typically used to manage symptoms of menopause, but it's not typically used to treat endometrial cancer, particularly in a patient with a history of this disease.
\hspace{1em} - Option E (Anticoagulation therapy) is used to treat or prevent blood clots, but it does not treat cancer itself.\\
Hence, based on the above discussion, the most appropriate long-term treatment for this patient would be radiotherapy or chemotherapy \\
\hline
\hline

\textbf{MCQG-SRefine CRITIQUE Step} \\
\hspace{0.1em} - After generating all the components in the initialization and answer steps, the model \( M \) is tasked with critiquing each component in this feedback step. \\
\hspace{0.1em} - The set of components to be critiqued, \( u_{critique} \), which includes the context (\( c \)), question (\( q \)), correct answer (\( a \)), distractor options (\( d \)), and rationale (\( r \)). \\
\hspace{0.1em} - Model \( M \) receives a scoring guide (\( G \)) containing all the aspects on which each component needs to be scored. \\
\hspace{0.1em} - The prompt also includes few-shot examples of already scored components and relevant information. \\
\hspace{0.1em} - The final output of this step is the feedback for all components, denoted as \( f \). It includes a short textual critique and a score for each aspect of the respective components. \\
\hline

\textbf{Rubric for the Feedback Output} \\
\textbf{Rubric for the Context:} \\
\hspace{1em} - \texttt{Relevant}: The context should be relevant to the topic provided. \\
\hspace{1em} - \texttt{Concision}: The context should be concise and shouldn't include extraneous information or just be a copy of the clinical note. \\
\hspace{1em} - \texttt{Coherent}: The context should be coherent enough and should organically be built for the question in the end. \\
\hspace{1em} - \texttt{Consistent}: The context is consistent with the information in the clinical note and the topic. \\
\hspace{1em} - \texttt{Specific}: The context should be specific and address the topic. \\
\hspace{1em} - \texttt{Fluent}: The context is fluent in terms of grammar and flow of words and ideas. \\
\hspace{1em} - \texttt{Clueing}: Instead of directly mentioning the diagnosis of a medical condition, it should be clued in through the mention of symptoms. \\
\hspace{1em} - \texttt{Completeness}: The context should be complete and free from gaps or missing information that could lead to ambiguity in answering the question accurately. \\
\hspace{1em} - \texttt{Misdirection}: Does the context avoid misleading the test taker intentionally or unintentionally? \\

\textbf{Rubric for the Question:} \\
\hspace{1em} - \texttt{Relevant}: The question should be answerable from the information provided in the context and should not be abrupt. \\
\hspace{1em} - \texttt{Clear}: The question should not be vague or ambiguous. \\
\hspace{1em} - \texttt{Concluding}: The flow of ideas from the context should organically result into the question. \\
\hspace{1em} - \texttt{Difficulty}: The question should not be too easy. \\
\hspace{1em} - \texttt{Clarity}: Is the question unambiguous to avoid incorrect interpretations caused by ambiguity or poor wording? \\

\textbf{Rubric for the Correct Answer:} \\
\hspace{1em} - \texttt{Relevant}: The correct answer should be either the keypoint or very related to it. \\
\hspace{1em} - \texttt{Occurrence}: The correct answer or any of its semantic or syntactic forms and directly related medical concepts should not occur in the context. \\
\hspace{1em} - \texttt{Justification}: Is the correct answer logically supported by the context and aligned with the provided information? \\
\hspace{1em} - \texttt{Depth of Understanding}: Does the correct answer demand nuanced understanding of the context or concepts, ensuring the test taker genuinely grasps the material? \\
\hspace{1em} - \texttt{Prevention of Guesswork}: Does the correct answer deter guessing and align with the context, avoiding common misconceptions? \\

\textbf{Rubric for the Distractors:} \\
\hspace{1em} - \texttt{Format}: The distractors should be in similar formats as the correct answer, i.e. if it is an abbreviation or an explanation. \\
\hspace{1em} - \texttt{Length}: The distractors should have similar length as the correct answer. \\
\hspace{1em} - \texttt{Relation}: The distractors should be related to the correct answer through some medical concepts or they should be the same kind of medical entities. \\
\hspace{1em} - \texttt{Variation}: Distractors should be distinct from each other and from the correct answer. \\
\hspace{1em} - \texttt{Plausibility}: Do the options align with the context and challenge critical thinking? \\
\hspace{1em} - \texttt{Differentiation}: Are the options distinct and does the correct answer clearly outshine the distractors based on context and available information? \\
\hspace{1em} - \texttt{Common Mistakes}: Distractors should align with common misconceptions to test genuine understanding? \\

\textbf{Rubric for the Chain of Thought/Reasoning:} \\
\hspace{1em} - \texttt{Logical Flow}: Does the chain of thought exhibit a coherent sequence of steps or considerations that are easy to follow? \\
\hspace{1em} - \texttt{Evidence-Based Reasoning}: Is the answer supported by evidence or information from the context, justifying the chosen response? \\
\hspace{1em} - \texttt{Consideration of Options}: Does the chain of thought demonstrate critical evaluation of each option, employing a systematic process to eliminate distractors with supporting evidence? \\
\hline

\textbf{Example of Distractors Feedback Output} \\
\hspace{0.1em} - \textbf{Format}: The format of the distractor options matches that of the correct answer. All options are treatment modalities, which is suitable for the question. \textbf{Score: 5/5}. \\
\hspace{0.1em} - \textbf{Length}: The length of the distractor options is similar to the length of the correct answer. \textbf{Score: 5/5}. \\
\hspace{0.1em} - \textbf{Relation}: The distractor options are related to the correct answer as they are all medical treatments, but not all are related to cancer treatment which is the specific context. \textbf{Score: 3/5}. \\
\hspace{0.1em} - \textbf{Variation}: The distractor options are distinct from each other and from the correct answer, which helps to test the test-taker's understanding of the context. \textbf{Score: 5/5}. \\
\hspace{0.1em} - \textbf{Plausibility}: The distractor options are plausible treatments for certain conditions, but not necessarily the condition described in the context, which is advanced cancer. \textbf{Score: 3/5}. \\
\hspace{0.1em} - \textbf{Differentiation}: The correct answer (radiotherapy or chemotherapy) clearly outshines the distractors based on the context and available information. \textbf{Score: 5/5}. \\
\hspace{0.1em} - \textbf{Common Mistakes}: The distractors do not align with common misconceptions about cancer treatment. They represent treatments for other conditions, not common mistakes in cancer management. \textbf{Score: 3/5}. \\
\textbf{Distractor Score}: Total Score: \textbf{29/35} \\
\hline
\hline

\textbf{MCQG-SRefine Correction Step} \\
\hspace{0.1em} - The feedback generated in the previous step is utilized to refine each of the generated components of the target output \( u \) from the initialization step. \\
\hspace{0.1em} - \( M \) is prompted with inputs including examples for refinement, the feedback (\( f \)), and the current components (\( c, q, a, d, aa, r \)). \\
\hspace{0.1em} - \( M \) generates \( u_{refined} \), aiming to improve its performance on all aspects critiqued in the feedback. \\
\hspace{0.1em} - \( u_{refined} \) is then subjected to the answer step and then to the feedback step. \\
\hspace{0.1em} - If the feedback scores surpass a threshold, the iterative refinement process halts. Otherwise, the process continues for up to four steps of feedback and refinement. \\
\hline

\textbf{MCQG-SRefine Correction Output} \\
\textbf{Context}: A 67-year-old female with a history of endometrial cancer presented with symptoms of rectal bleeding. She had previously undergone radiation therapy and brachytherapy for her cancer. On admission, a colonoscopy was performed due to her history and the presenting symptoms, which revealed a large mass in the sigmoid colon. Biopsy of the mass showed invasive and moderately differentiated carcinoma, with immunohistochemistry suggesting an endometrial primary. The patient underwent surgical resection of the mass followed by radiation therapy. \\
\textbf{Question}: Following surgical resection of a colon mass in a patient with a history of endometrial cancer, what would be the most appropriate long-term management approach? \\
\textbf{Correct answer}: radiotherapy or chemotherapy \\
\textbf{Distractor options}: \\
\hspace{1em} - a) oral corticosteroids \\
\hspace{1em} - b) aspirin therapy \\
\hspace{1em} - c) bisphosphonates \\
\hspace{1em} - d) hormone replacement therapy \\
\hspace{1em} - e) anticoagulation therapy \\
\hline
\caption{Demonstration of MCQG-SRefine Pipeline.}
\label{tab:demonstration_MCQG-SRefine}
\end{longtable}

\twocolumn

\onecolumn

\lstset{
    basicstyle=\ttfamily,
    columns=fullflexible,
    breaklines=true,
    postbreak=\mbox{\textcolor{black}{$\hookrightarrow$}\space}
}

\centering

\begin{longtable}{|p{1cm}|p{14cm}|}

\hline
\tiny{\textbf{Type}}  & \tiny{\textbf{Prompt}} \\ \hline
\tiny{INIT Prompt - Context} & 
{\tiny
\begin{lstlisting}
    USMLE context based questions: 
    Context and question: {context_question}
    Context and question: {context_question}
    Context and question: {context_question}
    Generate a context(not the question,in the format Context: ) based on the given topic from the clinical note:
    Clinical Note: {clinical_note}
    Topic: {topic}
    Keypoint: {keypoint}"
\end{lstlisting}
}
\\ \hline
\tiny{INIT Prompt - Question} &
{\tiny
\begin{lstlisting}
    USMLE context based questions: 
    Context and question: {context_question}
    Context and question: {context_question}
    Context and question: {context_question}
    Generate a one line question(in the format Question: ) based on the given context:
    Context: {context}
    Topic: {topic}
    Keypoint: {keypoint}"
\end{lstlisting}
}
\\ \hline
\tiny{INIT Prompt - Correct answer} &
{\tiny
\begin{lstlisting}
    USMLE context based questions with their correct answers: 
    Context and question: {context_question}
    Correct answer: {correct_answer}
    Context and question: {context_question}
    Correct answer: {correct_answer}
    Context and question: {context_question}
    Correct answer: {correct_answer}
    Generate the correct answer(in the format Correct answer: ) to the question based on the given context,topic and keypoint(to which it should be highly related to) :
    Context: {context}
    Question: {question}
    Topic: {topic}
    Keypoint: {keypoint}
    "
\end{lstlisting}
}
\\ \hline
\tiny{INIT Prompt - Correct answer} &
{\tiny
\begin{lstlisting}
    USMLE context based questions with their correct answers: 
    Context and question: {context_question}
    Correct answer: {correct_answer}
    Context and question: {context_question}
    Correct answer: {correct_answer}
    Context and question: {context_question}
    Correct answer: {correct_answer}
    Generate the correct answer(in the format Correct answer: ) to the question based on the given context,topic and keypoint(to which it should be highly related to) :
    Context: {context}
    Question: {question}
    Topic: {topic}
    Keypoint: {keypoint}
    "
\end{lstlisting}
}
\\ \hline
\tiny{INIT Prompt - Distractor options} &
{\tiny
\begin{lstlisting}
    USMLE context based questions with their correct answers and distractor options:  
    Context and Question: {question}
    Correct answer: {correct_answer}
    Distractor options: {distractor_options} 
    Generate distractor options(in the format Distractor options: ) for the context, question, and correct answer:
    Context: {context}
    Question: {question}
    Correct answer: {correct_answer}
    "
\end{lstlisting}
}
\\ \hline
\tiny{ANSWER Prompt} &
{\tiny
\begin{lstlisting}
    # few shot exemplars (X2)
    Context: {context}

    Question: {question}
    
    Options: {options}
    
    Correct answer: {answer}
    
    Reasoning: {reasoning}
    Answer the USMLE question and provide a step by step reasoning for reaching that particular answer and rejecting other options
    Context: {context}

    Question: {question}

    Options: {options}

    Correct answer: 
\end{lstlisting}
}
\\ \hline
\tiny{CRITIQUE Prompt} &
{\tiny
\begin{lstlisting}
    # Example USMLE component generated from the clinical note and its respective feedback:
    Clinical note:{clinical_note}
    Topic: {topic}
    Keypoint: {keypoint}
    Context:{context}
    Question:{question}
    Correct answer:{correct_answer}
    Distractor options:{distractor_options}
    {component_name} feedback: {feedback}
    {component_name} score: {score}
    In addition to the scoring rubrics in the examples above,give feedback and score the {component_name} using the attempted answer's(correct/incorrect) reasoning-based rubrics and their definitions below.
    Please include both the previous scoring rubrics and the following reasoning-based rubrics before giving the feedback for a particular aspect and add up the scores for all the aspects for the total scores of the {component_name}. 
    Many of these feedback points for the {component_name} depend upon the reasoning and the attempted answer correctness so consider that while giving feedback for the {component_name}.
    {component_name} reasoning-based rubrics: {reasoning_rubrics}
    Give the output in just this format: {format_instructions}
    Output just the JSON instance and nothing else.
    Clinical note: {clinical_note}
    Keypoint: {keypoint}
    Topic: {topic}
    Context: {context}
    Question: {question}
    Correct answer: {correct_answer}
    Attempted answer: {attempted_answer}
    Reasoning: {reasoning}
    Distractor options: {distractor_options}
\end{lstlisting}
}
\\ \hline
\tiny{CRITIQUE Prompt - Reasoning} &
{\tiny
\begin{lstlisting}
    Give supporting textual feedback for each aspect and score(out of 5 for each aspect, in the format "2/5" if the score for that aspect is 2, also give supporting evidence for that score) the {component_name} using the attempted answer's(correct/incorrect) reasoning-based rubrics and their definitions below.
    Please include the following reasoning-based rubrics before giving the feedback for a particular aspect and add up the scores for all the aspects for the total score of the {component_name}. 
    Many of these feedback points for the {component_name} depend upon the reasoning and the attempted answer correctness so consider that while giving feedback for the {component_name}.
    {component_name} rubrics: {reasoning_rubrics}
    Give the output in just this format: {format_instructions}
    Output just the JSON instance and nothing else.
    Clinical note: {clinical_note}
    Keypoint: {keypoint}
    Topic: {topic}
    Context: {context}
    Question: {question}
    Correct answer: {correct_answer}
    Attempted answer: {attempted_answer}
    Reasoning: {reasoning}
    Distractor options: {distractor_options}
\end{lstlisting}
}
\\ \hline
\tiny{CORRECTION Prompt} &
{\tiny
\begin{lstlisting}
    # few shot exemplars (x3), Eeach example is an improved version of the previous
    Context: {context}

    Question: {question}
    
    Correct answer: {correct_answer}
    
    Distractor Options: {distractor_options}
    
    Feedback for the above components:
    
    Context feedback: {context_feedback}
    
    Context score: {context_score}
    
    Question feedback: {question_feedback}
    
    Question score: {question_score}
    
    Correct answer feedback: {correct_answer_feedback}
    
    Correct answer score: {correct_answer_score}
    
    Distractor options feedback: {distractor_option_feedback}
    
    Distractor options score: {distractor_option_score}

    Improve the context,question, correct answer and distractor options using each previous components' feedback and the reasoning feedback.
    Generate a context, question, correct answer and distractor options that can achieve high scores on all the above feedback rubrics, given the clinical note, keypoint and topic. Do not generate the feedback for any of the component.:

    
    Clinical Note: {clinical_note}

    Topic: {topic}
    
    Keypoint: {keypoint}

    Context: {context}

    Question: {question}
    
    Attempted answer: {attempted_answer}
    
    Reasoning: {reasoning}
    
    Correct answer: {correct_answer}
    
    Distractor Options: {distractor_options}
    
    Feedback on the generated content with respect to various rubrics.
    
    Context feedback: {context_feedback}
    
    Context score: {context_score}
    
    Question feedback: {question_feedback}
    
    Question score: {question_score}
    
    Correct answer feedback: {correct_answer_feedback}
    
    Correct answer score: {correct_answer_score}
    
    Distractor options feedback: {distractor_option_feedback}
    
    Distractor options score: {distractor_option_score}
    
    Reasoning feedback: {reasoning_feedback}
    
    Reasoning score: {reasoning_score}
\end{lstlisting}
}

\\ \hline

\caption{All Prompts.}
\label{prompt}
\end{longtable}

\twocolumn

\onecolumn

\begin{longtable}{| p{15cm} |}
\hline
\textbf{Evaluation Guide} \\
\hline
Your task is to evaluate two system outputs, each consisting of a machine-generated USMLE question. Each question has four components: context, a question based on the context, the correct answer, and distractor options. \\\\

The questions are generated using a clinical (EHR) note, a topic from the official USMLE topic list, and a keypoint. The keypoint is a specific aspect on which the student is tested, either related to or being the correct answer itself. \\\\

You have to annotate the following: \\
1. \textbf{Preference:} Please indicate your preference for the better question in the \textbf{‘Preference’} column by writing \textbf{1} or \textbf{2}, signifying \textbf{Question 1} or \textbf{Question 2}, respectively. In the Remark column please give a short reasoning behind your choice. \\
2. \textbf{Difficulty level:} For each Question 1 and Question 2, you have to adjudge which level the question falls under: \textbf{Easy}, \textbf{Medium}, or \textbf{Hard}. Please indicate your choice in the respective Question difficulty level column, i.e., \textbf{Question 1 difficulty} and \textbf{Question 2 difficulty}. \\
3. \textbf{Errors:} For both Question 1 and Question 2, highlight any spans having errors, be it logical, syntactical, or medical. An example error could be if the question contradicts the clinical note. Write any such errors if they exist in the \textbf{Error} column. \\\\

Below are some rubrics you can use to evaluate the two outputs. You don't need to assess each question on every metric; simply use these as a guideline to understand what makes a good USMLE question before you start evaluating: \\\\

\textbf{Metrics for the context:} \\
\hspace{1em} - \texttt{Relevant:} The context should be relevant to the topic provided. \\
\hspace{1em} - \texttt{Concision:} The context should be concise and shouldn't include extraneous information or just be a copy of the clinical note. \\
\hspace{1em} - \texttt{Coherent:} The context should be coherent enough and should organically be built for the question in the end. \\
\hspace{1em} - \texttt{Consistent:} The context is consistent with the information in the clinical note and the topic. \\
\hspace{1em} - \texttt{Specific:} The context should be specific and address the topic. \\
\hspace{1em} - \texttt{Fluent:} The context is fluent in terms of grammar and flow of words and ideas. \\
\hspace{1em} - \texttt{Clueing:} Instead of directly mentioning the diagnosis of a medical condition, it should be clued in through the mention of symptoms. \\
\hspace{1em} - \texttt{Completeness:} The context should be complete and free from gaps or missing information that could lead to ambiguity in answering the question accurately. \\
\hspace{1em} - \texttt{Misdirection:} Does the context avoid misleading the test taker intentionally or unintentionally? \\\\

\textbf{Metrics for the question:} \\
\hspace{1em} - \texttt{Relevant:} The question should be answerable from the information provided in the context and should not be abrupt. \\
\hspace{1em} - \texttt{Clear:} The question should not be vague or ambiguous. \\
\hspace{1em} - \texttt{Concluding:} The flow of ideas from the context should organically result in the question. \\
\hspace{1em} - \texttt{Difficulty:} The question should not be too easy. \\
\hspace{1em} - \texttt{Clarity:} Is the question clear and unambiguous to avoid incorrect interpretations caused by ambiguity or poor wording? \\\\

\textbf{Metrics for the correct answer:} \\
\hspace{1em} - \texttt{Relevant:} The correct answer should be either the keypoint or very related to it. \\
\hspace{1em} - \texttt{Occurrence:} The correct answer or any of its semantic or syntactic forms and directly related medical concepts should not occur in the context. \\
\hspace{1em} - \texttt{Justification:} Is the correct answer logically supported by the context and aligned with the provided information? \\
\hspace{1em} - \texttt{Depth of Understanding:} Does the correct answer demand a nuanced understanding of the context or concepts, ensuring the test taker genuinely grasps the material? \\
\hspace{1em} - \texttt{Prevention of Guesswork:} Does the correct answer deter guessing and align with the context, avoiding common misconceptions? \\\\

\textbf{Metrics for the distractors:} \\
\hspace{1em} - \texttt{Format:} The distractors should be in similar formats as the correct answer, i.e., if it is an abbreviation or an explanation. \\
\hspace{1em} - \texttt{Length:} The distractors should have a similar length as the correct answer. \\
\hspace{1em} - \texttt{Relation:} The distractors should be related to the correct answer through some medical concepts, or they should be the same kind of medical entities. \\
\hspace{1em} - \texttt{Variation:} Distractors should be distinct from each other and from the correct answer. \\
\hspace{1em} - \texttt{Plausibility:} Do the options align with the context and challenge critical thinking? \\
\hspace{1em} - \texttt{Differentiation:} Are the options distinct, and does the correct answer clearly outshine the distractors based on context and available information? \\
\hspace{1em} - \texttt{Common Mistakes:} Distractors should align with common misconceptions to test genuine understanding. \\
\hline
\caption{Human evaluation guide.}
\label{tab:human_eval_guide}
\end{longtable}

\twocolumn

\onecolumn

\begin{table*}[!ht]
\centering
\begin{tabular}{c|rrrrrr|c}
\hline
Cases & gpt-4o & gpt4 & gpt3.5 & opus & sonnet & haiku & Expert X \\
\hline
1 & 1 & 2 & 1 & 1 & 1 & -1 & 1 \\
2 & 1 & 1 & -1 & -1 & 2 & 2 & 2 \\
3 & -1 & 2 & -1 & -1 & 2 & -1 & 2 \\
4 & 1 & 2 & -1 & -1 & -1 & 2 & 2 \\
5 & -1 & -1 & 2 & 1 & -1 & 1 & 2 \\
6 & 1 & -1 & 2 & 1 & 1 & -1 & 2 \\
7 & -1 & 1 & -1 & 2 & 2 & 2 & 0 \\
8 & 1 & 1 & -1 & -1 & -1 & -1 & 1 \\
9 & 1 & 1 & -1 & -1 & -1 & 2 & 2 \\
10 & -1 & 1 & -1 & 2 & -1 & 2 & 2\\
11 & -1 & -1 & -1 & -1 & -1 & 1 & 2 \\
12 & 2 & -1 & 2 & 1 & -1 & 2 & 2 \\
13 & 2 & 2 & 2 & 1 & 1 & -1 & 1 \\
14 & 2 & 1 & -1 & -1 & 1 & 2 & 2 \\
15 & 1 & -1 & 2 & -1 & -1 & 2 & 2 \\
16 & -1 & -1 & 2 & 1 & -1 & 2 & 0 \\
17 & 1 & 2 & 2 & 1 & 2 & -1 & 2 \\
18 & -1 & 2 & 2 & 1 & -1 & 2 & 2 \\
19 & -1 & -1 & 2 & 1 & 2 & -1 & 2 \\
20 & 1 & -1 & -1 & -1 & 2 & 2 & 2 \\
21 & -1 & 1 & -1 & -1 & -1 & 2 & 2 \\
22 & 1 & -1 & 1 & 1 & -1 & 2 & 0 \\
23 & 1 & 1 & 1 & -1 & 1 & 2 & 2 \\
24 & -1 & -1 & 1 & 1 & -1 & 2 & 0 \\
25 & -1 & -1 & 2 & 1 & -1 & 1 & 2 \\
26 & 1 & 2 & 2 & 1 & 2 & -1 & 2 \\
27 & 2 & 1 & -1 & -1 & -1 & -1 & 1 \\
28 & 1 & 2 & -1 & -1 & 1 & 1 & 1 \\
29 & 2 & 1 & 2 & 1 & 2 & -1 & 1 \\
30 & -1 & -1 & 1 & 1 & -1 & 2 & 1 \\
31 & 1 & -1 & 1 & 1 & -1 & -1 & 2 \\
32 & -1 & 1 & -1 & 1 & 2 & -1 & 2 \\
33 & 1 & 2 & 1 & -1 & -1 & 2 & 2 \\
34 & -1 & 1 & 1 & -1 & -1 & -1 & 2 \\
35 & -1 & 1 & 1 & -1 & -1 & -1 & 2 \\
36 & 1 & 2 & 1 & -1 & -1 & 2 & 2 \\
37 & -1 & 1 & -1 & -1 & 2 & -1 & 1 \\
38 & -1 & 1 & -1 & -1 & -1 & -1 & 2 \\
39 & 1 & -1 & -1 & 1 & -1 & -1 & 2 \\
40 & 1 & 2 & 1 & 1 & 2 & 2 & 2 \\
\hline
\end{tabular}
\caption{LLM-as-judge (comparison). 0 means tie, 1 means preferring GPT-4, 2 means preferring MCQG-SRefine, and -1 means position bias (LLM has different preferences in two different orders).
}
\label{tab:llm_as_judge_comparison_full}
\end{table*}

\twocolumn

\onecolumn

\begin{longtable}{| p{15cm} |}
\hline
\\
\large{\textbf{Case Study 1}} \\ 
\\
\hline
\\
\textbf{Clinical Note:} A 35-year-old man presented with the complaints of low back ache for 12 years, radiating to right leg for 4 months and numbness extending to lateral side of the sole of right foot. On examination, there was a 30\% sensory loss in right S1 dermatome as compared with contralateral limb, with no bladder bowel involvement. Patient had no motor deficit. Magnetic resonance imaging (MRI) of the lumbosacral spine was suggestive of IM cystic lesion at L2-3 hypointense on T1-weighted images and hyperintense on T2-weighted images. MRI brain did not reveal any abnormality. Lumbar puncture and serologic studies were not performed.
With the differential diagnosis of neoplastic lesion, the patient was taken up for posterior laminectomy. L2-3 laminectomy was done. A dural bulge was identified. On durotomy, the cord was found to be enlarged. Under microscopic guidance, posterior longitudinal myelotomy was done, the cysts were approached, and subtotal resection of cysts was done. Intraoperatively, three grayish white cysts were identified. Cysts were found to be adherent to the nerve roots causing their inflammation. As a result, one of the cysts could not be excised and was only decompressed. The remaining two cysts were completely excised. Histopathology revealed it to be NCC.
The patient improved postoperatively. Back pain was relieved, and there was significant reduction in radiating pain. He was started on albendazole (15 mg/kg body weight) for 4 weeks and steroids for 2 weeks. The patient was discharged on the 4th post-operative day. He was followed-up biweekly for the first month. Thereafter, monthly follow-up was done for the next 2 months. MRI done at 6 months confirmed resolution of the cystic lesion. Thereafter, 6 monthly follow-up was done. Patient is symptom free and not on any medication at 2 years of follow-up.\\

\hspace{1em} - \textbf{Topic:} 	select most appropriate laboratory or diagnostic study \\

\hspace{1em} - \textbf{Keypoint:} MRI \\\\
\hline
\\
\textbf{GPT-4 generated USMLE question:} \\
Context: A 35-year-old man has been experiencing low back ache for 12 years, with pain radiating to his right leg for the past 4 months. He also reports numbness extending to the lateral side of the sole of his right foot. Upon examination, a 30\% sensory loss in right S1 dermatome was found compared to the contralateral limb, but there was no bladder or bowel involvement and no motor deficit. An MRI of his lumbosacral spine revealed an IM cystic lesion at L2-3, which was hypointense on T1-weighted images and hyperintense on T2-weighted images. No abnormalities were found in an MRI of the brain. The patient underwent a posterior laminectomy and L2-3 laminectomy, during which three grayish-white cysts were identified. These cysts were found to be adhering to the nerve roots, causing inflammation. One cyst could not be completely excised and was only decompressed, while the other two were completely removed. Histopathology confirmed the diagnosis of NCC. Postoperative improvement was noted in the patient, with relief from back pain and significant reduction in radiating pain. He was prescribed albendazole and steroids and was discharged on the 4th post-operative day. Follow-ups were conducted biweekly for the first month, monthly for the next two months, and then every six months. An MRI taken six months post-operatively confirmed the resolution of the cystic lesion. After two years, the patient is symptom-free and not on any medication.\\
Question: What is the most appropriate diagnostic study to confirm the resolution of a cystic lesion in the lumbosacral spine?\\
Correct answer: MRI \\
Distractor options: \\
\hspace{1em} - A : CT scan \\
\hspace{1em} - B : Ultrasound \\
\hspace{1em} - C : X-ray \\
\hspace{1em} - D : EEG \\
\\
\hline
\\
\textbf{MCQG-SRefine generated USMLE question} \\
Context: A 35-year-old patient had been experiencing lower back pain for over a decade, which recently started radiating to his right leg and was accompanied by numbness in his right foot. Although the patient had no motor deficit, a 30\% sensory loss was detected in the right S1 dermatome. A key diagnostic tool was utilized to identify the cause of his symptoms.\\
Question: Which imaging technique was pivotal in diagnosing the patient's condition and guiding subsequent treatment?\\
Correct answer: MRI \\
Distractor options: \\
\hspace{1em} - A : CT scan \\
\hspace{1em} - B : Ultrasound \\
\hspace{1em} - C : X-ray \\
\hspace{1em} - D : EEG \\ 
\\
\hline
\\
\textbf{Human Evaluation:}\\
\textbf{Expert X}: \\
Preference: \\
\hspace{1em}MCQG-SRefine \\
Difficulty: \\
\hspace{1em}GPT-4: Easy \\
\hspace{1em}MCQG-SRefine: Hard \\
\textbf{Reasons:} \\
\hspace{1em} 1. Question (GPT-4) consists of the answer to the question in the context. \\
\hspace{1em} 2. Question (MCQG-SRefine) context more Concision. \\\\

\hline

\\
\large{\textbf{Case Study 2}} \\ 
\\
\hline
\\
\textbf{Clinical Note:} This 33-year-old female patient had typical COVID-19 symptoms such as high fever, dry cough, headache, and dyspnea about 1 week before ICU admission. She was intubated and proned due to rapid respiratory deterioration. For the following 6 days, her situation was unstable, and physical therapy consisted of prone positioning and prevention of secondary damage. From day 7 onwards, she started to improve rapidly and could be mobilized passively into a side-edge position. After extubation, she presented postextubation dysphagia and severe ICU-acquired weakness (MRC-SS 36/60). She also suffered from pronounced delirium and anxiety and said repeatedly that she had been abducted and that she believed she had to die. She seemed to feel threatened by us and it was difficult to calm her down. Due to the pandemic measures of the Swiss government, hospital visits were not generally allowed, but because her anxiety was limiting her rehabilitation, her husband was granted an exceptional permission to visit her. This seemed to give the patient a short sense of security, and she started to participate in some basic functional activities (CPAx 21/50). Nevertheless, the delirium did not resolve upon her transfer to a peripheral acute hospital.\\

\hspace{1em} - \textbf{Topic:} 	the cause/infectious agent or predisposing factor(s) \\

\hspace{1em} - \textbf{Keypoint:} covid-19 \\\\
\hline
\\
\textbf{GPT-4 generated USMLE question:} \\
Context: A 33-year-old female patient presented with typical symptoms of COVID-19, including high fever, dry cough, headache, and dyspnea. She was admitted to the ICU due to rapid respiratory deterioration and was intubated and proned. Despite physical therapy and prone positioning, her condition remained unstable for 6 days. From the seventh day, she started showing signs of improvement and was passively mobilized into a side-edge position. However, post-extubation, she experienced dysphagia and severe ICU-acquired weakness along with pronounced delirium and anxiety. Her rehabilitation was further hampered due to her delirium, which led to her belief of being abducted and fear of death. Due to her heightened anxiety, her husband was granted special permission to visit her, which provided her with a temporary sense of security and encouraged her to participate in basic functional activities. Despite this, her delirium persisted even after her transfer to a peripheral acute hospital.\\
Question: What infectious agent is responsible for the symptoms and complications experienced by the 33-year-old female patient?\\
Correct answer: SARS-CoV-2 (COVID-19)\\
Distractor options: \\
\hspace{1em} - A : Influenza A virus \\
\hspace{1em} - B : Human immunodeficiency virus (HIV) \\
\hspace{1em} - C : Mycobacterium tuberculosis \\
\hspace{1em} - D : Streptococcus pneumoniae \\
\\
\hline
\\
\textbf{MCQG-SRefine generated USMLE question} \\
Context: A 33-year-old female patient was admitted to the ICU following the onset of high fever, dry cough, headache, and dyspnea. Following intubation, her condition remained unstable for nearly a week. Upon extubation, she displayed severe weakness, dysphagia, and exhibited signs of delirium and anxiety.\\
Question: Given the patient's symptoms and the course of her illness, what is the most probable causative agent of her condition?\\
Correct answer: sars-cov-2 (covid-19)\\
Distractor options: \\
\hspace{1em} - A : Influenza a virus  \\
\hspace{1em} - B : Pneumococcus bacteria  \\
\hspace{1em} - C : Mycobacterium tuberculosis \\
\hspace{1em} - D : Epstein-Barr virus \\ 
\\
\hline
\\
\textbf{Human Evaluation:}\\
\textbf{Expert X}: \\
Preference: \\
\hspace{1em}GPT-4 \\
Difficulty: \\
\hspace{1em}GPT-4: Easy \\
\hspace{1em}MCQG-SRefine: Hard \\
\textbf{Reasons:} \\
\hspace{1em} 1. According to the context in Question (MCQG-SRefine), it is hard to get the right answer. Question (MCQG-SRefine) tries to make it harder, but at the same time, it becomes confusing.\\
\hspace{1em} 2. The answer to Question (MCQG-SRefine) can be all the following distractors. Because there is no evidence to suspect the infection of COVID-19. 
Question (GPT-4), however, starts the context by giving evidence that the patient was already diagnosed with Covid-19. But the answer also has COVID-19, so this is an error of the question. \\\\

\hline

\\
\large{\textbf{Case Study 3}} \\ 
\\
\hline
\\
\textbf{Clinical Note:} A 35-year-old male patient presented with a six-month history of pain in the anterior chest and neck. His pain initially had been mild but had become severe in the last three weeks. He was afebrile and did not have any constitutional symptoms. On examination, marked tenderness of the right sternoclavicular joint was noted. Initial laboratory investigations, tuberculosis workup, and chest X-rays were unremarkable. However, further study with MRI revealed subchondral bone marrow edema and enhancement involving the medial end of the right clavicle. In addition, mild effusion of the right sternoclavicular joint and surrounding soft-tissue edema was also seen (Figure , ). Given the patient's radiological and clinical findings, suspicion of SAPHO syndrome was raised. He was given an initial trial of paracetamol and NSAIDs and he showed remarkable improvement on his monthly follow-up visits. Though a biopsy was offered, the patient refused to undergo one and has shown no relapse to date thanks to the dramatic response to NSAIDs.\\

\hspace{1em} - \textbf{Topic:} 	select most likely diagnosis yes \\

\hspace{1em} - \textbf{Keypoint:} SAPHO syndrome \\
\\
\hline
\\
\textbf{GPT-4 generated USMLE question:} \\
Context: A 35-year-old male patient has been experiencing anterior chest and neck pain for six months, which has become severe in the last three weeks. Despite being afebrile and showing no constitutional symptoms, the right sternoclavicular joint was notably tender on examination. Initial investigations, including laboratory tests, tuberculosis workup, and chest X-rays, yielded no remarkable findings. However, an MRI revealed subchondral bone marrow edema and enhancement at the medial end of the right clavicle, along with mild effusion of the right sternoclavicular joint and surrounding soft-tissue edema. Given these clinical and radiological findings, the suspicion of SAPHO syndrome was raised. The patient was started on paracetamol and NSAIDs, which led to a significant improvement in his condition during subsequent monthly follow-ups. Although a biopsy was considered, the patient declined. He has remained free of relapse due to the effective response to NSAIDs.\\
Question: Based on the clinical and radiological findings, what is the most likely diagnosis for a 35-year-old male patient experiencing severe anterior chest and neck pain, and showing signs of subchondral bone marrow edema and enhancement at the medial end of the right clavicle on an MRI?\\
Correct answer: SAPHO syndrome \\
Distractor options: \\
\hspace{1em} - A : Rheumatoid arthritis \\
\hspace{1em} - B : Osteomyelitis. \\
\hspace{1em} - C : Gout. \\
\hspace{1em} - D : Bone cancer. \\
\hspace{1em} - E : Paget's disease of bone. \\
\\
\hline
\\
\textbf{MCQG-SRefine generated USMLE question} \\
Context: A 35-year-old male has been dealing with pain in his anterior chest and neck area for the past half a year, which has recently escalated in severity. Despite his discomfort, he had no other systemic symptoms and was afebrile. Upon physical examination, severe tenderness was noted in the right sternoclavicular joint. While initial investigations including laboratory tests and chest radiographs showed no abnormalities, an MRI scan revealed edema and enhancement in the bone marrow beneath the cartilage at the medial end of the right clavicle. There were also signs of mild effusion in the right sternoclavicular joint and surrounding soft tissue edema. After being administered paracetamol and NSAIDs, the patient showed significant improvement in his condition.\\
Question: What diagnosis does the clinical presentation and the results of the MRI scan most likely indicate?\\
Correct answer: SAPHO syndrome\\
\hspace{1em} - A : Rheumatoid arthritis  \\
\hspace{1em} - B : Osteomyelitis \\
\hspace{1em} - C : Gout \\
\hspace{1em} - D : Bone cancer \\ 
\hspace{1em} - E : Paget's disease of bone \\ 
\\
\hline
\\
\textbf{Human Evaluation:}\\
\textbf{Expert X}: \\
Preference:\\
\hspace{1em}MCQG-SRefine \\
Difficulty: \\
\hspace{1em}GPT-4: Easy \\
\hspace{1em}MCQG-SRefine: Hard \\
\textbf{Reasons:} \\
\hspace{1em} 1. In Question (GPT-4), the answer was directly listed in the context. \\
\hspace{1em} 2. In Question (MCQG-SRefine), only the signs/symptoms for SAPHO syndrome are given. " \\\\

\hline
\\
\large{\textbf{Case Study 4}} \\ 
\\
\hline
\\
\textbf{Clinical Note:} A 35-year-old male patient presented with a six-month history of pain in the anterior chest and neck. His pain initially had been mild but had become severe in the last three weeks. He was afebrile and did not have any constitutional symptoms. On examination, marked tenderness of the right sternoclavicular joint was noted. Initial laboratory investigations, tuberculosis workup, and chest X-rays were unremarkable. However, further study with MRI revealed subchondral bone marrow edema and enhancement involving the medial end of the right clavicle. In addition, mild effusion of the right sternoclavicular joint and surrounding soft-tissue edema was also seen (Figure , ). Given the patient's radiological and clinical findings, suspicion of SAPHO syndrome was raised. He was given an initial trial of paracetamol and NSAIDs and he showed remarkable improvement on his monthly follow-up visits. Though a biopsy was offered, the patient refused to undergo one and has shown no relapse to date thanks to the dramatic response to NSAIDs.\\

\hspace{1em} - \textbf{Topic:} 	knows signs/symptoms of selected disorders \\

\hspace{1em} - \textbf{Keypoint:} SAPHO syndrome \\
\\
\hline
\\
\textbf{GPT-4 generated USMLE question:} \\
Context: A 35-year-old male presents with a six-month history of anterior chest and neck pain that has progressively worsened over the last three weeks. The patient is afebrile and does not display any constitutional symptoms. Upon examination, significant tenderness is noted in the right sternoclavicular joint. Initial laboratory tests, including a tuberculosis workup and chest X-rays, do not reveal any abnormalities. However, an MRI shows subchondral bone marrow edema and enhancement at the medial end of the right clavicle, along with mild effusion of the right sternoclavicular joint and surrounding soft-tissue edema. The patient's clinical and radiological findings lead to a suspicion of SAPHO syndrome. An initial treatment with paracetamol and NSAIDs results in significant improvement, and the patient continues to show no signs of relapse. Despite the offer of a biopsy, the patient chooses not to undergo the procedure. \\
Question: What are the common signs and symptoms of SAPHO syndrome? \\
Correct answer: Anterior chest and neck pain, tenderness in the sternoclavicular joint, subchondral bone marrow edema and enhancement, mild effusion of the sternoclavicular joint and surrounding soft-tissue edema. \\
Distractor options: \\
\hspace{1em} - A : Swelling and redness over the affected joint, accompanied by intense pain. \\
\hspace{1em} - B : Chronic diarrhea, abdominal pain, and weight loss. \\
\hspace{1em} - C : Presence of skin rashes, dry mouth and eyes. \\
\hspace{1em} - D : Persistent cough, chest pain, and difficulty in breathing. \\
\hspace{1em} - E : Severe headache, nausea, and light sensitivity. \\
\\
\hline
\\
\textbf{MCQG-SRefine generated USMLE question} \\
Context: A 35-year-old male patient has been experiencing pain in his chest and neck for six months. This pain has been intensifying over the past three weeks. He does not have any fever or other systemic symptoms. During the physical exam, we noticed significant tenderness in his right sternoclavicular joint. While initial lab tests and chest X-rays didn't reveal anything unusual, an MRI showed signs of subchondral bone marrow edema in the medial end of the right clavicle, a mild effusion in the right sternoclavicular joint, and some soft-tissue edema surrounding the joint. These clinical and radiological findings are suggestive of a particular condition. The patient was treated with paracetamol and NSAIDs, which led to a significant improvement in his symptoms.\\
Question: Based on the described signs and symptoms, which syndrome is most likely to be suspected in this patient?\\
Correct answer: sapho syndrome \\
Distractor options: \\
\hspace{1em} - A : Ehlers-danlos syndrome  \\
\hspace{1em} - B : Marfan syndrome \\
\hspace{1em} - C : Reiter's syndrome \\
\hspace{1em} - D : Guillain-barre syndrome \\ 
\\
\hline
\\
\textbf{Human Evaluation:}\\
\textbf{Expert X}: \\
Preference: \\
\hspace{1em}MCQG-SRefine \\
Difficulty: \\
\hspace{1em}GPT-4: Easy \\
\hspace{1em}MCQG-SRefine: Hard \\
\textbf{Reasons:} \\
\hspace{1em} 1. For Question (GPT-4), the common signs and symptoms of SAPHO syndrome are not Anterior chest and neck pain, tenderness in the sternoclavicular joint, subchondral bone marrow edema and enhancement, mild effusion of the sternoclavicular joint and surrounding soft-tissue edema. This is just the symptom for the giving case. \\
\hspace{1em} 2. Question (MCQG-SRefine) consists the symptoms of SAPHO and asks to choose the right diagnosis for the patient which is a common USMLE type question. Question (GPT-4) also is a good way to test whether the test-taker knows the signs/symptoms of SAPHO. But the context is too long, and the answers are a little bit obvious. \\
\hspace{1em} 3. Question (GPT-4) is the missing context for diagnosis, but Question (MCQG-SRefine)'s  Distractor options are not that relevant and need correction. \\
\hspace{1em} 4. In the Question (MCQG-SRefine) context, there is a hint about the signs and symptoms of the patient's disease. \\\\
\hline

\caption{Human evaluation case study.}
\label{tab:human_eval_case}
\end{longtable}

\twocolumn

\onecolumn

\begin{table*}[!ht]
\centering
\begin{tabular}{cccc}
\hline
Component & Aspect & Percentage agreement & Cohen's kappa \\
\hline

Context	& consistent & 10.8108 & -0.0008196 \\
Context	& misdirection & 10.52631 & -0.001550 \\
Context & coherent & 26.31578 & 0.065026 \\
Context & clueing & 28.947368 & 0.0393258 \\
Context & concision & 26.3157 & 0.04488 \\
Context & relevant & 18.42105 & 0.010084 \\
Context & specific & 15.7894 & -2.22044 \\
Context & completeness & 13.15789 & 0.0079113 \\
Context & fluent & 10.5263 & -0.00310 \\
\hline
Question & difficulty & 27.50000 & -0.01665 \\
Question & relevant & 27.50000 & -0.03386 \\
Question & clarity & 40.0 & 0.08309 \\
Question & clear & 42.5 & 0.05056 \\
Question & concluding & 52.5 & 0.197465 \\
\hline
Correct Answer & depth of understanding & 30.7692 & -0.02631 \\
Correct Answer & relevant & 12.8205 & 0.008968 \\
Correct Answer & justification & 12.8205 & 0.008968 \\
Correct Answer & occurrence & 28.20512 & 0.07379 \\
Correct Answer & prevention of guesswork & 38.461 & 0.0370 \\
\hline
Distractor & variation & 10.2564 & 0.0 \\
Distractor & plausibility & 31.578 & 0.023715 \\
Distractor & common mistakes & 33.333 & 0.10582 \\
Distractor & length & 19.4444 & 0.035120 \\
Distractor & differentiation & 17.9487 & -0.012165 \\
Distractor & relation & 28.205 & -0.02535 \\
Distractor & format & 12.820 & -0.06763 \\
\hline
Reasoning & logical flow & 7.5 & -0.069364 \\
Reasoning & correctness & 7.5 & -0.048158 \\
Reasoning & consideration of options & 10.0 & -0.0020876 \\
Reasoning & evidence_based_reasoning & 5.0 & -0.081081 \\

\hline

\end{tabular}
\caption{Aspect vs Expert-X Preference correlation (Human Eval 40 data).}
\label{tab:aspect_human_corr}
\end{table*}

\begin{table*}[!ht]
\centering
\begin{tabular}{ccc}
\hline
Component & Percentage agreement & Cohen's kappa \\
\hline

Context & 0.38235 & 0.0338 \\
Question & 0.5 & 0.05472 \\
Correct Answer & 0.47058 & 0.15522 \\
Distractor & 0.3235 & -0.0358 \\
Reasoning & 0.08823 & -0.07168 \\

\hline

\end{tabular}
\caption{Component vs Expert-X Preference correlation (Human Eval 40 data). The correct answer is the highest, maybe because it is easier to judge a correct answer since it is the shortest lengthwise, and in case it is right or wrong, it is very easy to reject a particular question, whereas other components are considerably more subjective.}
\label{tab:component_human_corr}
\end{table*}

\twocolumn

\onecolumn

\begin{table*}[!ht]
\centering
\scalebox{0.85}{
\begin{tabular}{c|cccc}
\hline
& Component & Aspect & Percentage Agreement & Cohen's kappa \\
\hline

1 & question & concluding & 52.5 & 0.19746568109820484 \\
2 & question & clear & 47.5 & 0.08296943231441045 \\
3 & question & clarity & 47.5 & 0.05084745762711862 \\
4 & correct answer & prevention of guesswork & 62.5 & 0.1724137931034483\\
5 & distractor option & common mistakes of guesswork & 62.5 & 0.1689750692520774\\
6 & distractor option & plausibility & 60.0 & 0.10987482614742705 \\
7 & correct answer & depth of understanding & 60.0 & 0.11724137931034484 \\
8 & context & clueing & 65.0 & 0.1384615384615384 \\
9 & correct answer & occurrence of guesswork & 65.0 & 0.19308357348703176\\
10 & distractor option & relation & 67.5 & 0.21921921921921916\\
11 & question & relevant & 70.0 & 0.3083573487031701 \\
12 & question & difficulty & 70.0 & 0.30535455861070904 \\
13 & context & coherent & 72.5 & 0.34230194319880425 \\
14 & context & concision & 72.5 & 0.34230194319880425\\
15 & distractor option & length & 70.0 & 0.30535455861070904 \\
16 & context & relevant & 70.0 & 0.3083573487031701 \\
17 & distractor option & differentiation & 70.0 & 0.3083573487031701 \\
18 & context & specific & 70.0 & 0.3083573487031701 \\
19 & context & completeness & 70.0 & 0.3083573487031701 \\
20 & correct answer & relevant & 70.0 & 0.3083573487031701\\
21 & correct answer & justification & 70.0 & 0.3083573487031701\\
22 & distractor option & format & 67.5 & 0.22619047619047628 \\
23 & context & consistent & 67.5 & 0.22619047619047628 \\
24 & context & misdirection & 67.5 & 0.22619047619047628 \\
25 & context & fluent & 67.5 & 0.22619047619047628 \\
26 & distractor option & variation & 67.5 & 0.22619047619047628 \\
27 & reasoning & consideration of options & 67.5 & 0.22619047619047628 \\
28 & reasoning & logical_flow & 67.5 & 0.22619047619047628 \\
29 & reasoning & correctness & 67.5 & 0.22619047619047628 \\
30 & reasoning & evidence based reasoning & 67.5 & 0.22619047619047628\\

\hline

\end{tabular}
}
\caption{Percentage agreement based Greedy aspect filtering.}
\label{tab:aspect_filtering_PA}
\end{table*}

\begin{table*}[!ht]
\centering
\scalebox{0.85}{
\begin{tabular}{c|cccc}
\hline
& Component & Aspect & Percentage Agreement & Cohen's kappa \\
\hline

1 & question & concluding & 52.5 & 0.19746568109820484 \\
2 & distractor option & common mistakes & 60.0 & 0.2660550458715596 \\
3 & question & clarity & 57.5 & 0.16971916971916978 \\
4 & correct answer & occurrence & 67.5 & 0.3603936039360395 \\
5 & context & coherent & 72.5 & 0.4233289646133682 \\
6 & question & clear & 65.0 & 0.2660550458715596 \\
7 & context & concision & 67.5 & 0.31847968545216254 \\
8 & context & clueing & 65.0 & 0.24119241192411933 \\
9 & correct answer & prevention of guesswork & 70.0 & 0.30535455861070904 \\
10 & distractor option & length & 70.0 & 0.30535455861070904\\
11 & distractor option & plausibility & 70.0 & 0.3083573487031701 \\
12 & context & relevant & 70.0 & 0.30535455861070904 \\
13 & correct answer & relevant & 70.0 & 0.30535455861070904 \\
14 & correct answer & justification & 70.0 & 0.30535455861070904 \\
15 & context & completeness & 70.0 & 0.30535455861070904 \\
16 & distractor_option & variation & 70.0 & 0.3083573487031701 \\
17 & context & specific & 67.5 & 0.2737430167597765 \\
18 & context & consistent & 67.5 & 0.2737430167597765 \\
19 & context & misdirection & 67.5 & 0.2737430167597765 \\
20 & reasoning & consideration of options & 67.5 & 0.2737430167597765 \\
21 & context & fluent & 67.5 & 0.2737430167597765 \\
22 & distractor option & differentiation & 67.5 & 0.2737430167597765 \\
23 & question & difficulty & 67.5 & 0.2737430167597765 \\
24 & distractor option & relation & 67.5 & 0.22619047619047628 \\
25 & correct answer & depth of understanding & 70.0 & 0.25811437403400295 \\
26 & question & relevant & 70.0 & 0.3083573487031701 \\
27 & reasoning & correctness & 70.0 & 0.3083573487031701 \\
28 & distractor option & format & 67.5 & 0.22619047619047628 \\
29 & reasoning & logical flow & 67.5 & 0.22619047619047628 \\
30 & reasoning & evidence based reasoning & 67.5 & 0.22619047619047628\\

\hline

\end{tabular}
}
\caption{Cohen's kappa-based Greedy aspect filtering.}
\label{tab:aspect_filtering_CK}
\end{table*}

\twocolumn

\onecolumn

\begin{longtable}{p{1.5cm}lcc}

\hline
& Combination & Percentage Agreement & Cohen's kappa \\
\hline
Top 1 & question - concluding & 52.5 &  0.19746 \\
\hline
Top 2 & question - concluding & 67.5 & 0.39953 \\
& correct answer - occurrence & & \\
\hline
Top 3 & question - concluding & 72.5 & 0.44584 \\
& context - clueing & & \\
& correct answer - prevention of guesswork & & \\
\hline
Top 4 & question - concluding & 75.0 & 0.47984 \\
& context - concision & & \\
& correct answer - prevention of guesswork & & \\
& correct answer - occurrence & & \\
\hline
Top 5 & question - concluding & 77.5 & 0.51612 \\
& context - coherent & & \\
& correct answer - occurrence & & \\
& correct answer - depth of understanding & & \\
& distractor - common mistakes & & \\
\hline
Top 6 & question - concluding & 77.5 & 0.51612 \\
& context - coherent & & \\
& context - consistent & & \\
& correct answer - occurrence & & \\
& correct answer - depth of understanding & & \\
& distractor - common mistakes & & \\
\hline
Top 7 & question - concluding & 77.5 & 0.51612 \\
& context - coherent & & \\
& context - consistent & & \\
& context - misdirection & & \\
& correct answer - occurrence & & \\
& correct answer - depth of understanding & & \\
& distractor - common mistakes & & \\
\hline
Top 8 & question - concluding & 80.0 & 0.50541 \\
& question - relevant & & \\
& context - clueing & & \\
& context - concision & & \\
& correct answer - prevention of guesswork & & \\
& correct answer - occurrence & & \\
& correct answer - depth of understanding & & \\
& distractor - plausibility & & \\
\hline
Top 9 & question - concluding & 80.0 & 0.50541 \\
& question - relevant & & \\
& context - consistent & & \\
& context - clueing & & \\
& context - concision & & \\
& correct answer - occurrence & & \\
& correct answer - depth of understanding & & \\
& correct answer - prevention of guesswork & & \\
& distractor - plausibility & & \\
\hline
Top 10 & question - concluding & 80.0 & 0.50541\\
& question - relevant & & \\
& context - clueing & & \\
& context - completeness & & \\
& context - misdirection & & \\
& context - concision & & \\
& correct answer - occurrence & & \\
& correct answer - prevention of guesswork & & \\
& correct answer - depth of understanding & & \\
& distractor - plausibility & & \\
\hline

\caption{Percentage Agreement-based All-Combination aspect filtering.}
\label{tab:aspect_filtering_all_combination_PA}
\end{longtable}

\begin{longtable}{p{1.5cm}lcc}
\hline
& Combination & Percentage Agreement & Cohen's kappa \\
\hline
Top 1 & question - concluding & 52.5 &  0.19746 \\
\hline
Top 2 & question - concluding & 67.5 & 0.39953 \\
& correct answer - occurrence & & \\
\hline
Top 3 & question - concluding & 72.5 & 0.44584 \\
& correct answer - depth of understanding & & \\
& correct answer - occurrence & & \\
\hline
Top 4 & question - concluding & 75.0 & 0.47984 \\
& correct answer - depth of understanding & & \\
& correct answer - relevant & & \\
& correct answer - occurrence & & \\
\hline
Top 5 & question - concluding & 77.5 & 0.51612 \\
& context - coherent & & \\
& correct answer - occurrence & & \\
& correct answer - depth of understanding & & \\
& distractor - common mistakes & & \\
\hline
Top 6 & question - concluding & 77.5 & 0.51612 \\
& context - coherent & & \\
& context - consistent & & \\
& correct answer - occurrence & & \\
& correct answer - depth of understanding & & \\
& distractor - common mistakes & & \\
\hline
Top 7 & question - concluding & 77.5 & 0.51612 \\
& context - coherent & & \\
& context - consistent & & \\
& context - misdirection & & \\
& correct answer - occurrence & & \\
& correct answer - depth of understanding & & \\
& distractor - common mistakes & & \\
\hline
Top 8 & question - concluding & 77.5 & 0.51612 \\
& context - coherent & & \\
& context - consistent & & \\
& context - misdirection & & \\
& context - completeness & & \\
& correct answer - occurrence & & \\
& correct answer - depth of understanding & & \\
& distractor - common mistakes & & \\
\hline
Top 9 & question - concluding & 80.0 & 0.53890 \\
& question - clarity & & \\
& context - concision & & \\
& context - relevant & & \\
& correct answer - occurrence & & \\
& correct answer - depth of understanding & & \\
& distractor - common mistakes & & \\
& reasoning - evidence based reasoning & & \\
& reasoning - logical flow & & \\
\hline
Top 10 & question - concluding & 80.0 & 0.53890 \\
& question - clarity & & \\
& context - concision & & \\
& context - relevant & & \\
& context - misdirection & & \\
& correct answer - occurrence & & \\
& correct answer - depth of understanding & & \\
& distractor - common mistakes & & \\
& reasoning - evidence based reasoning & & \\
& reasoning - logical flow & & \\
\hline

\caption{Cohen's kappa-based All-Combination aspect filtering.}
\label{tab:aspect_filtering_all_combination_CK}
\end{longtable}

\end{document}